\pdfoutput=1 

\documentclass[10pt,twocolumn,letterpaper]{article}

\usepackage[pagenumbers]{cvpr} 

\usepackage{graphicx}
\usepackage{amsmath}
\usepackage{amssymb}
\usepackage{booktabs}
\usepackage{multirow}
\usepackage{bbding}
\usepackage{adjustbox}
\usepackage{xspace}
\usepackage{xcolor}

\newenvironment{packed_item}{
\begin{itemize}
\vspace{-8pt}
  \setlength{\itemsep}{0pt}
  \setlength{\parskip}{0pt}
  \setlength{\parsep}{0pt}
  \setlength{\topsep}{-10pt}
  \setlength{\partopsep}{0pt}
}{\end{itemize}}

\def\netname{SEDNet\xspace}
\def\kg{$\mathcal{L}_{log}$\xspace}

\newcommand{\best}[1]{\textbf{#1}}

\usepackage[pagebackref,breaklinks,colorlinks]{hyperref}

\def\cvprPaperID{2228} 
\def\confName{CVPR}
\def\confYear{2023}

\begin{document}

\title{Learning the Distribution of Errors in Stereo Matching \\ for Joint Disparity and Uncertainty Estimation}

\author{Liyan Chen\qquad Weihan Wang\qquad Philippos Mordohai\\
Stevens Institute of Technology\\
}

\maketitle


\pdfoutput=1
\begin{abstract}

We present a new loss function for joint disparity and uncertainty estimation in deep stereo matching. Our work is motivated by the need for precise uncertainty estimates and the observation that multi-task learning often leads to improved performance in all tasks. We show that this can be achieved by requiring the distribution of uncertainty to match the distribution of disparity errors via a KL divergence term in the network's loss function. A differentiable soft-histogramming technique is used to approximate the distributions so that they can be used in the loss. We experimentally assess the effectiveness of our approach and observe significant improvements in both disparity and uncertainty prediction on large datasets. Our code is available at \url{https://github.com/lly00412/SEDNet.git}.

\end{abstract}

\section{Introduction}
\label{sec:intro}
\pdfoutput=1

Many computer vision problems can be formulated as estimation tasks. Considering, however, that even high-performing estimators are not error-free, associating confidence or uncertainty with their estimates is of great importance, particularly in critical applications. In this paper, we focus on disparity estimation via stereo matching, but we are confident that our approach is applicable to other pixel-wise regression tasks after minor modifications.

We \textit{distinguish between confidence and uncertainty}: the former refers to a probability or likelihood of correctness, while the latter is related to the magnitude of the expected error of an estimate. Confidence can be used to reject estimates that are suspected to be incorrect, or to rank them  from most to least reliable. We argue that uncertainty is more valuable because it can also used for fusing multiple observations, e.g. in a Kalman filtering framework.
Most research has focused on confidence estimation for stereo matching \cite{hu2012quantitative,poggi2021confidence}. Moreover, most methods estimate confidence for pre-computed disparities that are not further improved. Joint estimation of disparity and confidence, which benefits both due to multi-task learning, is addressed infrequently \cite{shaked2017improved,kim2018unified,kim2020adversarial,mehltretter2022joint}.

Our work is partially inspired by the joint treatment of epistemic and aleatoric uncertainty by Kendall and Gal \cite{kendall2017uncertainties}, who propose novel loss functions that give rise to uncertainty estimates in pixel-wise vision tasks. Results on semantic segmentation and single-image depth estimation demonstrate how the primary task benefits from simultaneous uncertainty estimation.
Kendall and Gal argue that ``in many big data regimes (such as the ones common to deep learning with image data), it is most effective to model aleatoric uncertainty," while epistemic uncertainty can be reduced when large amounts of data are available. Here, we restrict our attention to aleatoric uncertainty. 

Our motivation is that ideally we should be able to predict the magnitude of the estimator's error at each pixel. Of course, this is unrealistic, since if it was possible, we could drive all errors down to zero. A feasible objective is to train an \textit{uncertainty estimator whose outputs follow the same distribution as the true errors of the disparity estimator.}

\newcommand\widf{.5\columnwidth}

\begin{figure}[b]
\begin{adjustbox}{width=\columnwidth,center}
\centering
\begin{tabular}{cc}
Left Image & Predicted Disparity \\
\includegraphics[width=\widf]{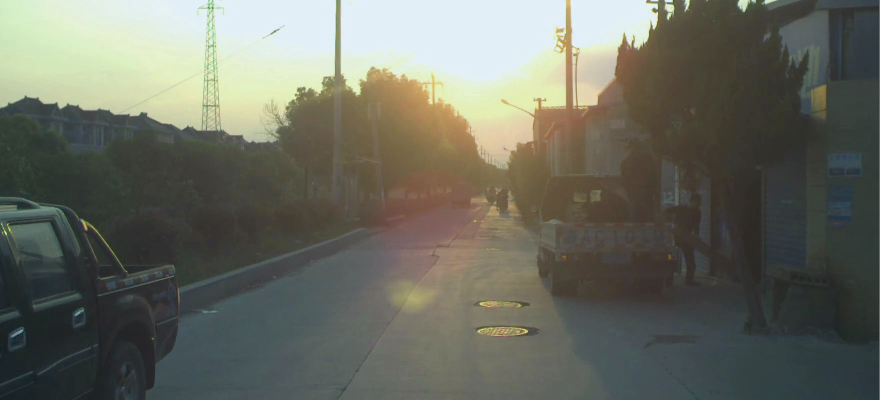} &
\includegraphics[width=\widf]{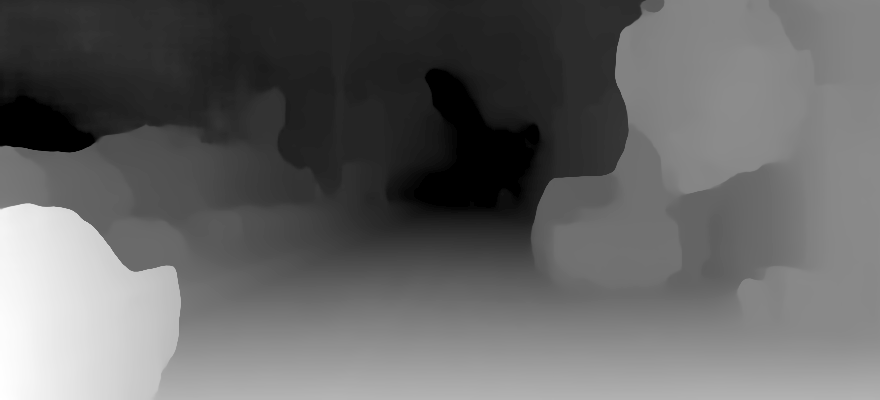} \\

\includegraphics[width=\widf]{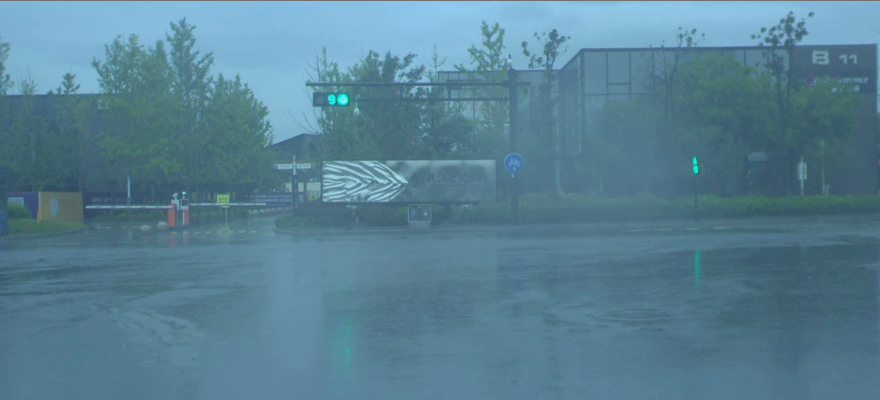} &
\includegraphics[width=\widf]{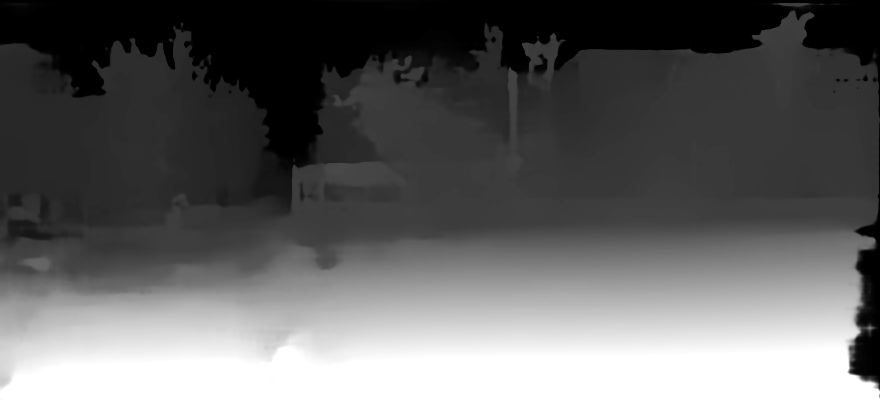} \\

\end{tabular}
\end{adjustbox}
\vspace{-4pt}
\caption{Examples of left images and predicted disparity maps by \netname on DrivingStereo \cite{Yang_2019_CVPR}. The first example is taken around sunset with over-exposure. The second example is taken on a rainy day with under-exposure. In both challenge cases, \netname predicts accurate disparity.}
\label{fig:sednet_ds}
\end{figure}

In this paper, we present an implementation of this concept via a deep network that jointly estimates disparity and its uncertainty from a pair of rectified images. We named the network \textbf{\netname}, for \textit{Stereo Error Distribution Network}. \netname includes a novel, lightweight uncertainty estimation subnetwork that predicts the aleatoric uncertainty of stereo matching, %
and a new loss  to match the distribution of uncertainties with that of disparity errors. To generate the inputs to this new loss, we approximate the distributions from the samples of disparity errors and uncertainty values in a differentiable way via a soft-histogramming technique.

We present extensive experimental validation of \netname's performance in disparity estimation and uncertainty prediction on large datasets with ground truth. \netname is superior to baselines with similar,  even identical, architecture, but without the proposed loss function.
Our main contributions are:
\begin{packed_item}
    \item a novel uncertainty estimation subnetwork that extracts information from the intermediate multi-resolution disparity maps generated by the disparity subnetwork,
    \item a differentiable soft-histogramming technique used to approximate the distributions of disparity errors and estimated uncertainties,
    \item a loss based on KL divergence applied on histograms obtained with the above technique.
\end{packed_item}

\section{Related Work}
\label{sec:related_works}
\pdfoutput=1

We refer readers to recent surveys on deep stereo matching \cite{poggi2021synergies} and on confidence estimation \cite{poggi2021confidence}. Here we summarize the most relevant publications to our work.

Stereo matching networks operate on a cost volume, which aggregates 2D features at each potential disparity for every pixel, 
and 
can be constructed via correlation or concatenation.
Correlation-based networks such as DispNetC~\cite{mayer2016large}, iResNet~\cite{liang2018learning_1st_Rob} and SegStereo~\cite{yang2018segstereo}, generate a single-channel correlation map between features extracted from the two views at each disparity level, favoring computation efficiency at the expense of losing the structural and semantic information in the feature representation. Concatenation-based networks, such as GCNet~\cite{kendall2017-gcnet}, PSMNet~\cite{chang2018psmnet} and GANet~\cite{zhang2019ga}, assemble features from both views at the disparity specified by the corresponding element of the cost volume. This promotes learning of contextual features but requires more parameters and a subsequent aggregation network.

We select GwcNet~\cite{guo2019group} as the foundation of our network. GwcNet takes a hybrid approach by reducing the dimension of the unary feature channels before concatenation in the cost volume. This is accomplished by a  \textit{Group Wise Correlation layer}, which takes as input $N_c$ unary feature channels, divides them into $N_g$ groups, computes the correlation between channels in each group at all disparity levels, and uses the resulting correlation scores to form the cost volume.
This reduces the size of the cost volume and the computational cost of 3D convolutions by a factor of $N_c:N_g$, with $N_g$ much smaller than $N_c$,
but still provides rich similarity-measure features to the disparity estimator.

Researchers have also focused on model reliability. In Bayesian Neural Networks (BNNs), different models are sampled from the distribution of weights to estimate the mean and variance of the target distribution in an empirical manner, yielding estimates of uncertainty~\cite{neal2012bayesian,Gal2016Uncertainty}. 
Additional empirical strategies such as Bootstrapped Ensembles~\cite{lakshminarayanan2017simple} and Monte Carlo Dropout~\cite{gal2016dropout} also sample from the distribution of weights.
On the other hand, Graves~\cite{graves2011practical} and Blundell et al.~\cite{blundell2015weight} proposed to replace the sampling with variational inference. 

Due to the high cost of training BNNs, methods for modeling the uncertainty or confidence in a predictive manner have also attracted interest. 
We distinguish between confidence and uncertainty: confidence is a binary variable trained with the BCE loss, while uncertainty is a continuous variable trained with L1 or L2 loss.
Nix and Weigend~\cite{nix1994estimating} introduce NNs with one output for model prediction and one for data noise (aleatoric uncertainty). 
In addition to aleatoric uncertainty which captures the data noise of the observations, epistemic uncertainty, which accounts for the uncertainty of the model parameters, can also be modeled ~\cite{der2009aleatory}.
To capture both types, 
Kendall et al.~\cite{kendall2017uncertainties, kendall2018multi} proposed to combine empirical and predictive methods in a joint framework. 

CNNs have been used to estimate confidence in stereo matching. The Confidence CNN (CCNN)~\cite{poggi2016bmvc}, Patch Based Confidence Prediction (PBCP)~\cite{seki2016patch}, the Early Fusion Network (EFN) and the Late Fusion Network (LFN)~\cite{fu2017stereo} and Multi Modal CNN (MMC)\cite{fu2018learning} only use small patches of the disparity maps. Conversely, the Global Confidence Network (ConfNet) and the Local-Global Confidence Network (LGC)~\cite{tosi2018beyond} introduce U-Net like architectures and take both the image and disparity map as inputs.
As a baseline, we use the Locally Adaptive Fusion Network (LAF)~\cite{Kim_2019_CVPR} which predicts the confidence map based on tri-modal inputs: the cost and disparity maps and the color image. An extension based on knowledge distillation has also been published \cite{kim2022stereo}.
These strategies are effective and  cheaper than the empirical ones, since they only require one forward pass. 

Only a subset of the confidence estimation literature has focused on joint disparity and confidence estimation.
The Reflective Confidence Network (RCN)~\cite{shaked2017improved} is the first to combine a disparity and a confidence loss.
The Unified Confidence Network (UCN)~\cite{kim2018unified} and the Adversarial Confidence Network (ACN)~\cite{kim2020adversarial} jointly estimate confidence and disparity from pre-computed cost volumes. UCN is self-supervised,
while ACN combines a generative cost aggregation network and a discriminative confidence estimation network in an adversarial manner. 
Mehltretter \cite{mehltretter2022joint} presents an approach that predicts both epistemic and aleatoric uncertainty using a Bayesian Neural Network, based on GCNet \cite{kendall2017-gcnet}.
KL divergence is used to measure the distance between the approximation of the distribution of network parameters estimated by variational difference and the exact posterior distribution.
(It should be noted that we use KL divergence for a completely different purpose on the distribution of disparity errors.)

Relevant research in adjacent areas of computer vision includes the work of Poggi et al.~\cite{Poggi_2020_CVPR} who comprehensively evaluate uncertainty estimation for self-supervised monocular depth estimation. 
Ilg et al.~\cite{ilg2018uncertainty} study empirical ensembles, predictive models and predictive ensembles as uncertainty models for optical flow estimation.

\section{Method}
\label{sec:method}
\pdfoutput=1

\begin{figure*}[t]
	\centering
	\includegraphics[width=2\columnwidth]{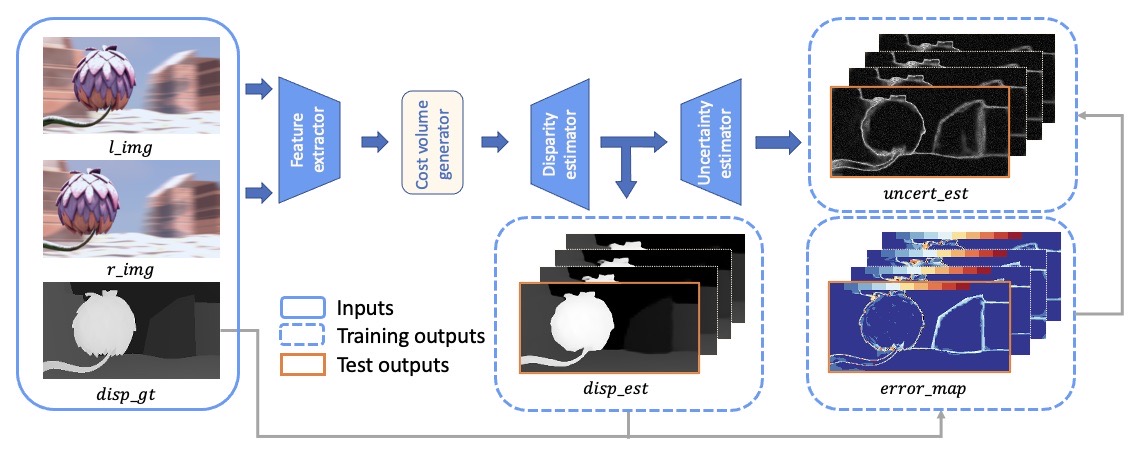}	
	\caption{An illustration of \netname. The stereo matching subnetwork consists of the feature extractor, cost volume generator and disparity estimator. It takes a rectified pair of images as input and predicts disparity maps at multiple resolutions. The uncertainty estimation subnetwork takes the predicted disparity maps as input and predicts corresponding uncertainty maps. The error maps, between the ground truth and predicted disparity, are used to supervise uncertainty estimation. 
	During training, the network keeps the output at all resolutions, but returns only the highest resolution disparity and uncertainty maps during testing.}
	\label{fig:Pipline}
\end{figure*}

The objective of our work is to jointly estimate the disparity and its uncertainty.
An important benefit of this joint formulation is that the multi-task network learns to predict more accurate disparities than the standalone disparity estimator when the uncertainty subnetwork is added.
Given a stereo image pair $\mathbf{X} = \{\mathbf{x}_{l},\mathbf{x}_{r}\}$, with image dimensions $H\times W$, and the corresponding ground truth disparity $\mathbf{d}$, the prediction ${\Hat{\mathbf{d}}}$ of a stereo-matching network ${f_{\theta}}$ can be represented as $\hat{\mathbf{d}} = f_{\theta}(\mathbf{x}_{l},\mathbf{x}_{r})$.
For each pixel $i$, the error $\epsilon^{(i)}$ of the prediction is calculated using the L1 loss.
 
Kendall and Gal \cite{kendall2017uncertainties} use the negative log-likelihood of the prediction model as the loss function to be minimized in pixel-wise tasks.
We take the formulation a step further by requiring that the network generate a distribution of uncertainties that matches the distribution of errors. To this end, we propose to minimize the divergence $\mathcal{D}$ between the distributions of predicted uncertainty and actual disparity error.

In the following subsections, we present aleatoric uncertainty estimation (Section~\ref{sec:kg_model}), the proposed KL divergence loss (Section~\ref{sec:sed_loss}), our network architecture (Section~\ref{sec:sednet}), and the combined loss function (Section~\ref{sec:total_loss}).

\subsection{Aleatoric Uncertainty Estimation}
\label{sec:kg_model}

In order to predict uncertainty and reduce the impact of noise, Kendall and Gal \cite{kendall2017uncertainties} minimize the pixel-wise negative log-likelihood 
of the prediction model, assuming that it follows a Gaussian distribution.
The subsequent work of Ilg et al.~\cite{ilg2018uncertainty} shows that the predicted distribution can be modeled as either Laplacian or Gaussian depending on whether the L1 or L2 loss is used for disparity estimation. 
Since we use the former, we can write the prediction model as:
\begin{equation}
\label{eq: s_distribution}
    p(\mathbf{d}|f_\theta(\mathbf{X})) = \mathcal{L}aplace(f_\theta(\mathbf{X}), \sigma)
\end{equation}
where the mean is given by the model output and  $\sigma$ is the observation noise scalar.

To model aleatoric uncertainty, Kendall and Gal \cite{kendall2017uncertainties} introduce pixel-specific noise parameters $\sigma^{(i)}$. We follow the approach of Ilg et al. \cite{ilg2018uncertainty}, who do the same for a Laplacian model, and obtain  the following pixel-wise loss function:
\begin{equation}
    \mathcal{L}_{log} = \frac{1}{n}\sum^n_{i=1} \frac{|\hat{\mathbf{d}}^{(i)} - \mathbf{d}^{(i)}|}{\exp(\mathbf{s}^{(i)})}  + \frac{1}{n}\sum^n_{i=1} \mathbf{s}^{(i)}\label{eq:log_loss}
\end{equation}
where $\hat{\mathbf{d}}^{(i)}$ and $\mathbf{d}^{(i)}$ are the predicted and ground truth disparity for pixel $i$, $\mathbf{s}^{(i)}$ is the log of the observation noise scalar $\mathbf{\sigma}^{(i)}$, and $n$ is equal to the total number of the pixels. 
Equation (\ref{eq:log_loss}) may be viewed as a robust loss function where the residual loss for a pixel is attenuated by its uncertainty, while the second term acts as a regularizer. 
We follow the authors' suggestion and train the network to predict the log of the observation noise scalar, $s$, for numerical stability.

\subsection{Matching the Distribution of Errors}
\label{sec:sed_loss}

Training a model using Eq. (\ref{eq:log_loss}) as the loss improves disparity estimation accuracy and favors uncertainty estimates correlated with the errors. Ideally, we would like each uncertainty estimate to be a precise predictor of the corresponding disparity error. Since this is infeasible, \textit{we would like the distribution of uncertainties to match the distribution of errors}. 

The Kullback-Leibler (KL) divergence \cite{kullback1951information} is a natural choice for measuring the dissimilarity between the distribution of $\epsilon$ and that of $\sigma$. 
Since the KL divergence is asymmetric, we choose the distribution of $\epsilon$ as the reference. Therefore, our network should also minimize the following objective function:
\begin{equation}
\label{eq:kl_loss}
\mathcal{L}_{div} = \mathcal{D}(P_{\sigma}(d)) \|Q_{\epsilon}(d)) = 
\int_{0}^{d_{max}} P_{\sigma}(d) \log \frac{P_{\sigma}(d)}{Q_{\epsilon}(d)} \,dd
\end{equation}
where $d$ spans the disparity range. %
Since the network regresses disparity, the continuous formulation of KL divergence is appropriate.

Minimizing Eq. (\ref{eq:kl_loss}) directly requires closed form expressions for the two distributions, which are not available to us. They could be modeled as Laplace distributions, but the maximum likelihood estimator is not differentiable. Moreover, fitting models to the data may be imprecise at the tails of the distributions. Therefore we choose non-parametric representations in the form of histograms. 

Histogramming is also not a differentiable operation, leading us to \textit{soft-histogramming} as a differentiable alternative. We specify a set of bins for the histograms based on the statistics of the errors $\epsilon$, since their distribution is the one that should be matched by the distribution of uncertainty estimates $\sigma$. Since the L1 loss, and our network in general, does not discriminate between positive and negative errors, we work with absolute values of $\epsilon$ and $\sigma$. 

For each batch during training, we compute the mean and standard deviation of the error, $\mu_\epsilon$ and $b_\epsilon$, set $C_0=\mu_\epsilon$ as the center of the first bin, and $C_m=\mu_\epsilon+\alpha_{m}b_\epsilon$ as the center of the last bin. (We use $b_\epsilon$ for the standard deviation to avoid overloading $\sigma$ or $s$. Also note that the last bin extends to the disparity range, which is also the maximum possible error.) We then define $m-1$ centers evenly spaced in a linear or logarithmic scale between the first and last center.

Given the bin centers, we compute a soft-histogram for the errors and one for the uncertainties as follows, considering all pixels with ground truth disparity, and error values. We present the steps for $\epsilon$ here. 
For each error $\epsilon^{(i)}$, we compute weights for every bin center which are inversely proportional to the distance. 
\begin{equation}
    w_{j}(\epsilon^{(i)}) = \lambda_1 \cdot \exp(-\frac{(\mu_{\epsilon} + \alpha_j b_{\epsilon} - \epsilon^{(i)})^2}{\lambda_2})
\end{equation}
where $\lambda_1$ and $\lambda_2$ are hyper-parameters.
Softmax is then applied to favor the nearest bins and the contributions of $n$ pixels are accumulated in the bins of the histogram $H_{\epsilon}$.
\begin{align}
    H_{\epsilon}(j) & = \frac{1}{n} \sum^n_{i=1}\frac{\exp ( w_{j}(\epsilon^{(i)}))}{\sum^m_{j=0}\exp ( w_{j}(\epsilon^{(i)}))}, \, \,  j \in [0, m]
\end{align}
The histogram for $\sigma$, $H_{\sigma}$, is obtained similarly. (Note that the bins of both histograms are defined in terms of $b_\epsilon$.)

The loss representing the discrete form of the KL divergence between the two histograms is given by:
\begin{equation}
\label{eq:kl_loss_dis}
\mathcal{L}_{div} = \sum_{j=0}^{m} H_\epsilon(j) \log \frac{H_\epsilon(j)}{H_\sigma(j)}
\end{equation}

\subsection{SEDNet}
\label{sec:sednet}

Our network architecture, named \textbf{\netname}, includes a disparity estimation subnetwork, an uncertainty estimation subnetwork, and is shown in Figure~\ref{fig:Pipline}. In all experiments, we have adopted GwcNet \cite{guo2019group} as the disparity estimation subnetwork, among other options, and we have designed a novel uncertainty estimation subnetwork that interfaces with GwcNet. It is worth noting that the uncertainty subnetwork is \textit{extremely small.}

The GwcNet subnetwork extracts features from the images using a ResNet-like feature extractor, generates the cost volume, and assigns disparities to pixels using the soft-argmax operator \cite{kendall2017-gcnet}. 
The output module of the disparity predictor generates $K$ disparity maps at different resolutions. 

We propose a new uncertainty estimator integrated with the stereo matching network. The uncertainty estimation subnetwork learns to predict the log of the observation noise scalar, the error, at each pixel. The proposed subnetwork takes the multi-resolution disparity predictions as input, computes the 
\textit{pairwise differences vector (PDV)}
and passes it to a pixel-wise MLP to regress the uncertainty maps. Specifically, the disparity estimator outputs $K$ disparity maps at different resolutions $\hat{\mathbf{d}}=\{ \Hat{\mathbf{d}}_{1},...,\hat{\mathbf{d}}_{K}\}$, which are first upsampled to full-resolution and then undergo  pairwise differencing to form the PVD, which consists of 
$K \choose 2$ elements.
The output set of uncertainty maps $\mathbf{S}$ also contains $K$ resolutions to match the disparity maps. We use $K = 4$ in all experiments.

\subsection{Loss Function}\label{sec:total_loss}

Our loss function combines two parts: (1) the log-likelihood loss to optimize the error and uncertainty, (2) the KL divergence loss to match the distribution of uncertainty with the error. The total loss considers all disparity and uncertainty maps upsampled to the highest 
resolution:
\begin{equation}
    \mathcal{L} = \sum^K_{k=1} c_{k} \cdot (\mathcal{L}_{log,k} + \mathcal{L}_{div,k})
\end{equation}
where $c_{k}$ denotes the coefficients for the $k^{th}$ resolution level, $\mathcal{L}_{log,k}$ and $\mathcal{L}_{div,k}$ are computed by Eq.~(\ref{eq:log_loss}) and Eq.~(\ref{eq:kl_loss_dis}) on the prediction of the corresponding resolution level.

\section{Experiment Results}
\label{sec:experiments}
\pdfoutput=1

\pdfoutput=1

\begin{table*}[htb]
\begin{adjustbox}{width=\textwidth,center}
    \centering
    \begin{tabular}{|c|c|cccccc|cc|cc|cc|cc|}
    \hline
\multirow{2}{*}{\textbf{Dataset}} &
  \multirow{2}{*}{\textbf{Method}} &
  \multicolumn{6}{c|}{\textbf{Loss}} &
  \multicolumn{2}{c|}{\textbf{Inliers}} &
  \multicolumn{2}{c|}{\textbf{Disparity}$\downarrow$} &
  \multicolumn{2}{c|}{\textbf{APE}$\downarrow$} &
  \multicolumn{2}{c|}{\textbf{AUC}$\downarrow$} \\
  \cline{3-16}
 &
   &
  BCE &
  L1 &
  Log &
  KL &
  Bins &
  Scale &
  Def. &
  Pct(\%) &
  EPE &
  D1(\%) &
  Avg. &
  Median &
  Opt. &
  Est. \\
  \hline
  \hline
\multirow{4}{*}{Scene Flow} &
  GwcNet &
  - &
  $\surd$ &
  - &
  - &
  - &
  - &
  - &
  - &
  0.7758 &
  4.127 &
  - &
  - &
  10.9291 &
  - \\
   &
  +LAF &
  $\surd$ &
  - &
  - &
  - &
  - &
  - &
  - &
  - &
  0.7758 &
  4.127 &
  - &
  - &
10.9291 &
  20.0813 \\
 &
  +$\mathcal{L}_{log}$ &
  - &
  $\surd$ &
  $\surd$ &
  - &
  - &
  - &
  EPE\textless{}5 &
  96.96 &
  0.7611 &
  4.131 &
  0.6999 &
  0.0728 &
  5.7449 &
  12.1121 \\
 &
  +SEDNet &
  - &
  $\surd$ &
  $\surd$ &
  $\surd$ &
  11 &
  log &
  EPE\textless{}$\mu_\epsilon$+3$b_\epsilon$ &
  98.42 &
  \best{0.6754} &
  \best{3.963} &
  \best{0.5797} &
  \best{0.0432} &
  \best{4.9134} &
  \best{8.7195} \\
  \hline
  \hline
\multirow{4}{*}{VK2-S6} &
  GwcNet &
  - &
  $\surd$ &
  - &
  - &
  - &
  - &
  - &
  - &
  0.4125 &
  1.763 &
  - &
  - &
  6.0962 &
  - \\
 &
  +$\mathcal{L}_{log}$ &
  - &
  $\surd$ &
  $\surd$ &
  - &
  - &
  - &
  EPE\textless{}5 &
  98.86 &
  0.3899 &
  1.584 &
  0.4136 &
  0.1753 &
  4.6872 &
  12.5320 \\
 &
  +SEDNet &
  - &
  $\surd$ &
  $\surd$ &
  $\surd$ &
  11 &
  log &
  EPE\textless{}$\mu_\epsilon$+3$b_\epsilon$ &
  99.24 &
  \best{0.3109} &
  \best{1.392} &
  0.5234 &
  0.1454 &
  \best{4.1726} &
  \best{9.7637} \\
   &
  +SEDNet &
  - &
  $\surd$ &
  $\surd$ &
  $\surd$ &
  11 &
  log &
  EPE\textless{}$\mu_\epsilon$+5$b_\epsilon$ &
  99.68 &
  0.3236 &
  1.427 &
  \best{0.3561} &
  \best{0.1096} &
  4.2767 &
  9.9843 \\
  \hline
  \hline
  \multirow{4}{*}{VK2-S6-Moving} &
  GwcNet &
  - &
   $\surd$ &
  - &
  - &
  - &
  - &
  - &
  - &
  0.4253 &
  1.689 &
  - &
  - &
  5.9184 &
  - \\
 &
  +$\mathcal{L}_{log}$ &
  - &
   $\surd$ &
   $\surd$ &
  - &
  - &
  - &
  EPE\textless{}5 &
  98.91 &
  0.4231 &
  1.537 &
  0.4575 &
  0.1890 &
  4.3663 &
  11.3532 \\
 &
  +SEDNet &
  - &
   $\surd$ &
   $\surd$ &
   $\surd$ &
  11 &
  log &
  EPE\textless{}$\mu_\epsilon$+3$b_\epsilon$ &
  99.62 &
  \best{0.3577} &
  \best{1.389} &
  0.5958 &
  0.1573 &
  \best{3.9012} &
  \best{8.8339} \\
   &
  +SEDNet &
  - &
   $\surd$ &
   $\surd$ &
   $\surd$ &
  11 &
  log &
  EPE\textless{}$\mu_\epsilon$+5$b_\epsilon$ &
  99.76 &
  0.3862 &
  1.420 &
  \best{ 0.4002} &
   \best{0.1164} &
  4.0423 &
  9.0631 \\
  \hline
  \hline
\multirow{2}{*}{DrivingStereo} &
  +$\mathcal{L}_{log}$(FT) &
  - &
  $\surd$ &
  $\surd$ &
  - &
  - &
  - &
  - &
  - &
  0.5332 &
  0.2641 &
  0.3449 &
  0.2297 &
  21.7002 &
  45.7096 \\
 &
  +SEDNet(FT) &
  - &
  $\surd$ &
  $\surd$ &
  $\surd$ &
  11 &
  log &
  EPE\textless{}$\mu_\epsilon$+5$b_\epsilon$ &
  99.86 &
  \best{0.5264} &
  \best{0.2439} &
  \best{0.3324} &
  \best{0.2267} &
  \best{21.2856} &
  \best{44.3297} \\
  \hline
  \hline
  \multirow{4}{*}{DS-Weather} &
  GwcNet &
  - &
  $\surd$ &
  - &
  - &
  - &
  - &
  - &
  - &
  1.6962 &
  8.313 &
  - &
  - &
  44.4896 &
  - \\
 &
  +$\mathcal{L}_{log}$ &
  - &
  $\surd$ &
  $\surd$ &
  - &
  - &
  - &
  EPE\textless{}5 &
  95.78 &
  2.3944 &
  6.666 &
  2.1443  &
  \best{0.4383} &
  41.1909 &
  95.4264 \\
 &
  +SEDNet &
  - &
  $\surd$ &
  $\surd$ &
  $\surd$ &
  11 &
  log &
  EPE\textless{}$\mu_\epsilon$+3$b_\epsilon$ &
  98.95 &
  \best{1.5637} &
  6.508 &
  2.3406 &
  0.5309 &
  \best{38.4871} &
  \best{86.1118} \\
 &
  +SEDNet &
  - &
  $\surd$ &
  $\surd$ &
  $\surd$ &
  11 &
  log &
  EPE\textless{}$\mu_\epsilon$+5$b_\epsilon$ &
  99.41 &
  1.7051 &
  \best{6.057} &
  \best{1.5842} &
  0.6104 &
  39.8057 &
  87.1882 \\
  \hline
\end{tabular}
    \end{adjustbox}
    \caption{Quantitative results: (1) \textit{within-domain} on SceneFlow, VK2-S6 and VK2-S6-Moving; (2) after finetuning (FT) on DrivingStereo; (3) \textit{cross-domain} on DS-Weather. The best results in each category in each experiment are in bold typeface. \netname outperforms the baselines with respect to disparity and uncertainty metrics in the majority of experiments. }
    \label{tab:in_domian}
\end{table*}

\newcommand\hcwdf{0.65\columnwidth}
\begin{figure*}[thb]
\begin{adjustbox}{width=\textwidth,center}
\centering
\begin{tabular}{ccc}
\includegraphics[width=\hcwdf]{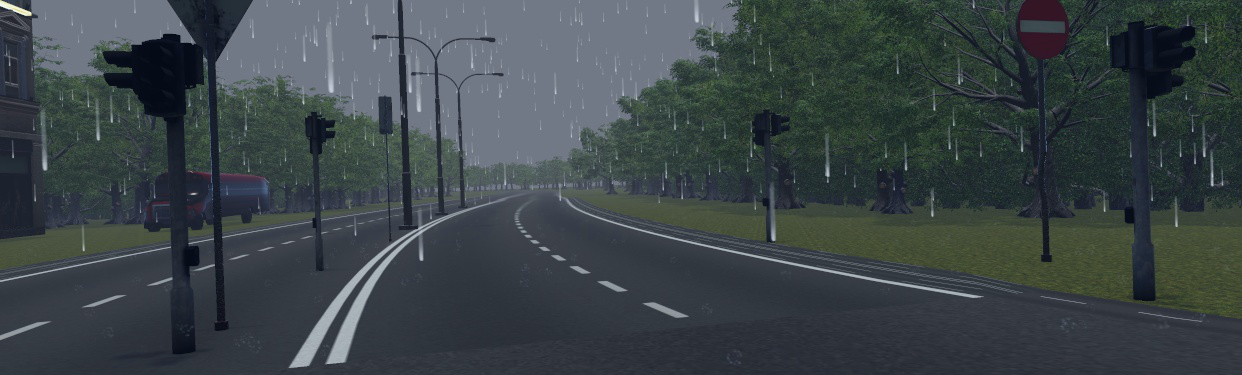} &
\includegraphics[width=\hcwdf]{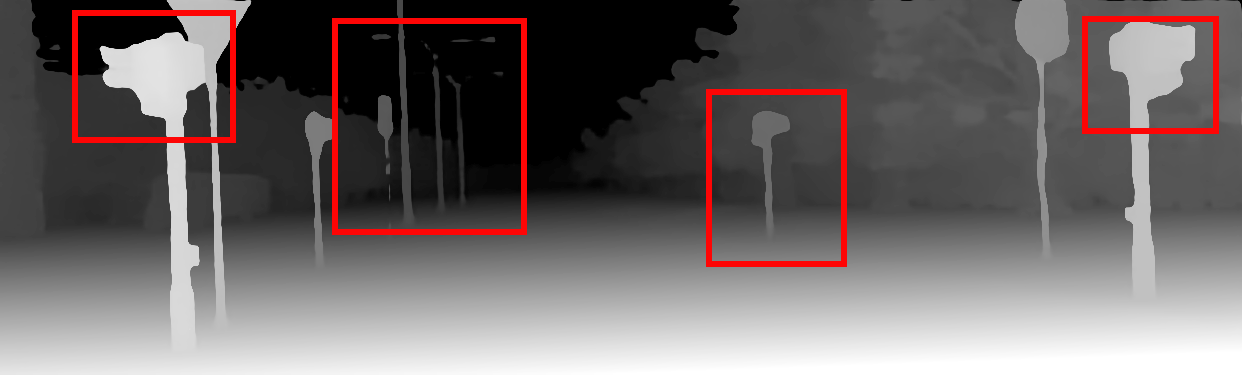} &
\includegraphics[width=\hcwdf]{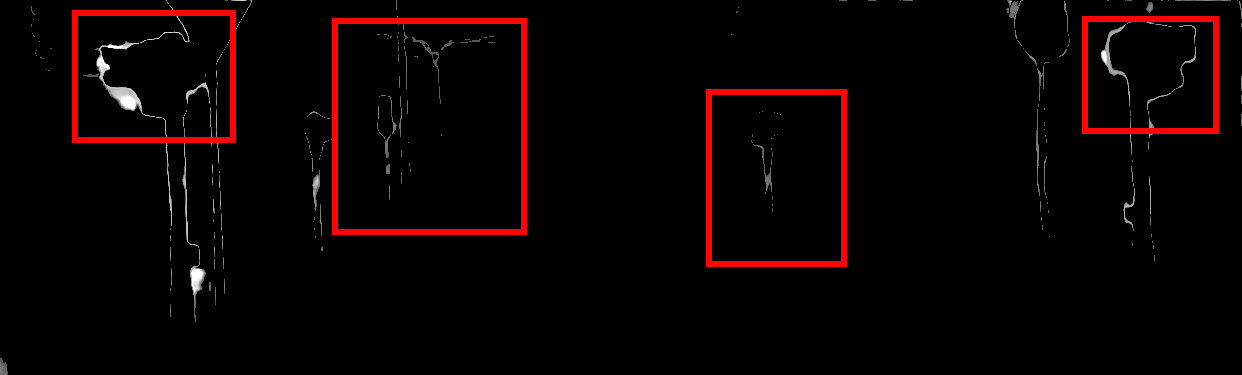} \\
Left Image & \kg Disparity & \kg Uncertainty \\

\includegraphics[width=\hcwdf]{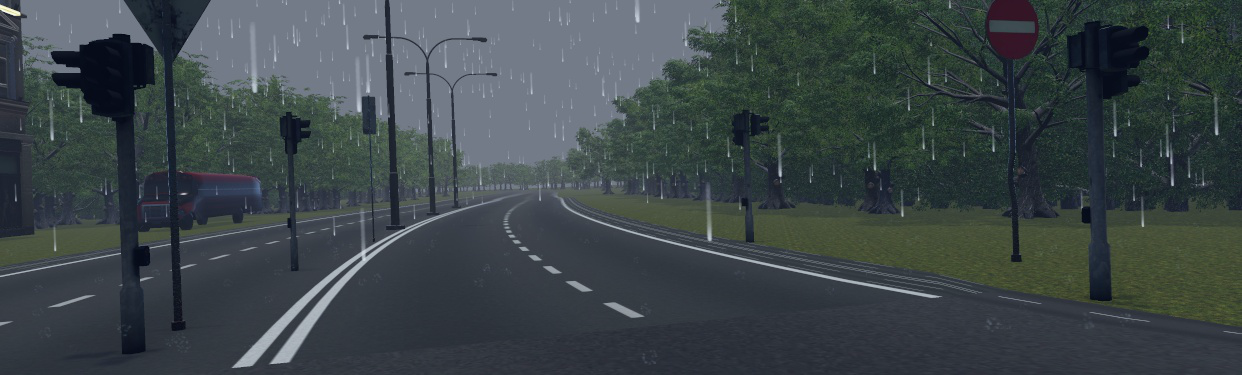} &
\includegraphics[width=\hcwdf]{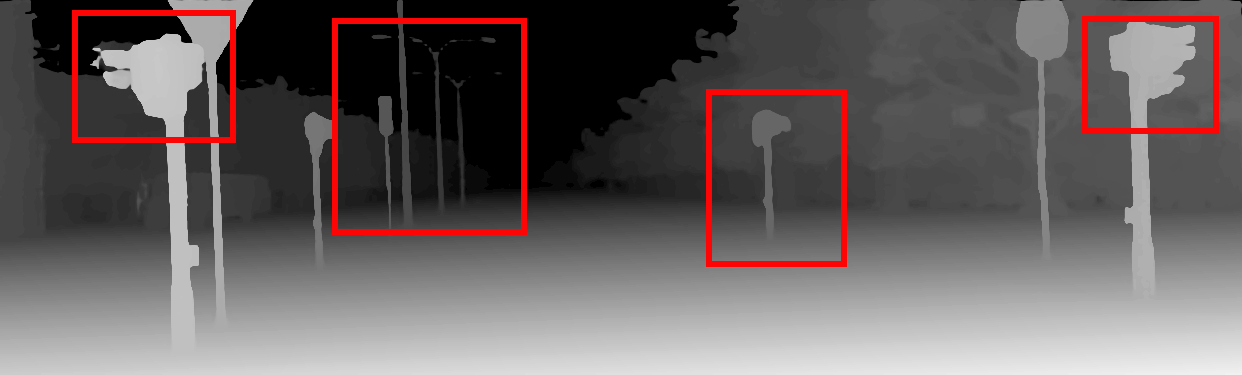} &
\includegraphics[width=\hcwdf]{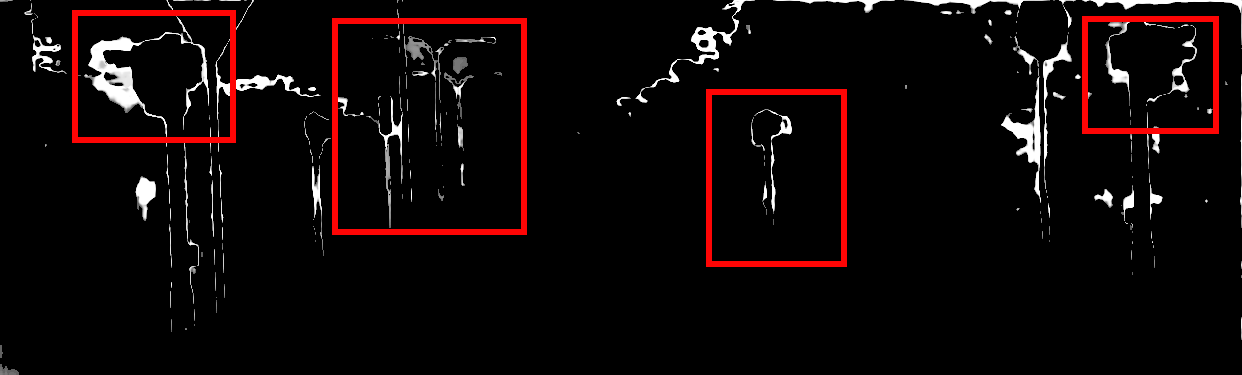}\\
Right Image & \netname Disparity & \netname Uncertainty \\

\end{tabular}
\end{adjustbox}
\vspace{-4pt}
\caption{Example from VK2-S6. In a rainy scene with poor illumination, recognizing objects far away from the camera is even difficult for human observers. \kg fails to predict the disparity of objects, such as the traffic sign near the left street light. On the other hand, \netname accurately predicts the disparity of these challenging objects, while its uncertainty map also captures more information, such as the  trees in the background, the traffic light near the trees and the street lights.}
\label{fig:vk_2}
\end{figure*}

In this section, we present our experimental setup and results on \textit{within-domain} and \textit{cross-domain} experiments. Datasets, evaluation metrics and baselines are described in Section~\ref{sec:dataset}, implementation details in Section~\ref{sec:imp_details},
and experimental results in Sections~\ref{sec:results} and~\ref{sec:match_error}. Additionally, a synthetic-to-real transfer evaluation is presented
in Section~\ref{sec:syn_to_real}.
We provide more quantitative and qualitative results, extending Section~\ref{sec:experiments}, in the Supplement, which also includes ablation studies on several aspects of \netname and the baselines,
as well as disparity, error, and uncertainty maps of difficult examples.

\subsection{Datasets and evaluation metrics}
\label{sec:dataset}

\noindent\textbf{SceneFlow}~\cite{mayer2016large} is a collection of three synthetic stereo datasets: FlyingThings3D, Driving, Monkaa. The datasets provide $35,454$ training and $4,370$ test stereo pairs in $960\times540$ pixel resolution with dense ground-truth disparity maps. We use the \textit{finalpass} versions of the rendered images which are more realistic because of the motion blur and depth of field effect.

\noindent\textbf{Virtual KITTI 2} \textit{(VK2)}~\cite{gaidon2016virtual} is a synthetic clone of KITTI~\cite{Geiger2012CVPR}. It consists of $21,260$ synthetic $1,242 \times 375$ stereo pairs from 6 driving scenes with 10 different imaging and weather conditions. 
Scene 006 \textit{(VK2-S6)} is specified as the test set by the authors of the dataset.
Since the car with the cameras is stopped for a long part of Scene 006 and only other cars move in the images, we split the last part of the scene where the car moves and denote it as \textit{VK2-S6-Moving}. We report results separately on this subset.
Since results on VK2-S6-Moving do not suffer from bias due to the almost constant background of the first part of the scene, evaluation on VK2-S6-Moving is more informative and fair.

\noindent\textbf{DrivingStereo}~\cite{yang2019drivingstereo} is a large real-world autonomous driving dataset. It contains $174,437$ training and $7,751$ test stereo pairs at $881\times400$ pixel resolution. The dataset provides sparse ground truth disparity as well as a challenging subset (\textit{DS-Weather}) of $2,000$ stereo pairs in 4 different weather conditions. 

For all datasets above, we exclude pixels with disparities $d > 192$ in training.

\noindent\textbf{Metrics.} To evaluate disparity estimation, we compute the endpoint error (\textbf{EPE}) and the percentage of outliers (\textbf{D1}) (i.e., the percentage of pixels with EPE $> 3$px or $\geq 5\%$ of the true depth). 

To evaluate uncertainty estimation, we use density-EPE ROC curves and the area under the curve (\textbf{AUC}) \cite{poggi2021confidence,hu2012quantitative,Kim_2019_CVPR}. The ROC curves in our case measure EPE (not binary correctness) at increasing disparity map density by successively adding pixels in increasing order of uncertainty. 
The optimal AUC is obtained by adding pixels in order of increasing EPE
and is therefore the lowest value any algorithm could achieve given the set of EPE values, while the estimated AUC is sorted by predicted uncertainty. To evaluate the precision of  uncertainty estimation, we also introduce the absolute prediction error (\textbf{APE}), which is the average L1 distance between the error and the observation noise scalar, $\sigma$. We report the average and median APE over all pixels. 

\noindent\textbf{Baselines.}  
Since we use GwcNet \cite{guo2019group} as the disparity estimation subnetwork of \netname, we
compare \netname with three baselines: (1) the original GwcNet trained with smooth L1 loss. (2) LAF-Net \cite{Kim_2019_CVPR} trained under the BCE loss on the left RGB images, the cost volumes and predicted disparity maps of GwcNet at the highest resolution. (3) {\netname} but only trained with the log-likelihood loss, therefore similar to Kendall and Gal's model \cite{kendall2017uncertainties}. We use \kg in tables and figures for this baseline. We selected GwcNet because it is the backbone of \netname, and LAF-Net as the confidence estimation baseline due to its strong performance within the training domain according to \cite{poggi2021confidence}. 

\subsection{Implementation Details}
\label{sec:imp_details}

We implemented all networks in PyTorch and used the Adam optimizer~\cite{kingmaadam} with $\beta_1 = 0.9$ and $\beta_2 = 0.999$ for all experiments. Training of all models was stopped before overfitting occurred.

Experiments on the VK2 dataset were performed on two NVIDIA RTX A6000 GPUs, each with 48 GB of RAM. For this dataset, we trained all models from scratch with an initial learning rate of 0.0001, down-scaled by 5 every 10 epochs. During training, we randomly cropped $512\times256$ patches from the images. During testing, we evaluated at the full resolution of VK2.

Experiments on the DrivingStereo dataset (DS) were also performed on two NVIDIA RTX A6000 GPUs. For this dataset, we did two experiments using the models pre-trained on VK2: (1) we finetuned on the DS training set with a learning rate starting from 0.0001, down-scaled by 2 every 3 epochs after epoch 10, then performed in-domain evaluation on the DS test set; (2) we skipped the finetuning step and performed cross-domain evaluation on DS-Weather subset. During training, we randomly cropped the inputs to be the same size as in the VK2 experiments. During testing, we padded the test samples to be the same resolution as VK2.

\begin{table}[b]
\vspace{-12pt}
    \centering
    \begin{tabular}{|c|c|c|}
    \hline
    \textbf{Architecture} & \textbf{Params} & \textbf{MACs(G)} \\
    \hline
    \hline
    GwcNet &  6,909,728 & 1075.82 \\
    \hline
    \netname & 6,909,918 &  1075.91 \\
    \hline
\end{tabular}
\vspace{-4pt}
    \caption{Comparison on number of parameters and computational complexity. MAC stands for multiply–accumulate operations. \netname only adds a 3-layer MLP, with 190 parameters, as an uncertainty decoder to GwcNet.}
    \label{tab:parames}
\end{table}

Experiments on the SceneFlow dataset were performed on an Nvidia TITAN RTX GPU with 24 GB memory. We trained all models from scratch on $256\times128$ patches cropped from half-resolution images to limit memory consumption. We set the initial learning rate to 0.001 and down-scaled it by 2 every 2 epochs after epoch 10.

For all the experiments above, we applied an inlier filtering strategy during training, which only back-propagates from the inliers. See the last paragraph of Section~\ref{sec:results} for details. Table~\ref{tab:parames} shows the number of parameters and computational complexity of GwcNet and \netname.

\subsection{Qualitative and Quantitative Results}
\label{sec:results}

In Table~\ref{tab:in_domian}, we present results: (1) \textit{within-domain} on SceneFlow and VK2; (2) on the DrivingStereo test set after finetuning the VK2 models on the DrivingStereo training set; (3) \textit{cross-domain} on the DS-Weather challenge test set by directly applying the model trained on VK2 without finetuning. An extended version of this table can be found in Table \ref{tab:in_domian_big} in the Supplement. 

\noindent\textbf{Disparity Estimation.} In all experiments, {\netname} achieves lower errors than all the baselines. See the EPE and D1 columns in Table~\ref{tab:in_domian}. %
Even in extreme weather like fog and rain, {\netname} predicts good disparity unaffected by poor illumination and blur. See Figure~\ref{fig:vk_2}.

\begin{figure*}[thb]
\begin{adjustbox}{width=\textwidth,center}
\centering
\begin{tabular}{ccc}
\includegraphics[width=0.65\columnwidth]{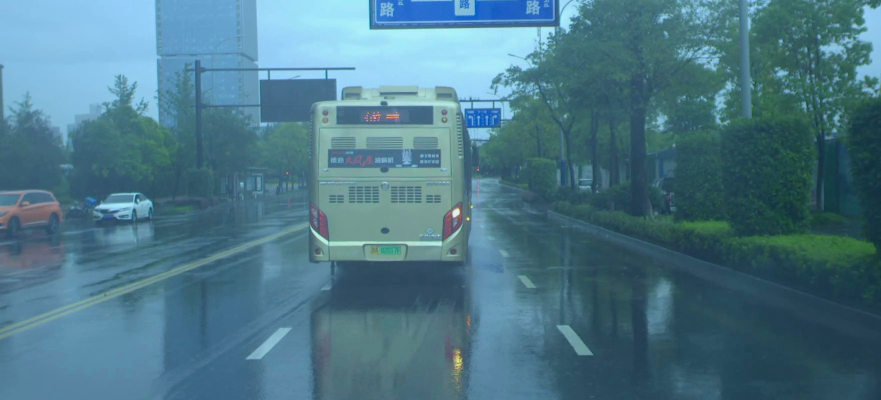} &
\includegraphics[width=0.65\columnwidth]{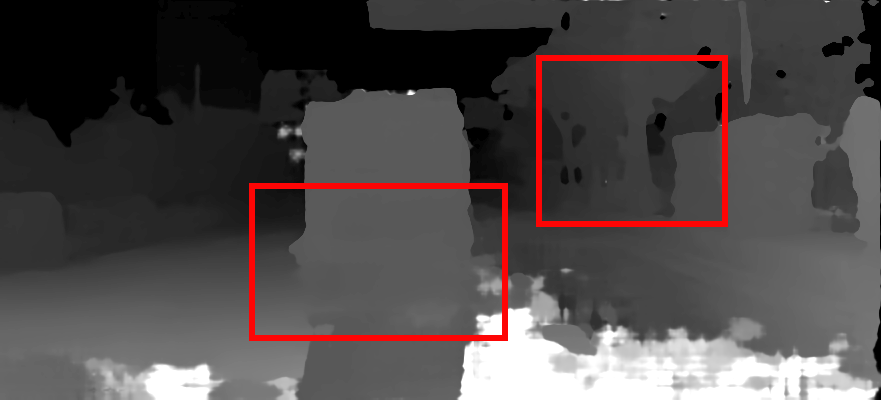} &
\includegraphics[width=0.65\columnwidth]{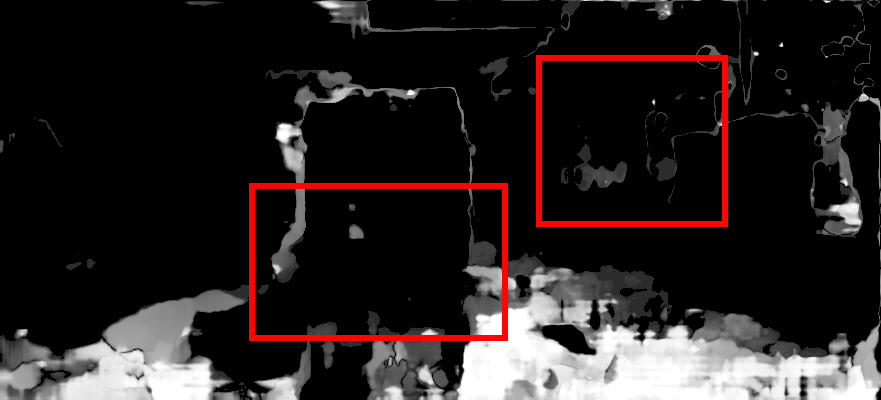} \\
Left Image & \kg Disparity & \kg Uncertainty \\
\includegraphics[width=0.65\columnwidth]{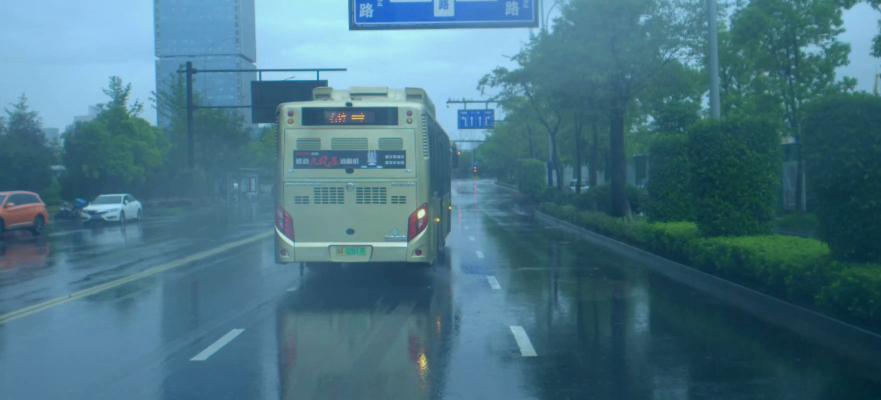} &
\includegraphics[width=0.65\columnwidth]{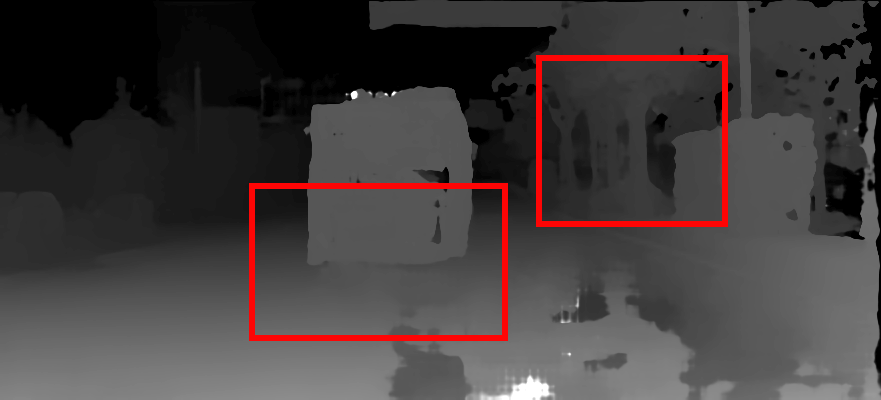} &
\includegraphics[width=0.65\columnwidth]{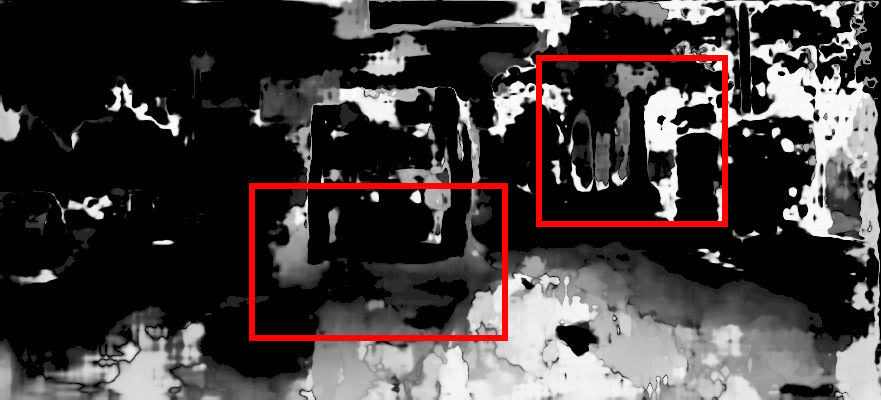}\\
 Right Image & \netname Disparity  & \netname Uncertainty \\

\end{tabular}
\end{adjustbox}
\vspace{-4pt}
\caption{Example from DS-Weather. Unlike the synthetic data, the rainy-day real images do not only suffer from poor illumination, but also pose challenges due to reflections in the 
water. In this example, the road is like a mirror, misleading the \kg model. Recall that the LIDAR ground truth disparity is very sparse, and is even sparser in reflective regions. The disparity map of \kg fails to distinguish the car and the reflection, but \netname is able to estimate the correct disparity and the uncertainty of the car. }
\label{fig:ds_2}
\end{figure*}

\begin{figure}[thb]
\centering
\begin{tabular}{c}
\includegraphics[width=0.7\columnwidth]{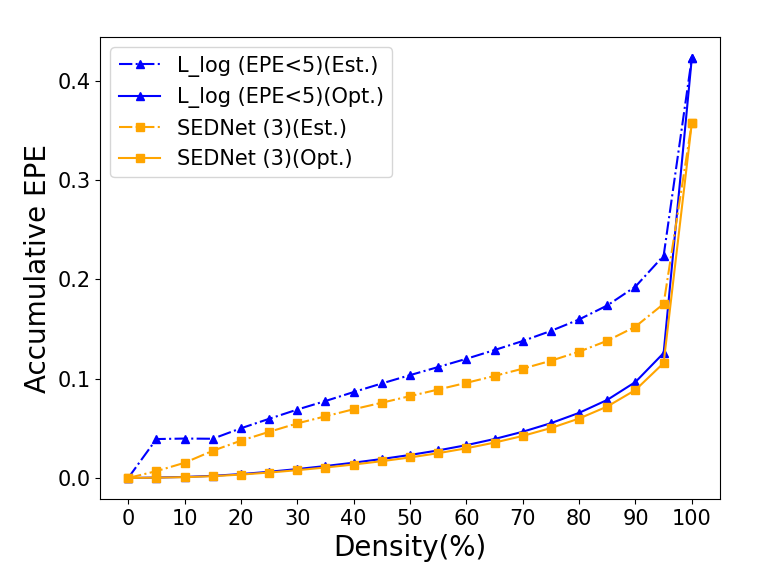} \\
\end{tabular}
\vspace{-4pt}
\caption{Comparison of density-EPE curves on VK2-S6-Moving. ''L\_log (EPE\textless{}5)'' and ''\netname (3)'' represent the \kg (with fixed inliers) and the \netname (with adaptive inliers of EPE\textless{}$\mu_\epsilon$+3$b_\epsilon$) of the \textit{VK2-S6-Moving} experiment in Table~\ref{tab:in_domian}. While their corresponding optimal AUCs are almost equal, the advantage of \netname in estimated AUC is significant. }
\label{fig:auc}
\end{figure}

\noindent\textbf{Uncertainty Estimation.} Our method outperforms the baselines in all experiments according to the AUC metric, as shown in the last two columns in Table~\ref{tab:in_domian}. Compared to \kg, \netname decreases the estimated AUC by $20\%$ -- $30\%$ in the in-domain experiments, with a $10\%$ decrease in optimal AUC, which depends on EPE. 
In the cross-domain evaluation, the advantage of {\netname} is even more evident.
Figure~\ref{fig:ds_2} shows uncertainty maps 
for real data, on which our method captures details more faithfully.
The ROC curves of the best \kg and \netname models based on EPE on VK2-S6-Moving are presented in Figure~\ref{fig:auc}.  

\noindent\textbf{Effects of Back-propagation from Inliers.} 
We apply two kinds of inlier filters: one with a fixed threshold that excludes all pixels that have an EPE larger than 5 from back-propagation; and one with adaptive threshold which excludes pixels that have EPE greater than a specified number of $b_\epsilon$ from the mean error. Back-propagation from the inliers only helps the network improve its performance on both disparity and uncertainty estimation. 
We attribute this to the suppression of harmful outliers that give rise to large gradients. 
Quantitative results for the baselines and the proposed method with different inlier settings are reported in the Supplement.
The results show that using adaptive thresholds is better than fixed thresholds. Fixed thresholds exclude more pixels, especially at lower resolutions and in the early stages of training, preventing the network from learning how to correct them.

\begin{figure}[b]
\begin{adjustbox}{width=\columnwidth,center}
\centering
\begin{tabular}{cc}
\includegraphics[width=0.5\columnwidth]{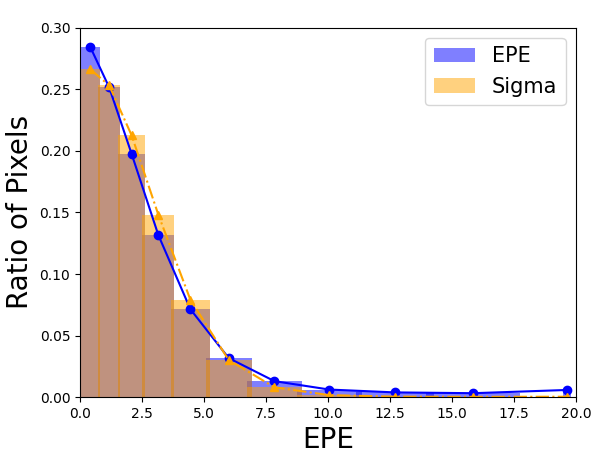} &
\includegraphics[width=0.5\columnwidth]{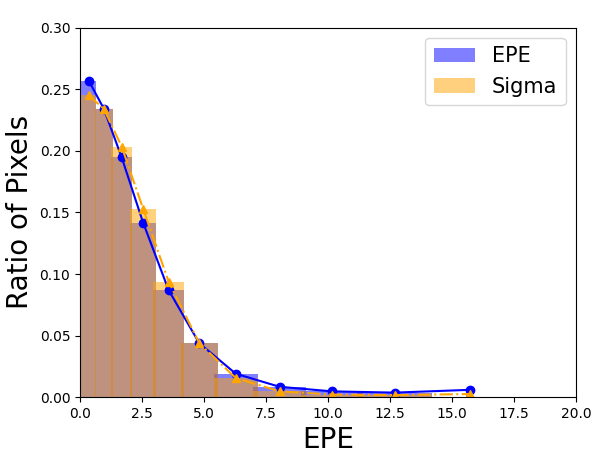} 
\\
\kg & \netname
\end{tabular}
\end{adjustbox}
\vspace{-4pt}
\caption{Distribution of error and predicted uncertainty. We pick the best model of \kg and \netname in \textit{VK2-S6-Moving}, then randomly sample $5,000,000$ points from the outputs of the two models. The distributions of error and $\sigma$ of \netname are much closer than those of the \kg model, especially in the first 5 bins that contain more than 95\% of the samples. The shorter tail of \netname also indicates fewer gross errors.}
\label{fig:s_and_error_distribution}
\end{figure}

\pdfoutput=1

\begin{table*}[htb]
\begin{adjustbox}{width=\textwidth,center}
    \centering
    \begin{tabular}{|c|c|cccccc|cc|cc|cc|cc|}
    \hline
\multirow{2}{*}{\textbf{Dataset}} &
  \multirow{2}{*}{\textbf{Method}} &
  \multicolumn{6}{c|}{\textbf{Loss}} &
  \multicolumn{2}{c|}{\textbf{Inliers}} &
  \multicolumn{2}{c|}{\textbf{Disparity}$\downarrow$} &
  \multicolumn{2}{c|}{\textbf{APE}$\downarrow$} &
  \multicolumn{2}{c|}{\textbf{AUC}$\downarrow$} \\
  \cline{3-16}
 &
   &
  BCE &
  L1 &
  Log &
  KL &
  Bins &
  Scale &
  Def. &
  Pct(\%) &
  EPE &
  D1(\%) &
  Avg. &
  Median &
  Opt. &
  Est. \\
  \hline
  \hline
  \multirow{3}{*}{VK2-S6-Morning} &
  GwcNet &
  - &
   $\surd$ &
  - &
  - &
  - &
  - &
  - &
  - &
  0.4642 &
  1.740 &
  - &
  - &
  6.1845 &
  - \\
 &
  +$\mathcal{L}_{log}$ &
  - &
   $\surd$ &
   $\surd$ &
  - &
  - &
  - &
  EPE\textless{}5 & 98.82
   & 0.4774
   & 1.624
   & \best{0.5067}
   & 0.1872
   & 4.6698
   & 12.5192
   \\
 &
  +SEDNet &
  - &
   $\surd$ &
   $\surd$ &
   $\surd$ &
  11 &
  log &
  EPE\textless{}$\mu_\epsilon$+3$b_\epsilon$ & 99.62
   & \best{0.4003}
  & \best{1.442}
  & 0.6183
   & \best{0.1553}
   & \best{4.1847}
   & \best{9.4063}
   \\
   \hline
    \multirow{3}{*}{VK2-S6-Sunset} &
  GwcNet &
  - &
   $\surd$ &
  - &
  - &
  - &
  - &
  - &
  - &
  0.4810 & 
  1.825 &
  - &
  - &
  6.6907 &
  \\
 &
  +$\mathcal{L}_{log}$ &
  - &
   $\surd$ &
   $\surd$ &
  - &
  - &
  - &
  EPE\textless{}5 & 98.84
   & 0.4863
   & 1.627
   & \best{0.5060}
   & 0.1827
   & 5.0075
   & 13.7848
 \\
 &
  +SEDNet &
  - &
   $\surd$ &
   $\surd$ &
   $\surd$ &
  11 &
  log &
  EPE\textless{}$\mu_\epsilon$+3$b_\epsilon$ & 99.61
   &\best{0.4108}
  &\best{1.475}
  & 0.6189
   & \best{0.1509}
   &\best{4.5840}
   &\best{10.7946}
   \\
   \hline
     \multirow{3}{*}{VK2-S6-Fog} &
  GwcNet &
  - &
   $\surd$ &
  - &
  - &
  - &
  - &
  - &
  - &
  0.4660 &
  1.812 &
  - &
  - &
  6.8355 &
  - \\
 &
  +$\mathcal{L}_{log}$ &
  - &
   $\surd$ &
   $\surd$ &
  - &
  - &
  - &
  EPE\textless{}5 & 98.98
   & 0.4425
   & 1.448
   &  \best{0.4609}
   & 0.1865
   & 4.8983
   & 12.1305
   \\
 &
  +SEDNet &
  - &
   $\surd$ &
   $\surd$ &
   $\surd$ &
  11 &
  log &
  EPE\textless{}$\mu_\epsilon$+3$b_\epsilon$ & 99.71
   &\best{0.3731}
  & \best{1.288}
  & 0.5517
   & \best{0.1547}
   & \best{4.4200}
   & \best{9.9380}
   \\
  \hline
    \multirow{3}{*}{VK2-S6-Rain} &
  GwcNet &
  - &
   $\surd$ &
  - &
  - &
  - &
  - &
  - &
  - &
  0.4618 &
  1.700 &
  - &
  - &
 6.6774  &
  - \\
 &
  +$\mathcal{L}_{log}$ &
  - &
   $\surd$ &
   $\surd$ &
  - &
  - &
  - &
  EPE\textless{}5 & 98.88
   & 0.4707
   & 1.571
   & \best{0.4899}
   & 0.1861
   & 4.9351
   & 13.3214
   \\
 &
  +SEDNet &
  - &
   $\surd$ &
   $\surd$ &
   $\surd$ &
  11 &
  log &
  EPE\textless{}$\mu_\epsilon$+3$b_\epsilon$ & 99.69
   &\best{0.3873}
  &\best{1.356}
  & 0.6685
   &  \best{0.1537}
   & \best{4.4013}
   &\best{10.3362}
   \\
   \hline
   \hline
     \multirow{3}{*}{DS-Cloudy} &
  GwcNet &
  - &
   $\surd$ &
  - &
  - &
  - &
  - &
  - &
  - &
   1.3413 &
  5.229 &
  - &
  - &
  37.4263 &
  - \\
 &
  +$\mathcal{L}_{log}$ &
  - &
   $\surd$ &
   $\surd$ &
  - &
  - &
  - &
  EPE\textless{}5 & 97.48
   & 1.4780 
   & \best{3.948}
   & 1.2617
   & \best{0.3513}
   & 34.4488
   & 82.5380
   \\
 &
  +SEDNet &
  - &
   $\surd$ &
   $\surd$ &
   $\surd$ &
  11 &
  log &
  EPE\textless{}$\mu_\epsilon$+3$b_\epsilon$ & 98.83
   & \best{1.3183}
  & 4.414
  & 1.5260
   & 0.4021 
   & \best{33.9037}
   & \best{73.6330}
   \\
   \hline
     \multirow{3}{*}{DS-Sunny} &
  GwcNet &
  - &
   $\surd$ &
  - &
  - &
  - &
  - &
  - &
  - &
  1.5448 &
  6.991 &
  - &
  - &
  38.7386 &
  - \\
 &
  +$\mathcal{L}_{log}$ &
  - &
   $\surd$ &
   $\surd$ &
  - &
  - &
  - &
  EPE\textless{}5 & 97.08
   & 1.4837
   & \best{4.631}
   & \best{1.2806}
   & \best{0.3835}
   & \best{35.5226}
   & 85.8715
   \\
 &
  +SEDNet &
  - &
   $\surd$ &
   $\surd$ &
   $\surd$ &
  11 &
  log &
  EPE\textless{}$\mu_\epsilon$+3$b_\epsilon$ & 98.64
   & 1.5548
  & 5.878
  & 3.0025 
   & 0.4808
   & 35.6523
   & \best{83.2573}
   \\
   \hline
     \multirow{3}{*}{DS-Foggy} &
  GwcNet &
  - &
   $\surd$ &
  - &
  - &
  - &
  - &
  - &
  - &
  1.5476 &
  8.859 &
  - &
  - &
  51.4640 &
  - \\
 &
  +$\mathcal{L}_{log}$ &
  - &
   $\surd$ &
   $\surd$ &
  - &
  - &
  - &
  EPE\textless{}5 & 94.89
   & 2.9553
   & 9.015
   &  2.6923
   & \best{0.5556}
   &48.7136
   & 101.7025
   \\
 &
  +SEDNet &
  - &
   $\surd$ &
   $\surd$ &
   $\surd$ &
  11 &
  log &
  EPE\textless{}$\mu_\epsilon$+3$b_\epsilon$ & 99.27
   & \best{1.5398}
  & 7.357
  &  \best{2.4109}
   &  0.7023
   & \best{47.7932}
   & \best{97.8627}
   \\
   \hline
     \multirow{3}{*}{DS-Rainy} &
  GwcNet &
  - &
   $\surd$ &
  - &
  - &
  - &
  - &
  - &
  - &
  3.1918 &
 17.356  &
  - &
  - &
  68.0346 &
  - \\
 &
  +$\mathcal{L}_{log}$ &
  - &
   $\surd$ &
   $\surd$ &
  - &
  - &
  - &
  EPE\textless{}5 & 98.79
   & 5.3539
   & 12.501
   & 4.9480
   & \best{0.5759}
   &  59.3952
   & 146.8906
   \\
 &
  +SEDNet &
  - &
   $\surd$ &
   $\surd$ &
   $\surd$ &
  11 &
  log &
  EPE\textless{}$\mu_\epsilon$+3$b_\epsilon$ & 99.10
   &\best{2.2165}
  & 11.020
  & \best{2.6599}
   & 0.6722
   & \best{50.8103}
   & \best{110.8360}
   \\
   \hline
\end{tabular}
    \end{adjustbox}
    \caption{Quantitative results of synthetic to real evaluation.  The top 4 subsets from VK2-S6-Moving are synthetic datasets, while the rest 4 subsets from DS-Weather are the real dataset. The best results in each category in each experiment are in bold typeface. \netname outperforms the baselines especially on the uncertainty estimation of the real data and under terrible weather (i.e., foggy and rainy).}
    \label{tab:syt_to_real}
\end{table*}

\subsection{Matching the Error Distribution.}
\label{sec:match_error}

Comparing the results of APE in all experiments in Table~\ref{tab:in_domian} and Figure~\ref{fig:s_and_error_distribution}, the difference between the errors and the predicted uncertainty is always close to the EPE, as expected. \netname, however, achieves lower APE, showing that it matches the true distribution better.

\subsection{Generalization from Synthetic to Real Data}
\label{sec:syn_to_real}
Stereo-matching networks are typically trained on synthetic data and fine-tuned on small amounts of data from the target domain due to the cost and difficulty of acquiring real data with ground truth depth.
In this section, we extend the experiments of VK2-S6 and DS-Weather in Table~\ref{tab:in_domian} to compare the generalization performance on unseen real domains of all methods trained only on synthetic data. We picked four synthetic subsets from VK2-S6, specifically \textit{Morning, Sunset, Fog and Rain}, that have similar illumination conditions, visibility level and weather with the four real subsets of DS-Weather~\cite{yang2019drivingstereo}, i.e., \textit{Cloudy, Sunny, Foggy and Rainy}. VK2-S6-Morning and VK2-S6-Sunset have similar illumination to DS-Cloudy and DS-Sunny, but the latter two are more challenging due to camera underexposure and overexposure. Even though they are acquired under the same weather, DS-Foggy and DS-Rainy are more difficult than VK2-S6-Fog and VK2-S6-Rain. This can be seen by comparing Figures~\ref{fig:vk_2} and \ref{fig:ds_2}. The synthetic examples only mimic the poor lighting conditions and challenges caused by the fog and rain, but ignore the Tyndall effect and reflections caused by the fog and stagnant water. 

As mentioned above, all models are trained only on Scenes 001, 002, 018 and 020 of VK2. Quantitative results on synthetic to real transfer are reported in Table~\ref{tab:syt_to_real}, where we only report the best model of each method. 
The top-performing variant of \netname outperforms the baselines in the majority of experiments. 
An extended version of this table can be found in Table \ref{tab:syt_to_real_big} in the Supplement.

\section{Conclusion}
\label{sec:conclusion}
\pdfoutput=1

We have presented a novel approach for joint disparity and uncertainty estimation from stereo image pairs. The key idea is a unique loss function based on the KL divergence between the distributions of disparity errors and uncertainty estimates. This is made possible by a differentiable histogramming scheme that we also introduce here. To implement our approach, we extended the GwcNet architecture to include an uncertainty estimation subnetwork with only 190 parameters.
Our experiments on multiple large datasets have demonstrated that our approach, named \netname, is effective in both disparity and uncertainty prediction. The success of our method is attributed to the novel loss function. 
\netname easily surpasses GwcNet in disparity estimation even though they have essentially the same capacity and almost identical architecture, up to the tiny uncertainty estimation subnetwork.
We are optimistic that our approach will be similarly successful in other pixel-wise regression tasks, which we plan to address in future research.

\noindent\textbf{Acknowledgment.} This research has been supported in part by the National Science Foundation under award 2024653.

{\small
\bibliographystyle{ieee_fullname}
\bibliography{ml,uncert}
}
\clearpage
\appendix
\pdfoutput=1

\renewcommand\best[1]{\color{white}\colorbox[rgb]{0.02,0.28,0.53}{#1}}
\newcommand{\second}[1]{\color{white}\colorbox[rgb]{0.53,0.8,0.9}{#1}}

\setcounter{table}{0}
\setcounter{figure}{0}

\renewcommand\thefigure{S.\arabic{figure}}  
\renewcommand\thetable{S.\arabic{table}}
\renewcommand\thesection{S.\arabic{section}}


\section*{Supplement}
In this document, we present more quantitative and qualitative results extending Section 
4 of the main paper. Ablation studies on the configuration of the histograms and on different settings of inlier filters are provided in Section~\ref{sec:ab_study}.
A detailed version of the synthetic-to-real transfer from Virtual KITTI 2 (VK2) to the DrivingStereo weather subsets (DS-Weather) is presented in Section~\ref{sec:syn_2_real}. 
Section~\ref{sec:images} includes additional qualitative results, such as disparity, error and uncertainty maps.

\section{Ablation Studies} 
\label{sec:ab_study}
\pdfoutput=1
\pdfoutput=1

\begin{table*}[p]
\begin{adjustbox}{width=\textwidth,center}
    \centering
    \begin{tabular}{|c|c|cccccc|cc|cc|cc|cc|}
    \hline
\multirow{2}{*}{\textbf{Dataset}} &
  \multirow{2}{*}{\textbf{Method}} &
  \multicolumn{6}{c|}{\textbf{Loss}} &
  \multicolumn{2}{c|}{\textbf{Inliers}} &
  \multicolumn{2}{c|}{\textbf{Disparity}$\downarrow$} &
  \multicolumn{2}{c|}{\textbf{APE}$\downarrow$} &
  \multicolumn{2}{c|}{\textbf{AUC}$\downarrow$} \\
  \cline{3-16}
 &
   &
  BCE &
  L1 &
  Log &
  KL &
  Bins &
  Scale &
  Def. &
  Pct(\%) &
  EPE &
  D1(\%) &
  Avg. &
  Median &
  Opt. &
  Est. \\
  \hline
  \hline
\multirow{12}{*}{Scene Flow} &
  GwcNet &
  - &
  $\surd$ &
  - &
  - &
  - &
  - &
  - &
  - &
  0.7758 &
  4.127 &
  - &
  - &
  10.9291 &
  - \\
 &
   GwcNet &
  - &
  $\surd$ &
  - &
  - &
  - &
  - &
  EPE\textless{}5 &
   97.09 &
  0.7799 &
   \best{3.940} &
  - &
  - &
  8.3413 &
  - \\
  &
   GwcNet &
  - &
  $\surd$ &
  - &
  - &
  - &
  log &
  EPE\textless{}$\mu_\epsilon$+3$b_\epsilon$ &
  98.41 &
  0.7981 &
  4.072 &
  - &
  - &
  9.6451 &
  - \\
   &
  +LAF &
  $\surd$ &
  - &
  - &
  - &
  - &
  - &
  - &
  - &
  0.7758 &
  4.127 &
  - &
  - &
  10.9291 &
  20.0813 \\
 &
  +$\mathcal{L}_{log}$ &
  - &
  $\surd$ &
  $\surd$ &
  - &
  - &
  - &
  - &
  - &
  0.7445 &
  4.522 &
  0.7133 &
   0.0795  &
  6.2567 &
  10.9635 \\
 &
  +$\mathcal{L}_{log}$ &
  - &
  $\surd$ &
  $\surd$ &
  - &
  - &
  - &
  EPE\textless{}5 &
  96.96 &
  0.7611 &
  4.131 &
  0.6999 &
  0.0728 &
  5.7449 &
  12.1121 \\
 &
  +$\mathcal{L}_{log}$ &
  - &
  $\surd$ &
  $\surd$ &
  - &
  - &
  log &
  EPE\textless{}$\mu_\epsilon$+3$b_\epsilon$ &
  98.33 &
   0.7890 &
  4.428 &
 0.7047 &
  0.0869 &
  6.6069 &
  12.4265 \\
  &
  +SEDNet &
  - &
  $\surd$ &
  $\surd$ &
  $\surd$ &
  - &
  - &
  EPE\textless{}5 &
  96.57 &
   1.0046 &
   4.455 &
   0.9092 &
   0.1327 &
  7.1444 &
  16.0036 \\
 &
  +SEDNet &
  - &
  $\surd$ &
  $\surd$ &
  $\surd$ &
  11 &
  lin &
  EPE\textless{}$\mu_\epsilon$+3$b_\epsilon$ &
  98.42 &
  0.6827 &
  4.022 &
  0.5877 &
  0.0450 &
  5.1258 &
  9.0113 \\
 &
  +SEDNet &
  - &
  $\surd$ &
  $\surd$ &
  $\surd$ &
  11 &
  log &
  EPE\textless{}$\mu_\epsilon$+3$b_\epsilon$ &
  98.42 &
  \best{0.6754} &
  3.963 &
  \best{0.5797} &
  0.0432 &
  \best{4.9134} &
  \best{8.7195} \\
 &
  +SEDNet &
  - &
  $\surd$ &
  $\surd$ &
  $\surd$ &
  20 &
  log &
  EPE\textless{}$\mu_\epsilon$+3$b_\epsilon$ &
  98.42 &
  0.6762 &
  3.966 &
  0.5821 &
  0.0433 &
  4.9845 &
  8.8103 \\
 &
  +SEDNet &
  - &
  $\surd$ &
  $\surd$ &
  $\surd$ &
  50 &
  log &
  EPE\textless{}$\mu_\epsilon$+3$b_\epsilon$ &
  98.42 &
  0.6894 &
  4.016 &
  0.5931 &
  \best{0.0426} &
  5.0216 &
  8.9412 \\
  \hline
  \hline
\multirow{10}{*}{VK2-S6} &
  GwcNet &
  - &
  $\surd$ &
  - &
  - &
  - &
  - &
  - &
  - &
  0.4125 &
  1.763 &
  - &
  - &
  6.0962 &
  - \\
  &
   GwcNet &
  - &
  $\surd$ &
  - &
  - &
  - &
  - &
  EPE\textless{}5 & 98.82
   & 0.4339
   & 1.634
   &
  - &
  - &6.0598
   &
  - \\
  &
   GwcNet &
  - &
  $\surd$ &
  - &
  - &
  - &
  log &
  EPE\textless{}$\mu_\epsilon$+3$b_\epsilon$ & 99.50
   &0.4597
   & 1.847
   &
  - &
  - &6.5506
   &
  - \\
 &
  +$\mathcal{L}_{log}$ &
  - &
  $\surd$ &
  $\surd$ &
  - &
  - &
  - &
  - &
  - &
  0.4360 &
  1.957 &
  0.5876 &
  0.1123 &
  4.9389 &
  10.2407 \\
 &
  +$\mathcal{L}_{log}$ &
  - &
  $\surd$ &
  $\surd$ &
  - &
  - &
  - &
  EPE\textless{}5 &
  98.86 &
  0.3899 &
  1.584 &
  0.4136 &
  0.1753 &
  4.6872 &
  12.5320 \\
  &
  +$\mathcal{L}_{log}$ &
  - &
  $\surd$ &
  $\surd$ &
  - &
  - &
  log &
  EPE\textless{}$\mu_\epsilon$+3$b_\epsilon$ &99.52
   &0.4079
   & 1.673
  &0.4549
  &0.2261
   &5.1675
   &13.3036
  \\
  &
  +SEDNet &
  - &
  $\surd$ &
  $\surd$ &
  $\surd$ &
  - &
  - &
  EPE\textless{}5 &98.59
   &0.5197
   & 1.905
   &0.5382
   &0.1779
   &5.2803
   &13.1777
   \\
 &
  +SEDNet &
  - &
  $\surd$ &
  $\surd$ &
  $\surd$ &
  11 &
  log &
  EPE\textless{}$\mu_\epsilon$+1$b_\epsilon$ &
  98.30 &
  0.3896 &
  1.582 &
  0.3876 &
  0.1330 &
  4.4569 &
  11.8449 \\
 &
  +SEDNet &
  - &
  $\surd$ &
  $\surd$ &
  $\surd$ &
  11 &
  log &
  EPE\textless{}$\mu_\epsilon$+3$b_\epsilon$ &
  99.24 &
  \best{0.3109} &
  \best{1.392} &
  0.5234 &
  0.1454 &
  \best{4.1726} &
  \best{9.7637} \\
 &
  +SEDNet &
  - &
  $\surd$ &
  $\surd$ &
  $\surd$ &
  11 &
  log &
  EPE\textless{}$\mu_\epsilon$+5$b_\epsilon$ &
  99.68 &
  0.3236 &
  1.427 &
  \best{0.3561} &
  \best{0.1096} &
  4.2767 &
  9.9843 \\
  \hline
  \hline
  \multirow{10}{*}{VK2-S6-Moving} &
  GwcNet &
  - &
   $\surd$ &
  - &
  - &
  - &
  - &
  - &
  - &
  0.4253 &
  1.689 &
  - &
  - &
  5.9184 &
  - \\
   &
   GwcNet &
  - &
  $\surd$ &
  - &
  - &
  - &
  - &
  EPE\textless{}5 &98.85
   &0.4543
   &1.593
   &- &- 
   &5.7884
   &
  - \\
  &
   GwcNet &
  - &
  $\surd$ &
  - &
  - &
  - &
  log &
  EPE\textless{}$\mu_\epsilon$+3$b_\epsilon$ &99.60
   &0.4708
   &1.707
   &
  - &
  - &6.2409
   &
  - \\
 &
  +$\mathcal{L}_{log}$ &
  - &
   $\surd$ &
   $\surd$ &
  - &
  - &
  - &
  - &
  - &
  0.4618 &
  1.930 &
  0.5592 &
  0.1176 &
  4.6928 &
  9.3604 \\
 &
  +$\mathcal{L}_{log}$ &
  - &
   $\surd$ &
   $\surd$ &
  - &
  - &
  - &
  EPE\textless{}5 &
  98.91 &
  0.4231 &
  1.537 &
  0.4575 &
  0.1890 &
  4.3663 &
  11.3532 \\
   &
  +$\mathcal{L}_{log}$ &
  - &
  $\surd$ &
  $\surd$ &
  - &
  - &
  log &
  EPE\textless{}$\mu_\epsilon$+3$b_\epsilon$ &99.63
   &0.4280
   &1.632
  &0.4885
  &0.2406
   &4.7709
   &12.5484
  \\
  &
  +SEDNet &
  - &
  $\surd$ &
  $\surd$ &
  $\surd$ &
  - &
  - &
  EPE\textless{}5 &98.67
   &0.5581
   &1.823
   &0.5806
   &0.1910
   &4.9404
   &12.1068
   \\
 &
  +SEDNet &
  - &
   $\surd$ &
   $\surd$ &
   $\surd$ &
  11 &
  log &
  EPE\textless{}$\mu_\epsilon$+1$b_\epsilon$ &
  98.77 &
  0.4220 &
  1.567 &
  0.4277 &
  0.1442 &
  4.1986 &
  10.8037 \\
 &
  +SEDNet &
  - &
   $\surd$ &
   $\surd$ &
   $\surd$ &
  11 &
  log &
  EPE\textless{}$\mu_\epsilon$+3$b_\epsilon$ &
  99.62 &
  \best{0.3577} &
  \best{1.389} &
  0.5958 &
  0.1573 &
  \best{3.9012} &
  \best{8.8339} \\
 &
  +SEDNet &
  - &
   $\surd$ &
   $\surd$ &
   $\surd$ &
  11 &
  log &
  EPE\textless{}$\mu_\epsilon$+5$b_\epsilon$ &
  99.76 &
  0.3862 &
  1.420 &
  \best{0.4002} &
   \best{0.1164} &
  4.0423 &
  9.0631 \\
  \hline
  \hline
\multirow{2}{*}{DrivingStereo} &
  +$\mathcal{L}_{log}$(FT) &
  - &
  $\surd$ &
  $\surd$ &
  - &
  - &
  - &
  - &
  - &
  0.5332 &
  0.2641 &
  0.3449 &
  0.2297 &
  21.7002 &
  45.7096 \\
 &
  +SEDNet(FT) &
  - &
  $\surd$ &
  $\surd$ &
  $\surd$ &
  11 &
  log &
  EPE\textless{}$\mu_\epsilon$+5$b_\epsilon$ &
  99.86 &
  \best{0.5264} &
  \best{0.2439} &
  \best{0.3324} &
  \best{0.2267} &
  \best{21.2856} &
  \best{44.3297} \\
  \hline
  \hline
  \multirow{10}{*}{DS-Weather} &
  GwcNet &
  - &
  $\surd$ &
  - &
  - &
  - &
  - &
  - &
  - &
  1.6962 &
  8.313 &
  - &
  - &
  44.4896 &
  - \\
   &
   GwcNet &
  - &
  $\surd$ &
  - &
  - &
  - &
  - &
  EPE\textless{}5 &73.05
   &18.2891
   &30.856
   &
  - &
  - &358.1623
   &
  - \\
  &
   GwcNet &
  - &
  $\surd$ &
  - &
  - &
  - &
  log &
  EPE\textless{}$\mu_\epsilon$+3$b_\epsilon$ &97.87
   &14.3547
   &31.835
   &
  - &
  - &315.8265
   &
  - \\
 &
  +$\mathcal{L}_{log}$ &
  - &
  $\surd$ &
  $\surd$ &
  - &
  - &
  - &
  - &
  - &
  1.9458 &
  8.700 &
  6.2375 &
  0.8295 &
  43.3146 &
  127.4829 \\
 &
  +$\mathcal{L}_{log}$ &
  - &
  $\surd$ &
  $\surd$ &
  - &
  - &
  - &
  EPE\textless{}5 &
  95.78 &
  2.3944 &
  6.666 &
  2.1443  &
  \best{0.4383} &
  41.1909 &
  95.4264 \\
   &
  +$\mathcal{L}_{log}$ &
  - &
  $\surd$ &
  $\surd$ &
  - &
  - &
  log &
  EPE\textless{}$\mu_\epsilon$+3$b_\epsilon$ &98.32
   &5.4850
   &15.136
  &5.1234
  &0.7290
   &90.5541
   &175.9993
  \\
  &
  +SEDNet &
  - &
  $\surd$ &
  $\surd$ &
  $\surd$ &
  - &
  - &
  EPE\textless{}5 &94.33
   &3.7375
   &9.297
   &3.3794
   &0.5249
   &49.9049
   &124.0092
   \\
 &
  +SEDNet &
  - &
  $\surd$ &
  $\surd$ &
  $\surd$ &
  11 &
  log &
  EPE\textless{}$\mu_\epsilon$+1$b_\epsilon$ &
  95.54 &
  6.9335 &
  9.346 &
  6.1682 &
  0.4946 &
  53.5638 &
  99.2891 \\
 &
  +SEDNet &
  - &
  $\surd$ &
  $\surd$ &
  $\surd$ &
  11 &
  log &
  EPE\textless{}$\mu_\epsilon$+3$b_\epsilon$ &
  98.95 &
  \best{1.5637} &
  6.508 &
  2.3406 &
  0.5309 &
  \best{38.4871} &
  \best{86.1118} \\
 &
  +SEDNet &
  - &
  $\surd$ &
  $\surd$ &
  $\surd$ &
  11 &
  log &
  EPE\textless{}$\mu_\epsilon$+5$b_\epsilon$ &
  99.41 &
  1.7051 &
  \best{6.057} &
  \best{1.5842} &
  0.6104 &
  39.8057 &
  87.1882 \\
  \hline
\end{tabular}
    \end{adjustbox}
    \caption{Quantitative results: (1) \textit{within-domain} on SceneFlow, VK2-S6 and VK2-S6-Moving; (2) after finetuning (FT) on DrivingStereo; (3) \textit{cross-domain} on DS-Weather. The best results in each category in each experiment are highlighted in dark blue. The top-performing variant of \netname, namely \netname with EPE\textless{}$\mu_\epsilon$+3$b_\epsilon$, outperforms the baselines with respect to disparity and uncertainty metrics in the majority of experiments. }
    \label{tab:in_domian_big}
\end{table*}
We conduct ablation studies to explore the impact of the number and spacing between the histogram bins, as well as the inlier threshold. 
Table~\ref{tab:in_domian_big} summarizes the quantitative results obtained from different baselines and our proposed method trained on the three datasets.
(We selected the best variant of each method and presented a brief version of this table as Table \ref{tab:in_domian} in the main paper.)

\begin{figure}[tb]
\centering
\begin{tabular}{c}
\includegraphics[width=0.8\columnwidth]{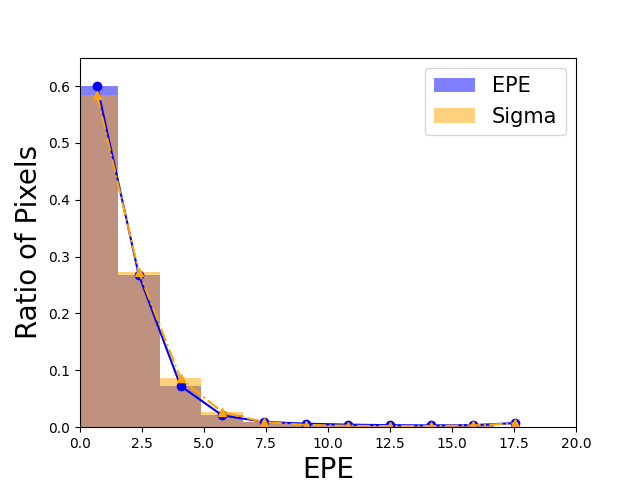}  \\
Linear bins \\
\includegraphics[width=0.8\columnwidth]{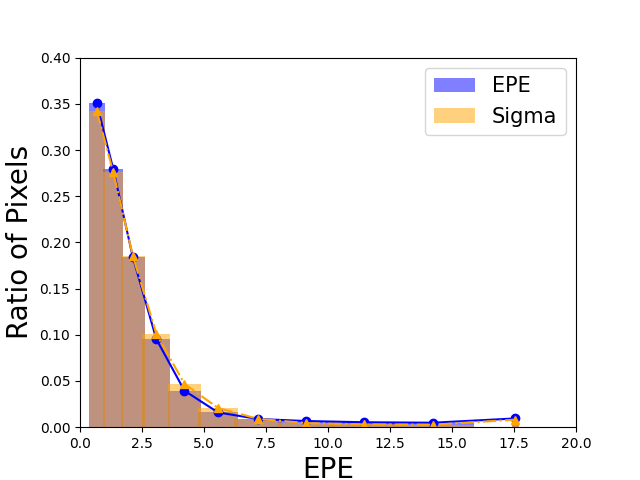} \\
Logarithmic bins \\
\end{tabular}
\vspace{-4pt}
\caption{Distributions of error and predicted uncertainty using different scaling of bins. The distributions correspond to the two rows of \netname with 11 bins in the \textit{Scene Flow} section of Table~\ref{tab:in_domian_big}.}
\label{fig:match_distribution}
\end{figure}

The first ablation study is to explore the configuration of the histograms, i.e. the $\alpha$ and $m$ in Section 3.2 in the main paper, with respect to the \textit{Bins} and \textit{Scale} under \textit{Loss} category in all tables. We performed this experiment on the Scene Flow dataset. When training \netname, we varied the number of bins and toggled between linear and logarithmic scaling. The results reveal that using bins defined in log space is better than linear space.
This is due to the fact that the distributions of errors and uncertainty are approximately Laplace distributions, and as a result, most samples are concentrated around the means. See the two rows corresponding to \netname with 11 bins in the \textit{Scene Flow} section of Table~\ref{tab:in_domian_big} for a comparison of linearly and logarithmically spaced bins. The distributions corresponding to these two rows are also plotted in Figure~\ref{fig:match_distribution}. Moreover, we find that increasing the number of bins for soft-histogramming does not increase the accuracy but only the computational cost. See the last four rows in the \textit{Scene Flow} section in Table~\ref{tab:in_domian_big} that shows \netname results with 11, 20 or 50 bins. 

The second ablation study revolves around the selection of the inlier threshold, i.e. the \textit{Inliers} category in all tables.
We first tried GwcNet, \kg and \netname with a fixed threshold (EPE\textless{}$5$) and then an adaptive threshold (\textless{}$\mu_\epsilon$+1$b_\epsilon$) when training on Scene Flow and VK2. We find that GwcNet and \kg are more compatible with a fixed threshold, while \netname works better when using an adaptive threshold. We also changed the threshold of \netname to be $\mu_\epsilon$+1$b_\epsilon$ and $\mu_\epsilon$+5$b_\epsilon$ when training models on VK2. In most cases, models with the threshold at $3b_\epsilon$ are better than others.
This does not hold for the evaluation on DS-Weather, where the network lacks knowledge of the unseen domain. Information from the error maps available when the threshold is larger may support a better prediction.

We also computed the precise percentage of the inliers in each experiment, see \textit{Pct} in Table~\ref{tab:in_domian_big}.
The results indicate that the threshold, fixed or adaptive, should not be too restrictive because the network does not improve if back-propagation only occurs on pixels with small errors. At the same time, outliers contaminate the solution and sometimes hinder convergence.

\section{Generalization from Synthetic to Real Data}
\label{sec:syn_2_real}
\pdfoutput=1

Here, we present more quantitative results on the synthetic to real data experiments supplementing Table \ref{tab:syt_to_real} in the main paper. Table~\ref{tab:syt_to_real_big} presents results from multiple variants of each method, including with and without inlier filtering, and different  fixed or adaptive thresholds. The results are consistent with the findings in the ablation studies. 

It is worth noting that on \textit{DS-Rainy}, \netname with an inlier threshold of $\mu_\epsilon+1b_\epsilon$ exhibits very poor performance due to the fact that only 85.82\% of the pixels are considered inliers. An overly restrictive inlier threshold, fixed or adaptive, is harmful to the performance of the network, since the back-propagation only occurs on pixels with small errors and the network does not benefit from hard examples. 

\pdfoutput=1

\begin{table*}[p]
\begin{adjustbox}{width=0.8\textwidth,center}
    \centering
    \begin{tabular}{|c|c|cccccc|cc|cc|cc|cc|}
    \hline
\multirow{2}{*}{\textbf{Dataset}} &
  \multirow{2}{*}{\textbf{Method}} &
  \multicolumn{6}{c|}{\textbf{Loss}} &
  \multicolumn{2}{c|}{\textbf{Inliers}} &
  \multicolumn{2}{c|}{\textbf{Disparity}$\downarrow$} &
  \multicolumn{2}{c|}{\textbf{APE}$\downarrow$} &
  \multicolumn{2}{c|}{\textbf{AUC}$\downarrow$} \\
  \cline{3-16}
 &
   &
  BCE &
  L1 &
  Log &
  KL &
  Bins &
  Scale &
  Def. &
  Pct(\%) &
  EPE &
  D1(\%) &
  Avg. &
  Median &
  Opt. &
  Est. \\
  \hline
  \hline
  \multirow{10}{*}{VK2-S6-Morning} &
  GwcNet &
  - &
   $\surd$ &
  - &
  - &
  - &
  - &
  - &
  - &
  0.4642 &
  1.740 &
  - &
  - &
  6.1845 &
  - \\
    &
   GwcNet &
  - &
  $\surd$ &
  - &
  - &
  - &
  - &
  EPE\textless{}5 & 
  98.79 & 
  0.5107 & 
  1.649 &
  - &
  - &
  6.0792 &
  - \\
  &
   GwcNet &
  - &
  $\surd$ &
  - &
  - &
  - &
  log &
  EPE\textless{}$\mu_\epsilon$+3$b_\epsilon$ &
  99.57 &
  0.5065 &
  1.742 &
  - &
  - &
 6.4706  &
  - \\ 
  &
  +$\mathcal{L}_{log}$ &
  - &
   $\surd$ &
   $\surd$ &
  - &
  - &
  - &
  - &
  - &
  0.5117  &
  1.998 &
   0.6616 &
   0.1162&
   5.0563 &
   10.0704
   \\
 &
  +$\mathcal{L}_{log}$ &
  - &
   $\surd$ &
   $\surd$ &
  - &
  - &
  - &
  EPE\textless{}5 & 98.82
   & 0.4774
   & 1.624
   & 0.5067
   & 0.1872
   & 4.6698
   & 12.5192
   \\
   &
  +$\mathcal{L}_{log}$ &
  - &
   $\surd$ &
   $\surd$ &
  - &
  11 &
  log &
  EPE\textless{}$\mu_\epsilon$+3$b_\epsilon$ & 99.59
   & 0.4571
   & 1.614
   & 0.5063
 & 0.2231
  & 4.8135
  & 12.2921
 \\
  &
  +SEDNet &
  - &
  $\surd$ &
  $\surd$ &
  $\surd$ &
  - &
  - &
  EPE\textless{}5 &
  98.58 &
  0.6136 &
  1.928 &
  0.6300 &
  0.1897 &
  5.2304 &
  13.1250 \\
 &
  +SEDNet &
  - &
   $\surd$ &
   $\surd$ &
   $\surd$ &
  11 &
  log &
  EPE\textless{}$\mu_\epsilon$+1$b_\epsilon$ & 98.83
   & 0.4774
   & 1.626
   & 0.4768
   & 0.1431
   & 4.4936
   & 11.8164
   \\
 &
  +SEDNet &
  - &
   $\surd$ &
   $\surd$ &
   $\surd$ &
  11 &
  log &
  EPE\textless{}$\mu_\epsilon$+3$b_\epsilon$ & 99.62
   & \best{0.4003}
  & \best{1.442}
  & 0.6183
   & 0.1553
   & \best{4.1847}
   & \best{9.4063}
   \\
 &
  +SEDNet &
  - &
   $\surd$ &
   $\surd$ &
   $\surd$ &
  11 &
  log &
  EPE\textless{}$\mu_\epsilon$+5$b_\epsilon$ & 99.71
   & 0.4265
   & 1.481
   & \best{0.4356}
   & \best{0.1150}
    & 4.3216
   & 9.6694
   \\
   \hline
    \multirow{10}{*}{VK2-S6-Sunset} &
  GwcNet &
  - &
   $\surd$ &
  - &
  - &
  - &
  - &
  - &
  - &
  0.4810 & 
  1.825 &
  - &
  - &
  6.6907 &
  - \\
   &
   GwcNet &
  - &
  $\surd$ &
  - &
  - &
  - &
  - &
  EPE\textless{}5 & 
  98.76 & 
  0.5222 & 
  1.701 &
  - &
  - &
  6.5110 &
  - \\
  &
   GwcNet &
  - &
  $\surd$ &
  - &
  - &
  - &
  log &
  EPE\textless{}$\mu_\epsilon$+3$b_\epsilon$ &
  99.57 &
  0.5345 &
  1.795 &
  - &
  - &
  6.9508 &
  - \\ 
 &
  +$\mathcal{L}_{log}$ &
  - &
   $\surd$ &
   $\surd$ &
  - &
  - &
  - &
  - &
  - &
  0.5112 &
  2.040 &
  0.6134 &
  0.1137 &
  5.4426 & 11.1299
   \\
 &
  +$\mathcal{L}_{log}$ &
  - &
   $\surd$ &
   $\surd$ &
  - &
  - &
  - &
  EPE\textless{}5 & 98.84
   & 0.4863
   & 1.627
   & 0.5060
   & 0.1827
   & 5.0075
   & 13.7848
   \\
   &
  +$\mathcal{L}_{log}$ &
  - &
   $\surd$ &
   $\surd$ &
  - &
  11 &
  log &
  EPE\textless{}$\mu_\epsilon$+3$b_\epsilon$ & 99.62
   & 0.4678
   & 1.646
   & 0.5046
 & 0.2151
  & 5.2551
  & 13.8256
 \\
 &
  +SEDNet &
  - &
  $\surd$ &
  $\surd$ &
  $\surd$ &
  - &
  - &
  EPE\textless{}5 &
  98.54 &
  0.6506 &
  1.981 &
  0.6558 &
  0.1827 &
  5.7348 &
  14.6926 \\
 &
  +SEDNet &
  - &
   $\surd$ &
   $\surd$ &
   $\surd$ &
  11 &
  log &
  EPE\textless{}$\mu_\epsilon$+1$b_\epsilon$ & 98.81
   & 0.4871
   & 1.664
   & 0.4764
   & 0.1356
   & 4.9970
   & 13.4379
   \\
 &
  +SEDNet &
  - &
   $\surd$ &
   $\surd$ &
   $\surd$ &
  11 &
  log &
  EPE\textless{}$\mu_\epsilon$+3$b_\epsilon$ & 99.61
   &\best{0.4108}
  &\best{1.475}
  & 0.6189
   & 0.1509
   &\best{4.5840}
   &\best{10.7946}
   \\
 &
  +SEDNet &
  - &
   $\surd$ &
   $\surd$ &
   $\surd$ &
  11 &
  log &
  EPE\textless{}$\mu_\epsilon$+5$b_\epsilon$ & 99.72
   & 0.4422
   & 1.505
   & \best{0.4399}
   &\best{0.1103}
    & 4.8745
   & 11.0680
   \\ 
   \hline
     \multirow{10}{*}{VK2-S6-Fog} &
  GwcNet &
  - &
   $\surd$ &
  - &
  - &
  - &
  - &
  - &
  - &
  0.4660 &
  1.812 &
  - &
  - &
  6.8355 &
  - \\
   &
   GwcNet &
  - &
  $\surd$ &
  - &
  - &
  - &
  - &
  EPE\textless{}5 & 
  98.91 & 
  0.4817 & 
  1.556 &
  - &
  - &
  6.4733 &
  - \\
  &
   GwcNet &
  - &
  $\surd$ &
  - &
  - &
  - &
  log &
  EPE\textless{}$\mu_\epsilon$+3$b_\epsilon$ &
  99.66 &
  0.5073 &
  1.810 &
  - &
  - &
  7.1501 &
  - \\ 
 &
  +$\mathcal{L}_{log}$ &
  - &
   $\surd$ &
   $\surd$ &
  - &
  - &
  - &
  - &
  - &
  0.4919 &
  1.859 &
  0.5574 &
  0.1174 &
  5.3870 &
  11.1750
   \\
 &
  +$\mathcal{L}_{log}$ &
  - &
   $\surd$ &
   $\surd$ &
  - &
  - &
  - &
  EPE\textless{}5 & 98.98
   & 0.4425
   & 1.448
   &  0.4609
   & 0.1865
   & 4.8983
   & 12.1305
   \\
   &
  +$\mathcal{L}_{log}$ &
  - &
   $\surd$ &
   $\surd$ &
  - &
  11 &
  log &
  EPE\textless{}$\mu_\epsilon$+3$b_\epsilon$ & 99.69
   & 0.4330
   & 1.490
   & 0.4671
 & 0.2211
  & 5.2211
  & 12.5494
 \\
 &
  +SEDNet &
  - &
  $\surd$ &
  $\surd$ &
  $\surd$ &
  - &
  - &
  EPE\textless{}5 &
  98.88 &
  0.5410 &
  1.579 &
  0.5398 &
   0.1880  &
  5.4246 &
  12.2987 \\
 &
  +SEDNet &
  - &
   $\surd$ &
   $\surd$ &
   $\surd$ &
  11 &
  log &
  EPE\textless{}$\mu_\epsilon$+1$b_\epsilon$ & 98.90
   & 0.4657
   & 1.533
   & 0.4459
   & 0.1415
   & 4.9125
   & 12.1241
   \\
 &
  +SEDNet &
  - &
   $\surd$ &
   $\surd$ &
   $\surd$ &
  11 &
  log &
  EPE\textless{}$\mu_\epsilon$+3$b_\epsilon$ & 99.71
   &\best{0.3731}
  & \best{1.288}
  & 0.5517
   & 0.1547
   & \best{4.4200}
   & \best{9.9380}
   \\
 &
  +SEDNet &
  - &
   $\surd$ &
   $\surd$ &
   $\surd$ &
  11 &
  log &
  EPE\textless{}$\mu_\epsilon$+5$b_\epsilon$ & 99.77
   & 0.4108
   & 1.339
   &\best{0.4162 }
   & \best{0.1156}
    & 4.5310
   & 10.0341
   \\
  \hline
    \multirow{10}{*}{VK2-S6-Rain} &
  GwcNet &
  - &
   $\surd$ &
  - &
  - &
  - &
  - &
  - &
  - &
  0.4618 &
  1.700 &
  - &
  - &
 6.6774  &
  - \\
   &
   GwcNet &
  - &
  $\surd$ &
  - &
  - &
  - &
  - &
  EPE\textless{}5 & 
  98.84 & 
  0.5030 & 
  1.633 &
  - &
  - &
  6.5983 &
  - \\
  &
   GwcNet &
  - &
  $\surd$ &
  - &
  - &
  - &
  log &
  EPE\textless{}$\mu_\epsilon$+3$b_\epsilon$ &
  99.61 &
  0.5197 &
  1.835 &
  - &
  - &
  7.1739 &
  - \\ 
 &
  +$\mathcal{L}_{log}$ &
  - &
   $\surd$ &
   $\surd$ &
  - &
  - &
  - &
  - &
  - &
  0.5160 &
  1.989 &
  0.6064 &
  0.1199  &
  5.3429 & 10.8433
   \\
 &
  +$\mathcal{L}_{log}$ &
  - &
   $\surd$ &
   $\surd$ &
  - &
  - &
  - &
  EPE\textless{}5 & 98.88
   & 0.4707
   & 1.571
   & 0.4899
   & 0.1861
   & 4.9351
   & 13.3214
   \\
   &
  +$\mathcal{L}_{log}$ &
  - &
   $\surd$ &
   $\surd$ &
  - &
  11 &
  log &
  EPE\textless{}$\mu_\epsilon$+3$b_\epsilon$ & 99.66
   & 0.4543
   & 1.565
   & 0.4914
 & 0.2198
  & 5.1625
  & 13.7641
 \\
 &
  +SEDNet &
  - &
   $\surd$ &
   $\surd$ &
   $\surd$ &
  11 &
  log &
  EPE\textless{}$\mu_\epsilon$+1$b_\epsilon$ & 98.82
   & 0.4701
   & 1.611
   & 0.4710
   & 0.1413
   & 4.7792
   & 12.8168
   \\
   &
  +SEDNet &
  - &
  $\surd$ &
  $\surd$ &
  $\surd$ &
  - &
  - &
  EPE\textless{}5 &
  98.75 &
  0.5868 &
  1.751 &
  0.5833 &
  0.1865 &
  5.5200 &
  14.5175 \\
 &
  +SEDNet &
  - &
   $\surd$ &
   $\surd$ &
   $\surd$ &
  11 &
  log &
  EPE\textless{}$\mu_\epsilon$+3$b_\epsilon$ & 99.69
   &\best{0.3873}
  &\best{1.356}
  & 0.6685
   &  0.1537
   & \best{4.4013}
   &\best{10.3362}
   \\
 &
  +SEDNet &
  - &
   $\surd$ &
   $\surd$ &
   $\surd$ &
  11 &
  log &
  EPE\textless{}$\mu_\epsilon$+5$b_\epsilon$ & 99.74
   & 0.4383
   & 1.461
   &\best{0.4557}
   &\best{0.1171}
    & 4.5840
   & 10.6757
   \\
   \hline
   \hline
     \multirow{10}{*}{DS-Cloudy} &
  GwcNet &
  - &
   $\surd$ &
  - &
  - &
  - &
  - &
  - &
  - &
   1.3413 &
  5.229 &
  - &
  - &
  37.4263 &
  - \\
     &
   GwcNet &
  - &
  $\surd$ &
  - &
  - &
  - &
  - &
  EPE\textless{}5 &84.75
   &7.3789
   &18.559
   &
  - &
  - &71.6779
   &
  - \\
  &
   GwcNet &
  - &
  $\surd$ &
  - &
  - &
  - &
  log &
  EPE\textless{}$\mu_\epsilon$+3$b_\epsilon$ &97.19
   &7.8976
   &21.581
   &
  - &
  - &88.6479
   &
  - \\
 &
  +$\mathcal{L}_{log}$ &
  - &
   $\surd$ &
   $\surd$ &
  - &
  - &
  - &
  - &
  - &
  1.8379  &
  6.731 &
  3.3032  &
    0.6810 &
  37.8238  & 159.3730
   \\
 &
  +$\mathcal{L}_{log}$ &
  - &
   $\surd$ &
   $\surd$ &
  - &
  - &
  - &
  EPE\textless{}5 & 97.48
   & 1.4780 
   & 3.948
   & 1.2617
   & \best{0.3513}
   & 34.4488
   & 82.5380
   \\
   &
  +$\mathcal{L}_{log}$ &
  - &
   $\surd$ &
   $\surd$ &
  - &
  11 &
  log &
  EPE\textless{}$\mu_\epsilon$+3$b_\epsilon$ & 98.78
   & 1.9784
   & 5.719
   & 1.7444
 & 0.3785
  & 39.2511
  & 91.5997
 \\
 &
  +SEDNet &
  - &
  $\surd$ &
  $\surd$ &
  $\surd$ &
  - &
  - &
  EPE\textless{}5 &96.96
   &2.0096
   &4.844
   &1.7492
   &0.3930
   &37.8752
   &95.2040
   \\
 &
  +SEDNet &
  - &
   $\surd$ &
   $\surd$ &
   $\surd$ &
  11 &
  log &
  EPE\textless{}$\mu_\epsilon$+1$b_\epsilon$ & 97.11
   & 4.0438
   & 6.101
   & 3.5618
   & 0.3936
   & 40.3969
   & 82.5675
   \\
 &
  +SEDNet &
  - &
   $\surd$ &
   $\surd$ &
   $\surd$ &
  11 &
  log &
  EPE\textless{}$\mu_\epsilon$+3$b_\epsilon$ & 98.83
   & 1.3183
  & 4.414
  & 1.5260
   & 0.4021 
   & 33.9037
   & 73.6330
   \\
 &
  +SEDNet &
  - &
   $\surd$ &
   $\surd$ &
   $\surd$ &
  11 &
  log &
  EPE\textless{}$\mu_\epsilon$+5$b_\epsilon$ & 99.41
   & \best{1.1901}
   &\best{3.772}
   &\best{1.0795}
   & 0.4736
    &\best{32.5237}
   &\best{69.6368}
   \\
   \hline
     \multirow{10}{*}{DS-Sunny} &
  GwcNet &
  - &
   $\surd$ &
  - &
  - &
  - &
  - &
  - &
  - &
  1.5448 &
  6.991 &
  - &
  - &
  38.7386 &
  - \\
     &
   GwcNet &
  - &
  $\surd$ &
  - &
  - &
  - &
  - &
  EPE\textless{}5 &88.95
   &5.2974
   &14.292
   &
  - &
  - &56.5606
   &
  - \\
  &
   GwcNet &
  - &
  $\surd$ &
  - &
  - &
  - &
  log &
  EPE\textless{}$\mu_\epsilon$+3$b_\epsilon$ &97.61
   &4.8873
   &17.035
   &
  - &
  - &62.4926
   &
  - \\
 &
  +$\mathcal{L}_{log}$ &
  - &
   $\surd$ &
   $\surd$ &
  - &
  - &
  - &
  - &
  - &
 1.5645   &
  6.039 &
  3.1431 &
  0.8429 &
  36.3650 & 76.7900
   \\
 &
  +$\mathcal{L}_{log}$ &
  - &
   $\surd$ &
   $\surd$ &
  - &
  - &
  - &
  EPE\textless{}5 & 97.08
   & 1.4837
   & 4.631
   & 1.2806
   & 0.3835
   & 35.5226
   & 85.8715
   \\
   &
  +$\mathcal{L}_{log}$ &
  - &
   $\surd$ &
   $\surd$ &
  - &
  11 &
  log &
  EPE\textless{}$\mu_\epsilon$+3$b_\epsilon$ & 98.57
   & 1.7809
   & 6.131
   & 1.5580
 & \best{0.3828}
  & 38.5942
  & 97.1704
 \\
 &
  +SEDNet &
  - &
  $\surd$ &
  $\surd$ &
  $\surd$ &
  - &
  - &
  EPE\textless{}5 &95.43
   &2.6221
   &7.001
   &2.3380
   &0.4468
   &41.8861
   &108.8739
   \\
 &
  +SEDNet &
  - &
   $\surd$ &
   $\surd$ &
   $\surd$ &
  11 &
  log &
  EPE\textless{}$\mu_\epsilon$+1$b_\epsilon$ & 97.24
   & 3.5945
   & 6.274
   & 3.6359
   & 0.3894
   & 38.2571
   & 85.1056
   \\
 &
  +SEDNet &
  - &
   $\surd$ &
   $\surd$ &
   $\surd$ &
  11 &
  log &
  EPE\textless{}$\mu_\epsilon$+3$b_\epsilon$ & 98.64
   & 1.5548
  & 5.878
  & 3.0025 
   & 0.4808
   & 35.6523
   & 83.2573
   \\
 &
  +SEDNet &
  - &
   $\surd$ &
   $\surd$ &
   $\surd$ &
  11 &
  log &
  EPE\textless{}$\mu_\epsilon$+5$b_\epsilon$ & 99.27
   &\best{1.4164}
   &\best{5.219}
   & \best{1.3549}
   & 0.6110
    & \best{34.0465}
   & \best{83.2316}
   \\
   \hline
     \multirow{10}{*}{DS-Foggy} &
  GwcNet &
  - &
   $\surd$ &
  - &
  - &
  - &
  - &
  - &
  - &
 1.5476  &
  8.859 &
  - &
  - &
  51.4640 &
  - \\
     &
   GwcNet &
  - &
  $\surd$ &
  - &
  - &
  - &
  - &
  EPE\textless{}5 &83.04
   &10.1839
   & 22.130
   &
  - &
  - &105.1603
   &
  - \\
  &
   GwcNet &
  - &
  $\surd$ &
  - &
  - &
  - &
  log &
  EPE\textless{}$\mu_\epsilon$+3$b_\epsilon$ &97.34
   &5.0526
   &20.534
   &
  - &
  - &80.6149
   &
  - \\
 &
  +$\mathcal{L}_{log}$ &
  - &
   $\surd$ &
   $\surd$ &
  - &
  - &
  - &
  - &
  - &
  1.6435  &
  9.694 &
  4.0449 &
  0.8879 &
 49.2533  & \best{95.5706}
   \\
 &
  +$\mathcal{L}_{log}$ &
  - &
   $\surd$ &
   $\surd$ &
  - &
  - &
  - &
  EPE\textless{}5 & 94.89
   & 2.9553
   & 9.015
   &  2.6923
   & 0.5556
   &48.7136
   & 101.7025
   \\
   &
  +$\mathcal{L}_{log}$ &
  - &
   $\surd$ &
   $\surd$ &
  - &
  11 &
  log &
  EPE\textless{}$\mu_\epsilon$+3$b_\epsilon$ & 98.93
   & 1.6931 
   & 8.979
   & \best{1.4534}
 & 0.6143
  & 50.7000
  & 106.2925
 \\
 &
  +SEDNet &
  - &
  $\surd$ &
  $\surd$ &
  $\surd$ &
  - &
  - &
  EPE\textless{}5 &94.64
   &3.3875
   &11.676
   &3.0463
   &0.6894
   &57.0915
   &137.2929
   \\
 &
  +SEDNet &
  - &
   $\surd$ &
   $\surd$ &
   $\surd$ &
  11 &
  log &
  EPE\textless{}$\mu_\epsilon$+1$b_\epsilon$ & 96.15
   & 6.1046
   & 9.996
   & 5.1343
   & \best{0.6046}
   & 56.8943
   & 104.6067
   \\
 &
  +SEDNet &
  - &
   $\surd$ &
   $\surd$ &
   $\surd$ &
  11 &
  log &
  EPE\textless{}$\mu_\epsilon$+3$b_\epsilon$ & 99.27
   & \best{1.5398}
  & 7.357
  &  2.4109
   &  0.7023
   & 47.7932
   & 97.8627
   \\
 &
  +SEDNet &
  - &
   $\surd$ &
   $\surd$ &
   $\surd$ &
  11 &
  log &
  EPE\textless{}$\mu_\epsilon$+5$b_\epsilon$ & 99.50
   &  1.6536
   & \best{7.145}
   &  1.5196
   & 0.7310
    &\best{44.5539}
   & 99.5714
   \\
   \hline
     \multirow{10}{*}{DS-Rainy} &
  GwcNet &
  - &
   $\surd$ &
  - &
  - &
  - &
  - &
  - &
  - &
  3.1918 &
 17.356  &
  - &
  - &
  68.0346 &
  - \\
     &
   GwcNet &
  - &
  $\surd$ &
  - &
  - &
  - &
  - &
  EPE\textless{}5 &35.47
   &49.6661
   &68.441
   &
  - &
  - &1199.2505
   &
  - \\
  &
   GwcNet &
  - &
  $\surd$ &
  - &
  - &
  - &
  log &
  EPE\textless{}$\mu_\epsilon$+3$b_\epsilon$ &99.32
   &39.5814
   &68.191
   &
  - &
  - &1031.5507
   &
  - \\
 &
  +$\mathcal{L}_{log}$ &
  - &
   $\surd$ &
   $\surd$ &
  - &
  - &
  - &
  - &
  - &
  3.6950 &
  17.079 &
  24.7441 &
  1.0236 &
 67.2717  & 253.0992
   \\
 &
  +$\mathcal{L}_{log}$ &
  - &
   $\surd$ &
   $\surd$ &
  - &
  - &
  - &
  EPE\textless{}5 & 98.79
   & 5.3539
   & 12.501
   & 4.9480
   & \best{0.5759}
   &  59.3952
   & 146.8906
   \\
   &
  +$\mathcal{L}_{log}$ &
  - &
   $\surd$ &
   $\surd$ &
  - &
  11 &
  log &
  EPE\textless{}$\mu_\epsilon$+3$b_\epsilon$ & 96.98
   & 16.4877 
   & 39.713
   & 15.7376
 & 1.5403
  & 233.6712
  & 408.9347
 \\
 &
  +SEDNet &
  - &
  $\surd$ &
  $\surd$ &
  $\surd$ &
  - &
  - &
  EPE\textless{}5 &90.28
   &6.9306
   &13.668
   &6.3839
   &0.5705
   &62.7668
   &154.6661
   \\
 &
  +SEDNet &
  - &
   $\surd$ &
   $\surd$ &
   $\surd$ &
  11 &
  log &
  EPE\textless{}$\mu_\epsilon$+1$b_\epsilon$ & 85.82
   & 22.9318
   & 22.408
   & 20.1601
   & 0.8772
   & 219.9731
   & 278.7730
   \\
 &
  +SEDNet &
  - &
   $\surd$ &
   $\surd$ &
   $\surd$ &
  11 &
  log &
  EPE\textless{}$\mu_\epsilon$+3$b_\epsilon$ & 99.10
   &\best{2.2165}
  & 11.020
  & \best{2.6599}
   & 0.6722
   & \best{50.8103}
   & \best{110.8360}
   \\
 &
  +SEDNet &
  - &
   $\surd$ &
   $\surd$ &
   $\surd$ &
  11 &
  log &
  EPE\textless{}$\mu_\epsilon$+5$b_\epsilon$ & 98.95
   & 3.6734
   & \best{10.975}
   & 3.4255
   & 0.7346
    & 54.1441
   & 129.2394
   \\
   \hline
\end{tabular}
    \end{adjustbox}
    \caption{Quantitative results of synthetic to real evaluation. This is the extended version of Table 3 in the main paper. The best results in each category in each experiment are in dark blue. The top-performing Variant of \netname, namely \netname with EPE\textless{}$\mu_\epsilon$+3$b_\epsilon$, outperforms the baselines in the majority of experiments, especially in uncertainty estimation on real data and under adverse weather (i.e., foggy and rainy).}
    \label{tab:syt_to_real_big}
\end{table*}

\section{Additional Qualitative Results}
\label{sec:images}
\pdfoutput=1

In the following pages, we provide more qualitative results to demonstrate the effectiveness of the proposed method. We select examples from the three datasets mentioned in Section 4.1. The main differences 
are highlighted in red boxes.

\noindent\textbf{Scene Flow.} Figure~\ref{fig:sf_1} and Figure~\ref{fig:sf_2} show two examples from the Flying3D dataset. Unlike VK2
and DrivingStereo that include images of street views, this dataset provides image pairs of indoor objects. We find that \netname is good at capturing the uncertainty of the boundaries of overlapping objects, and at predicting more accurate disparity for textureless objects and objects with complicated structure such as holes.

\noindent\textbf{VK2.} Figure~\ref{fig:vk_weather} presents a comparison of disparity estimation under different weather conditions on the synthetic datasets. To further illustrate the strength of \netname in predicting disparity as well as accurate uncertainty, we pick two hard examples, in Figures~\ref{fig:vk_1} and \ref{fig:vk_s2}, from the foggy subset. \netname still performs very well considering the surrounding environment is hazy, while \kg fails to figure out the background.

\noindent\textbf{DS-Weather.} Real data acquired under adverse weather conditions exhibit more challenges than synthetic data under simulated similar weather conditions.
In addition to poor illumination and opacity, real data also suffer from 
reflections and the Tyndall effect. Another large challenge for this dataset is that the LIDAR ground truth is sparse. Figure~\ref{fig:ds_s_est} presents uncertainty estimates by the different methods under diverse illumination conditions. Figures~\ref{fig:ds_1}, \ref{fig:ds_s2} and \ref{fig:ds_3} further illustrate how \netname outperforms the baselines under adverse weather.

\renewcommand\hcwdf{0.45\textwidth}
\newcommand\twdf{.35\textwidth}


\begin{figure*}[p]
\centering
\begin{tabular}{cc}
\includegraphics[width=\hcwdf]{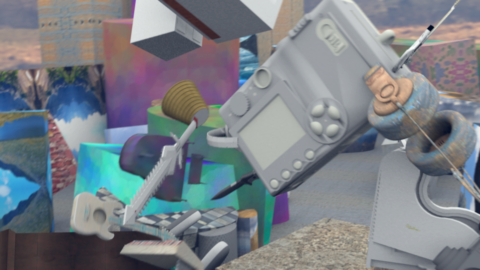} &
\includegraphics[width=\hcwdf]{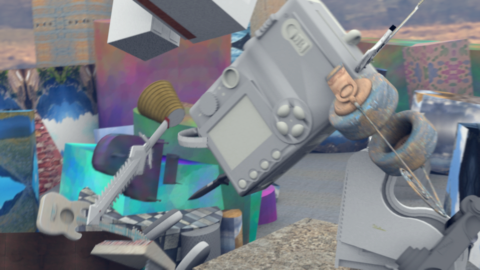} \\
Left Image & Right Image \\
\includegraphics[width=\hcwdf]{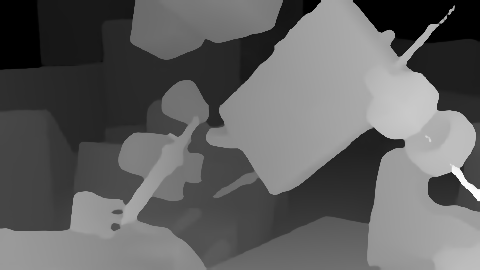} &
\includegraphics[width=\hcwdf]{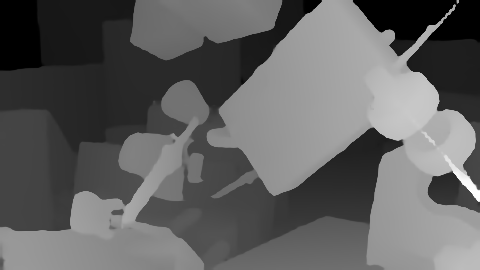}\\
\kg Disparity & \netname Disparity \\
\includegraphics[width=\hcwdf]{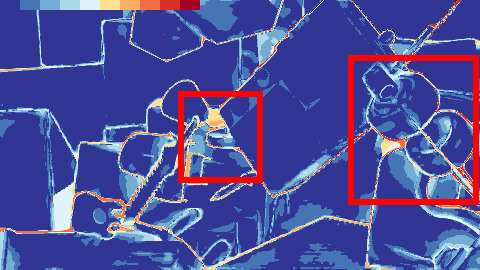} &
\includegraphics[width=\hcwdf]{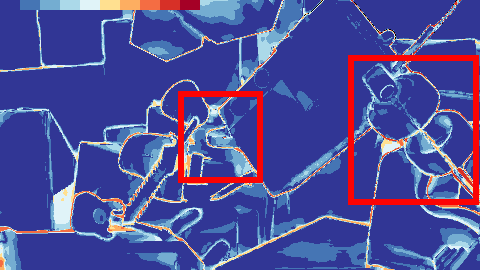}\\
\kg Error & \netname Error \\
\includegraphics[width=\hcwdf]{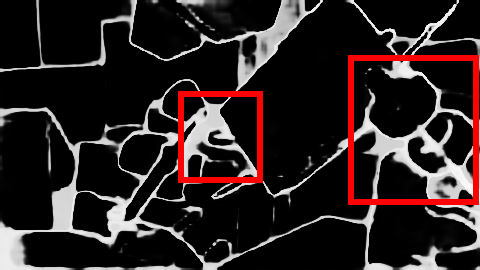} &
\includegraphics[width=\hcwdf]{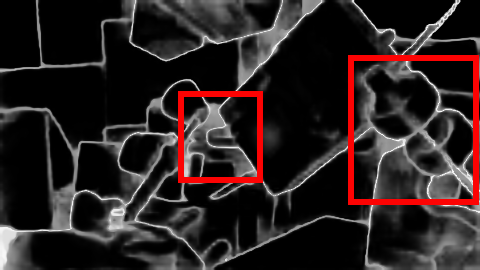}\\
\kg Uncertainty & \netname Uncertainty \\

\end{tabular}
\vspace{-4pt}
\caption{Example from Scene Flow. When objects overlap with each other and depth ordering is unclear, \netname captures the uncertainty more precisely according to the error map. In the regions outlined in red, \netname successfully detects the pull ring of the camera and the lid of the wheel, while the \kg model fails to estimate their disparity.}
\label{fig:sf_1}
\end{figure*}

\begin{figure*}[p]
\centering
\begin{tabular}{cc}
\includegraphics[width=\hcwdf]{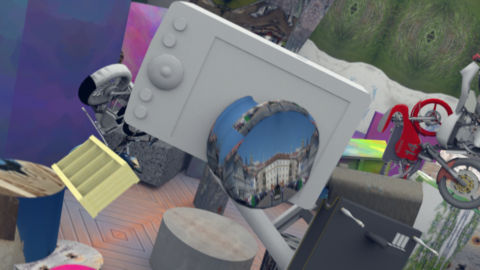} &
\includegraphics[width=\hcwdf]{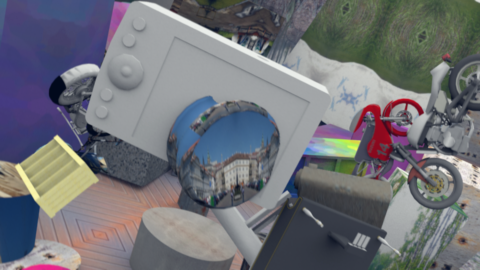} \\
Left Image & Right Image \\
\includegraphics[width=\hcwdf]{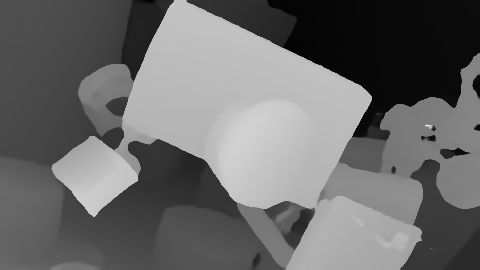} &
\includegraphics[width=\hcwdf]{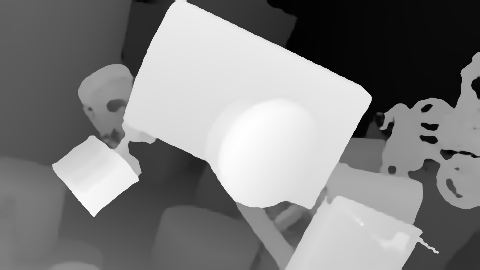}\\
\kg Disparity & \netname Disparity \\
\includegraphics[width=\hcwdf]{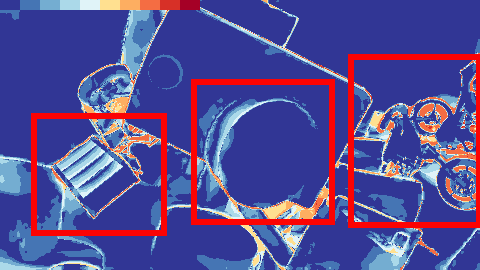} &
\includegraphics[width=\hcwdf]{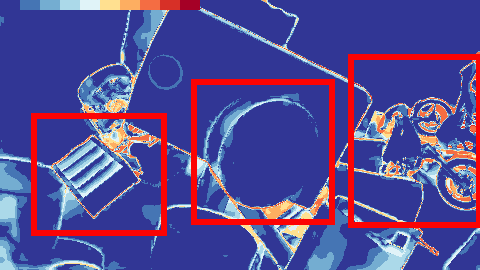}\\
\kg Error & \netname Error \\
\includegraphics[width=\hcwdf]{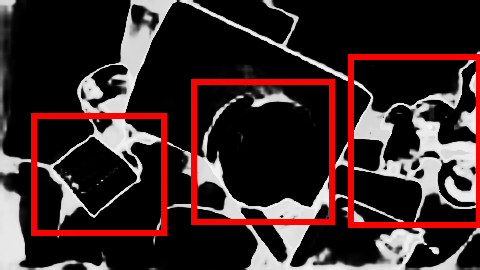} &
\includegraphics[width=\hcwdf]{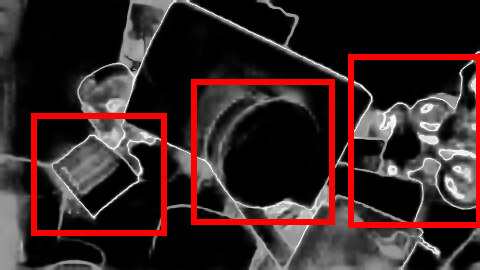}\\
\kg Uncertainty & \netname Uncertainty \\

\end{tabular}
\vspace{-4pt}
\caption{Example from Scene Flow. The disparity and uncertainty maps of \netname include some structure information of the objects, such as the bookcase on the left side which has several openings, the cylinder which has many intersecting surfaces, and the wheels of the motorcycle. The prediction of \kg lacks these details.}
\label{fig:sf_2}
\end{figure*}


\begin{figure*}[p]
\begin{adjustbox}{width=\textwidth,center}
\centering
\begin{tabular}{cccc}
Sunny &  Foggy & Rainy \\
\includegraphics[width=\twdf]{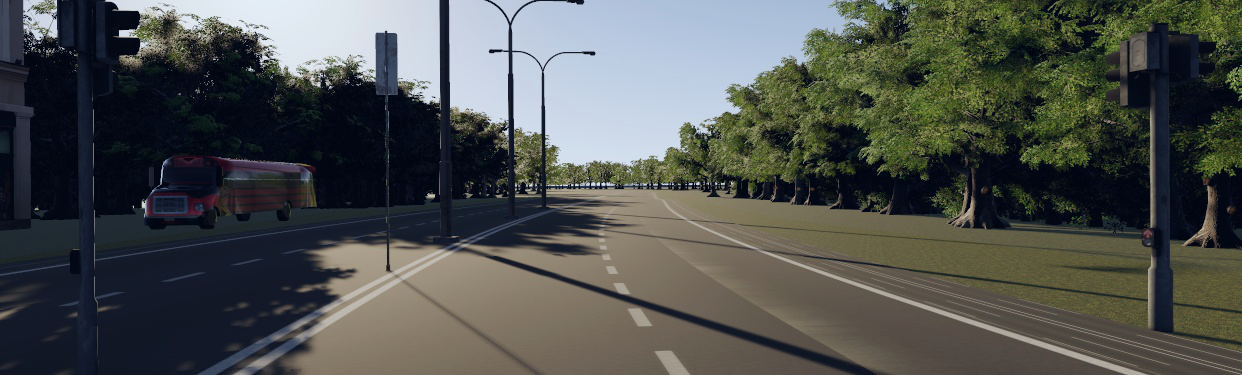} &
\includegraphics[width=\twdf]{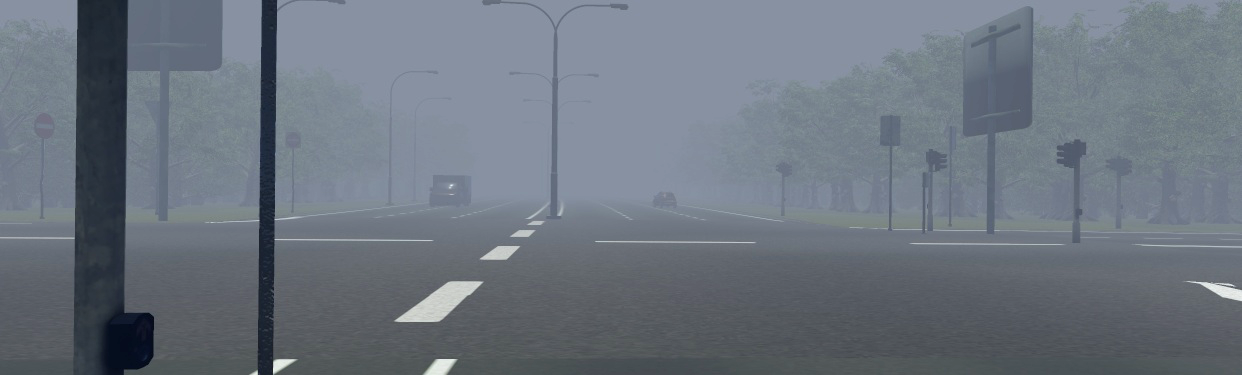} &
\includegraphics[width=\twdf]{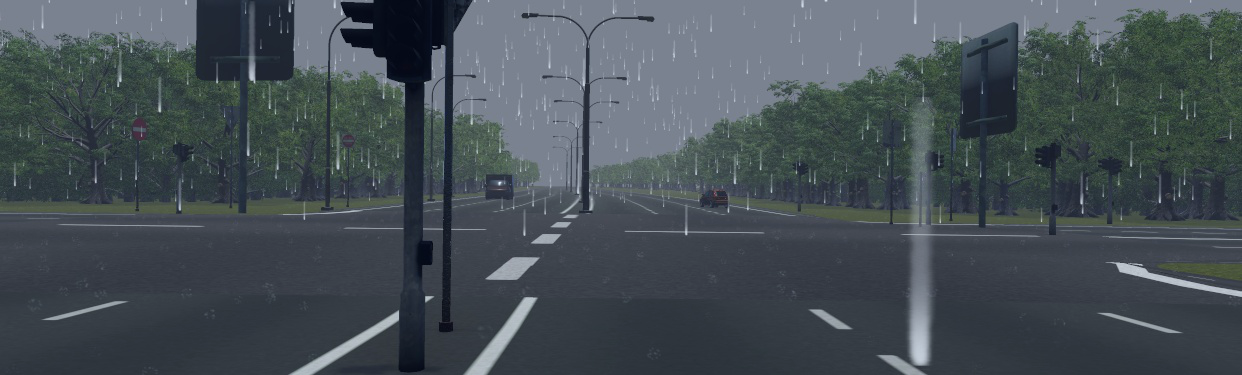} &
\\
&  Left Images & \\
\includegraphics[width=\twdf]{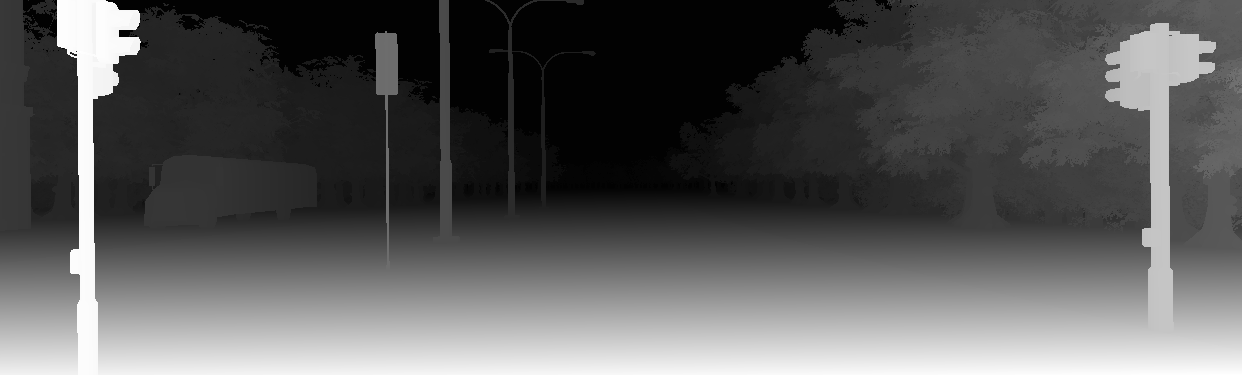} &
\includegraphics[width=\twdf]{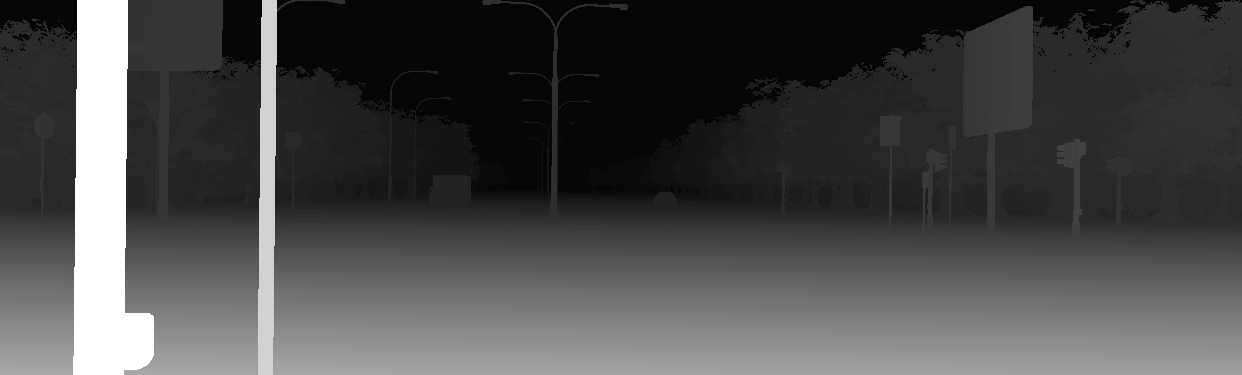} &
\includegraphics[width=\twdf]{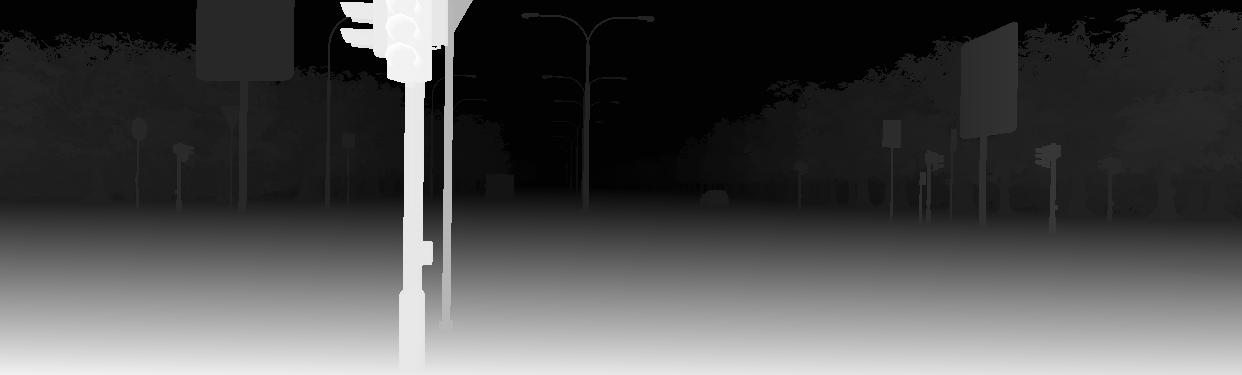} &
\\
& Ground Truth &\\
\includegraphics[width=\twdf]{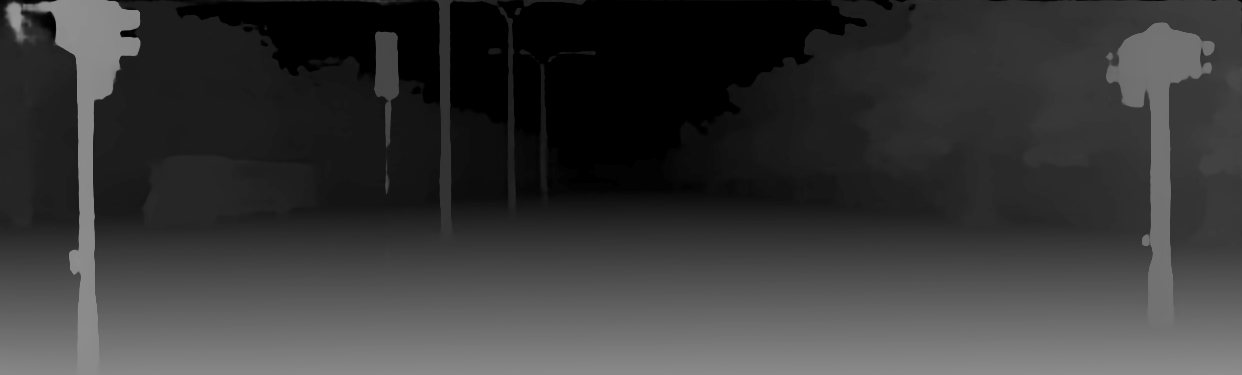} &
\includegraphics[width=\twdf]{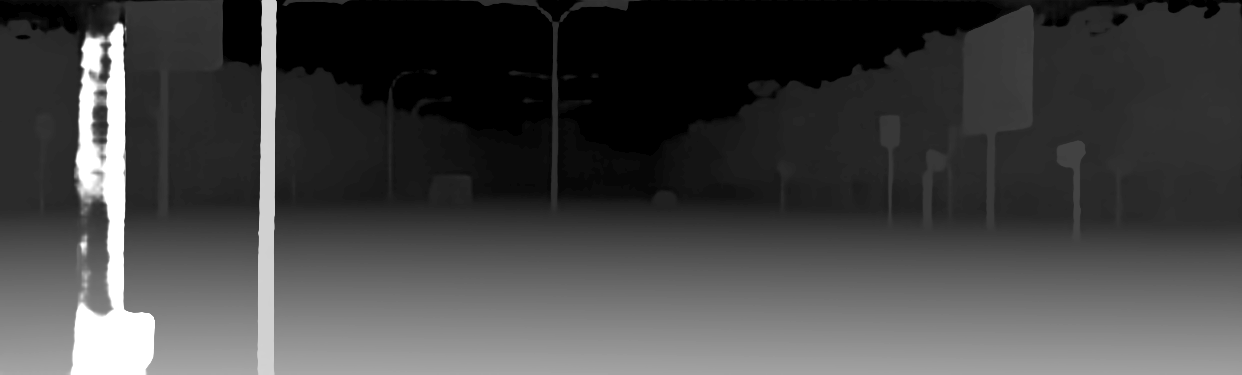} &
\includegraphics[width=\twdf]{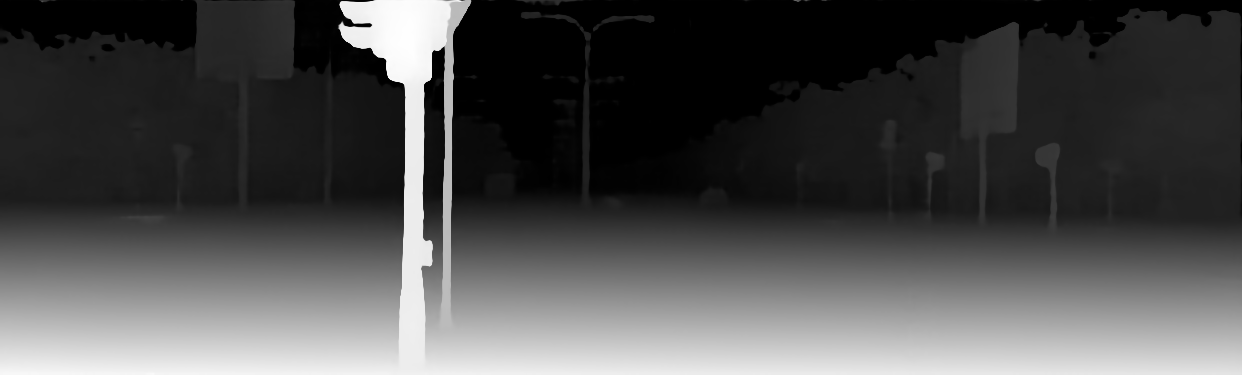} &
\\
& GwcNet &\\
\includegraphics[width=\twdf]{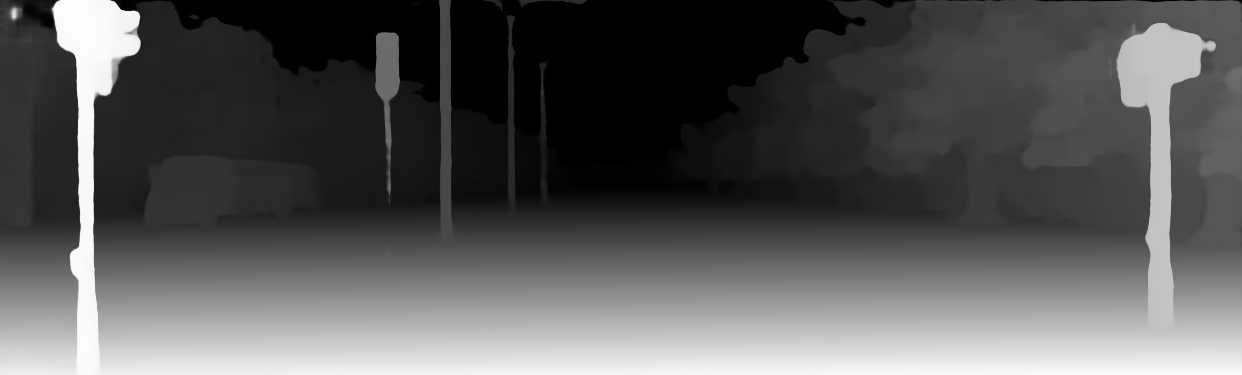} &
\includegraphics[width=\twdf]{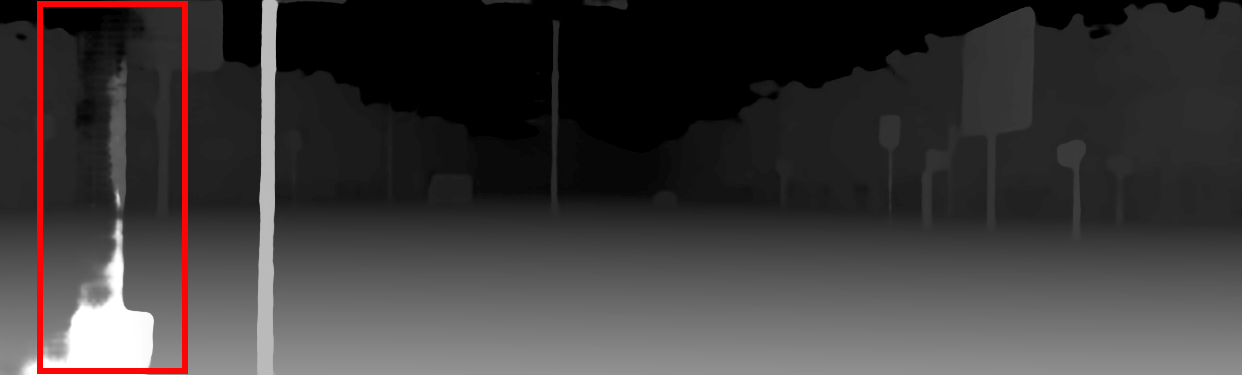} &
\includegraphics[width=\twdf]{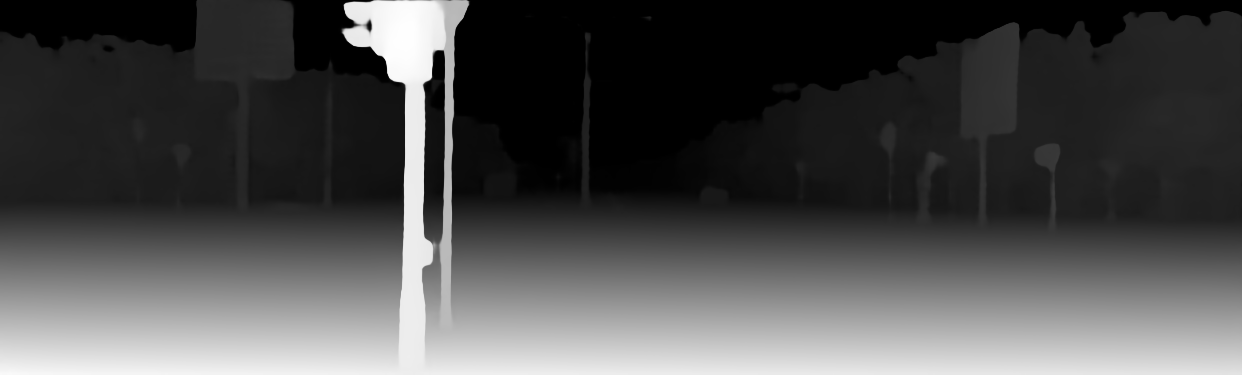} &
\\
& \kg &\\
\includegraphics[width=\twdf]{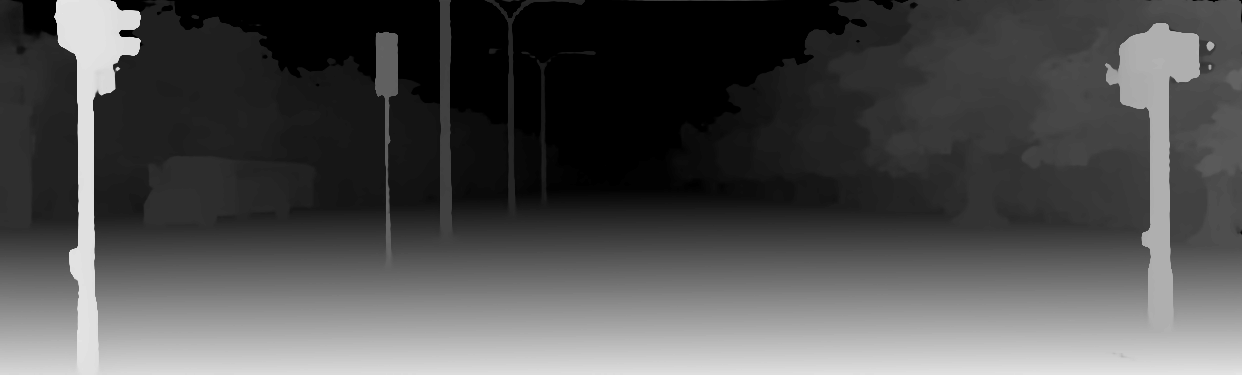} &
\includegraphics[width=\twdf]{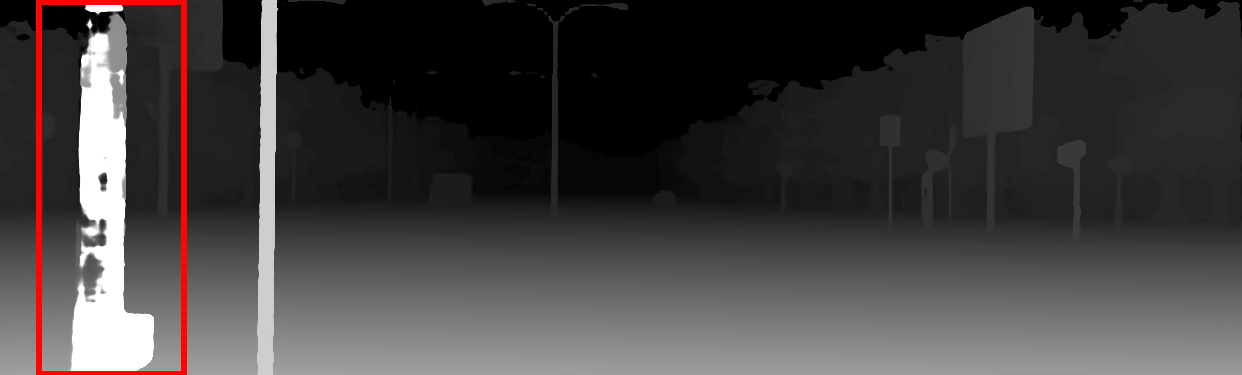} &
\includegraphics[width=\twdf]{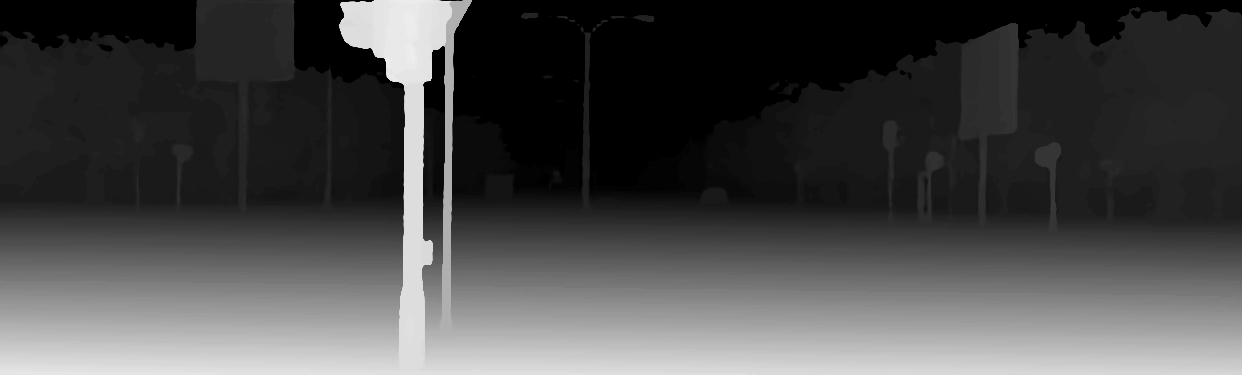} &
\\

& \netname &\\

\end{tabular}
\end{adjustbox}
\vspace{-4pt}
\caption{Examples of driving in different weather conditions from VK2-S6. We pick the best \kg and \netname based on EPE in Table~\ref{tab:in_domian}. (Please zoom in to see details.) \netname is better at predicting fine, challenging details in the disparity maps, such as the precise shape of the traffic light and the car under the left shadow in the sunny picture, the traffic sign behind the light post and the street light in the middle of the image in the fog. }
\label{fig:vk_weather}
\end{figure*}

\begin{figure*}[p]
\centering
\begin{tabular}{cc}
\includegraphics[width=\hcwdf]{fig/exp/VK2/GWC/foggy/l_img.png} &
\includegraphics[width=\hcwdf]{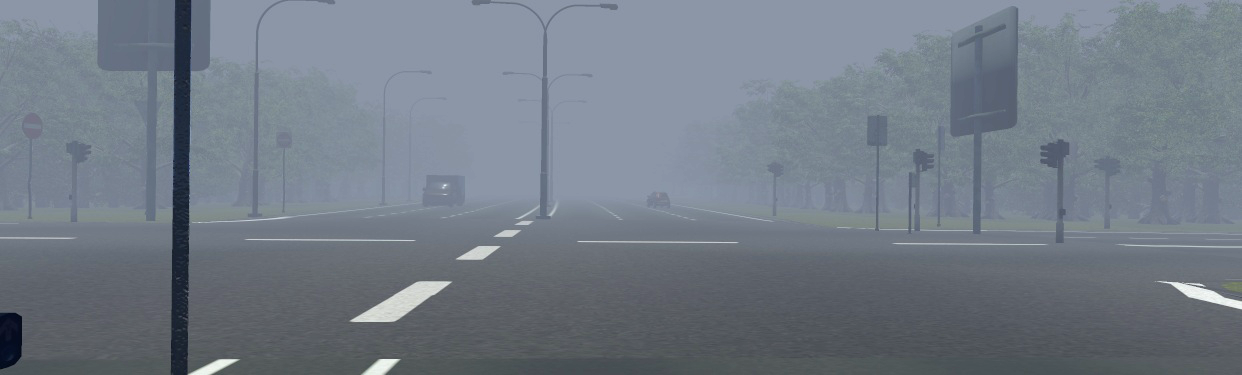}\\
Left Image & Right Image \\
\includegraphics[width=\hcwdf]{fig/exp/VK2/KG/foggy/d_est.png} &
\includegraphics[width=\hcwdf]{fig/exp/VK2/SED/foggy/d_est.png}\\
\kg Disparity & \netname Disparity \\
\includegraphics[width=\hcwdf]{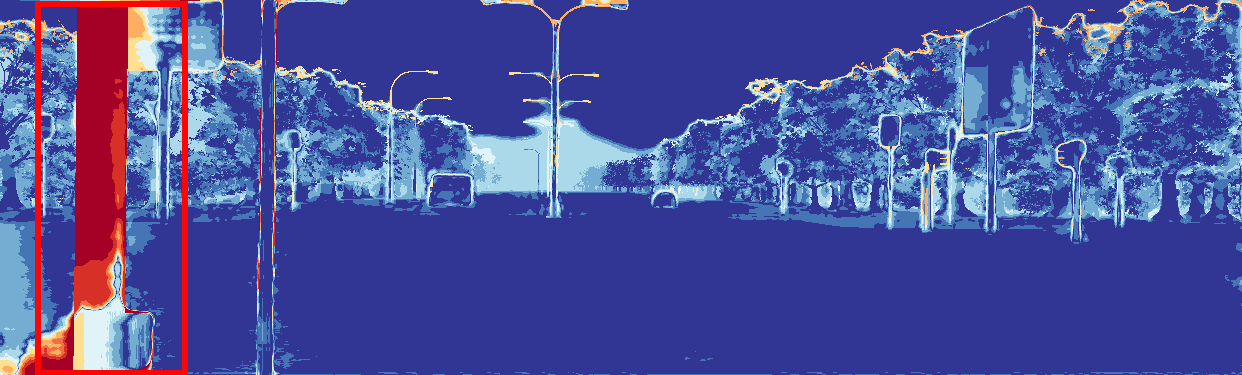} &
\includegraphics[width=\hcwdf]{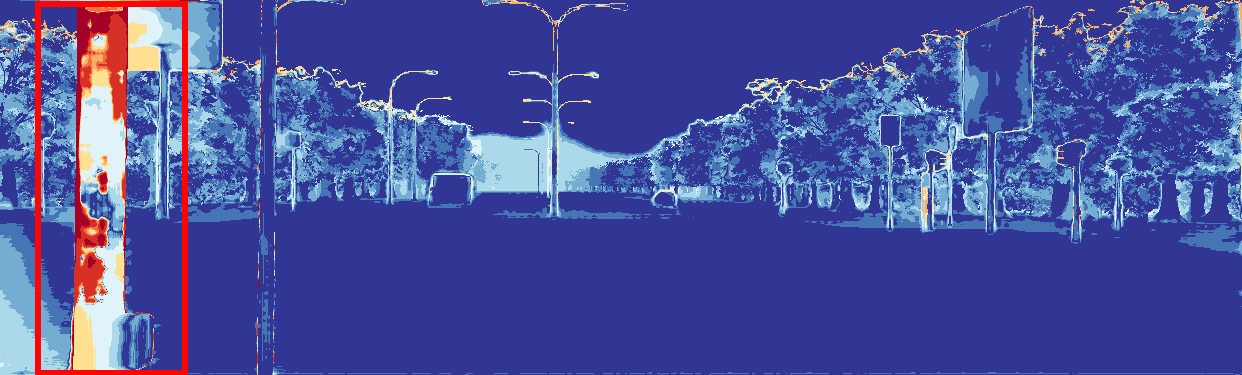}\\
\kg Error & \netname Error \\
\includegraphics[width=\hcwdf]{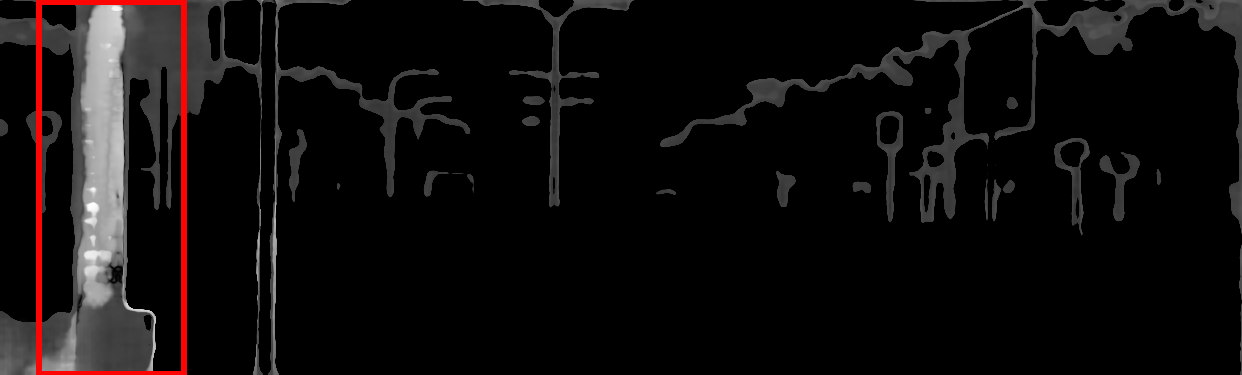} &
\includegraphics[width=\hcwdf]{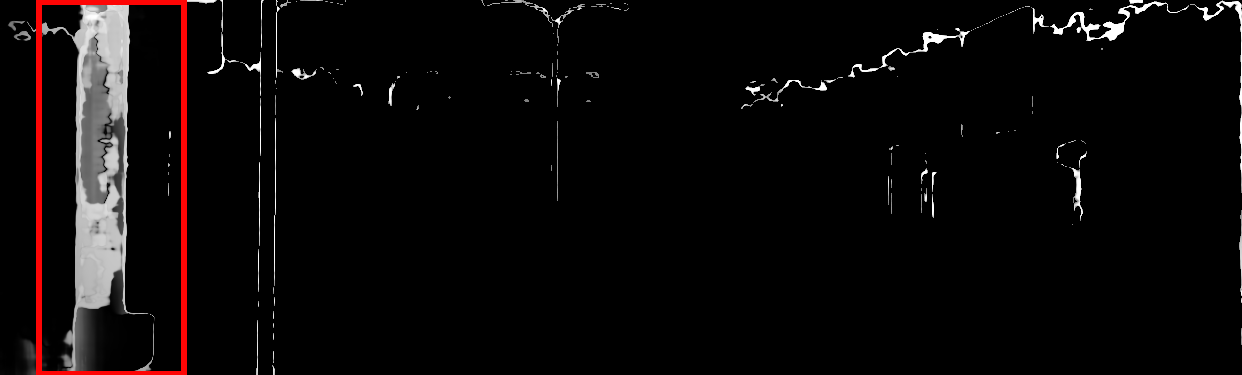}\\
\kg Uncertainty & \netname Uncertainty \\

\end{tabular}
\vspace{-4pt}
\caption{Example from VK2-S6-Fog. The post only appears in the left image, \kg fails to predict its disparity, but \netname does. The uncertainty map of \netname matches the error map better than that of \kg. }
\label{fig:vk_1}
\end{figure*}

\begin{figure*}[p]
\centering
\begin{tabular}{cc}
\includegraphics[width=\hcwdf]{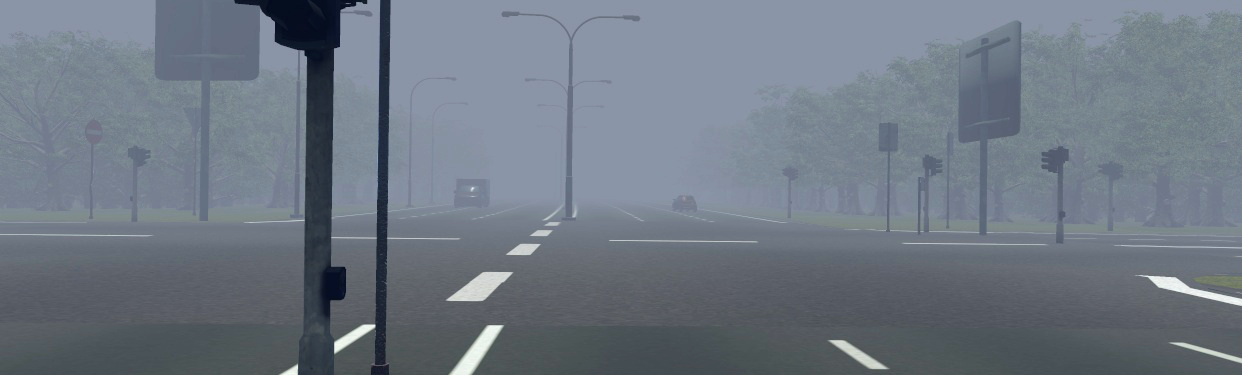} &
\includegraphics[width=\hcwdf]{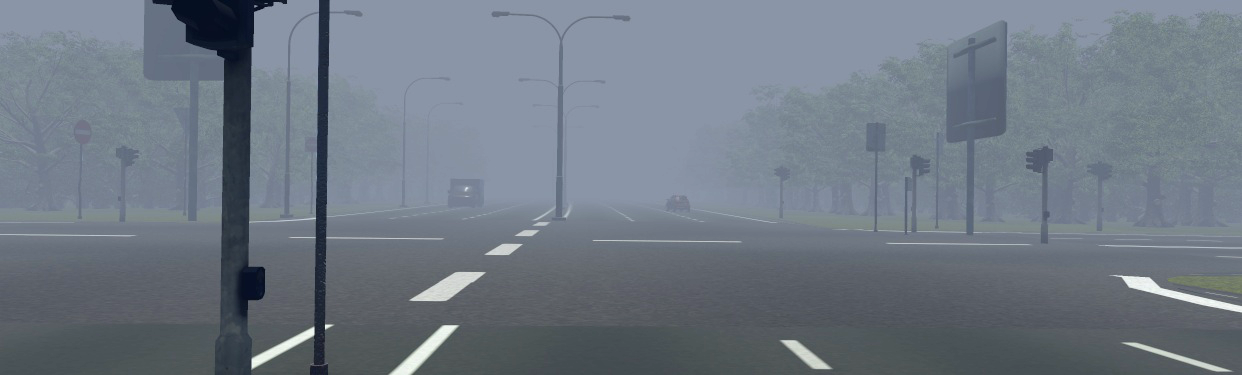}\\
Left Image & Right Image \\
\includegraphics[width=\hcwdf]{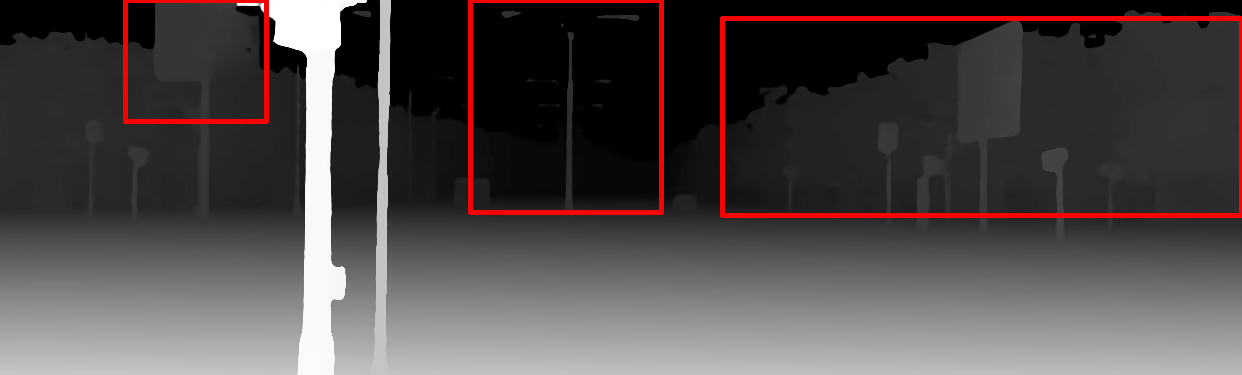} &
\includegraphics[width=\hcwdf]{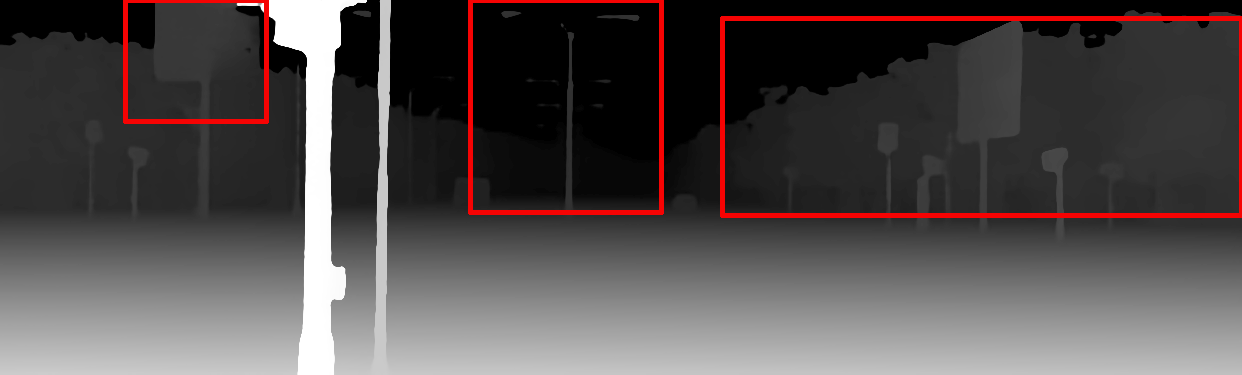}\\
\kg Disparity & \netname Disparity \\
\includegraphics[width=\hcwdf]{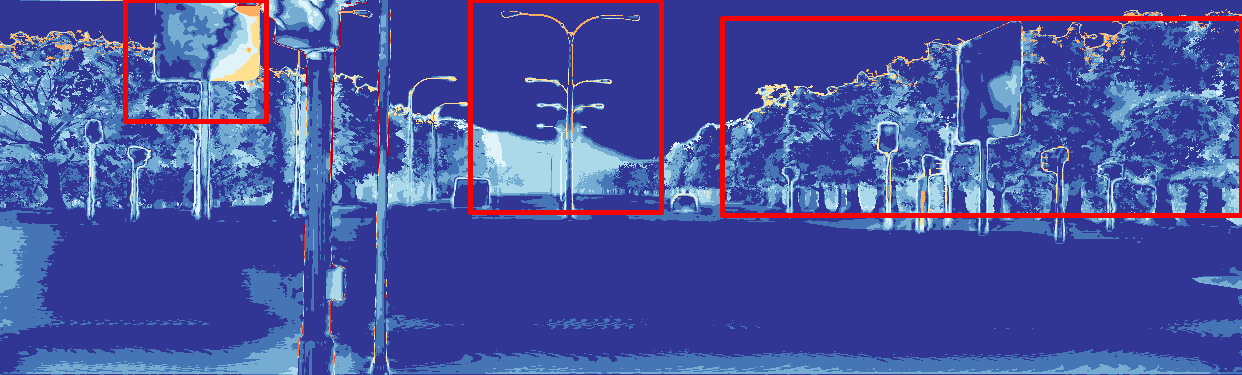} &
\includegraphics[width=\hcwdf]{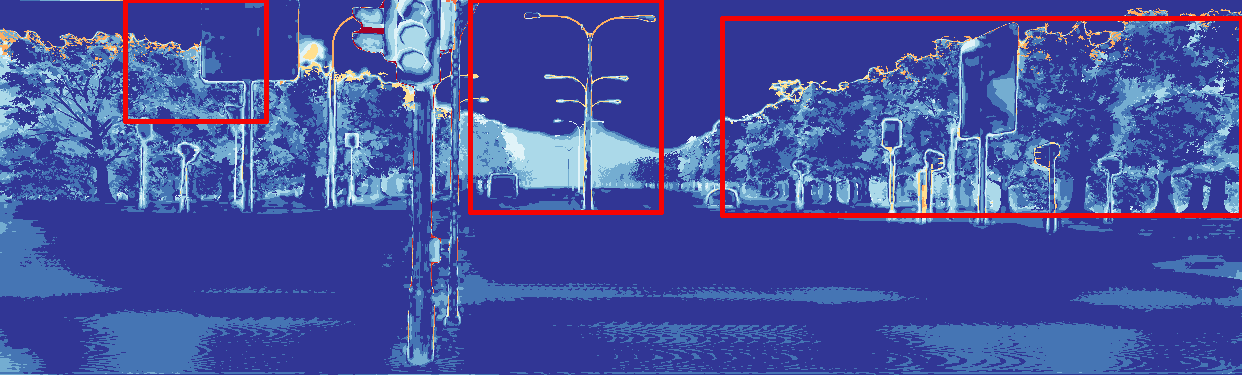}\\
\kg Error & \netname Error \\
\includegraphics[width=\hcwdf]{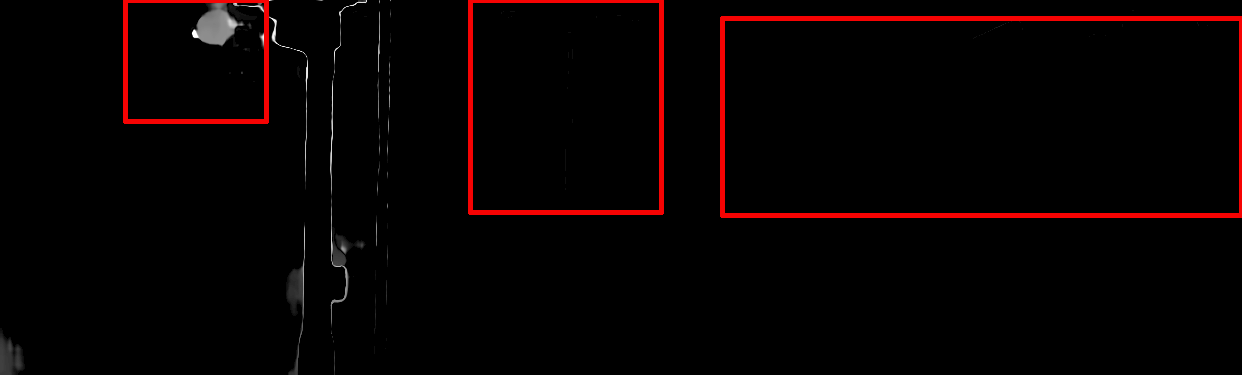} &
\includegraphics[width=\hcwdf]{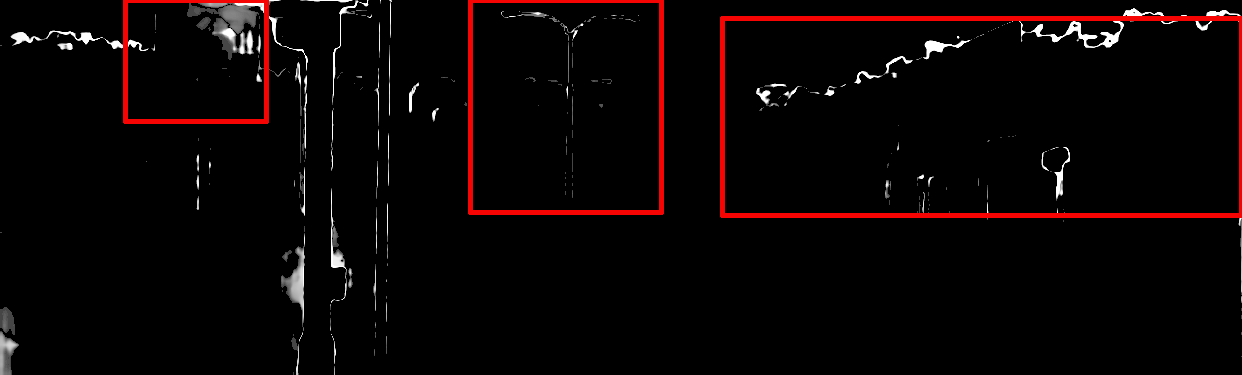}\\
\kg Uncertainty & \netname Uncertainty \\

\end{tabular}
\vspace{-4pt}
\caption{Example from VK2-S6-Fog. Contrasting error and uncertainty maps, we observe that \netname predicts more accurate uncertainty in the regions covered by fog, such as the edges of the traffic signs and the trees in the background.} 
\label{fig:vk_s2}
\end{figure*}


\begin{figure*}[p]
\begin{adjustbox}{width=\textwidth,center}
\centering
\begin{tabular}{ccc}
\includegraphics[width=\twdf]{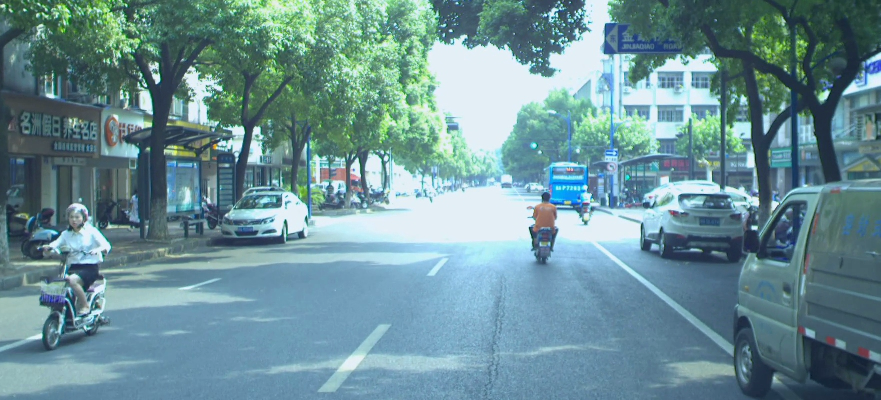} &
\includegraphics[width=\twdf]{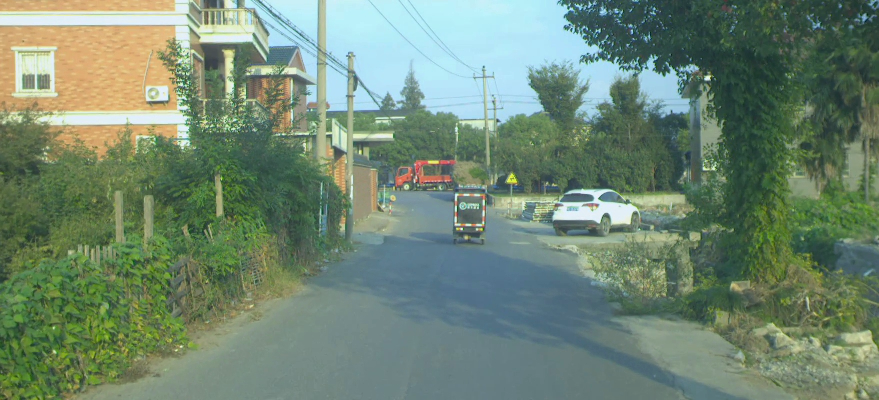} &
\includegraphics[width=\twdf]{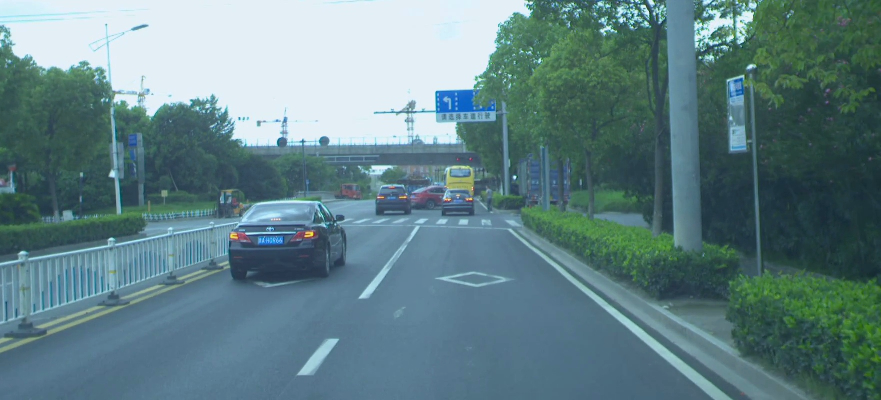} \\
& Left Image & \\

\includegraphics[width=\twdf]{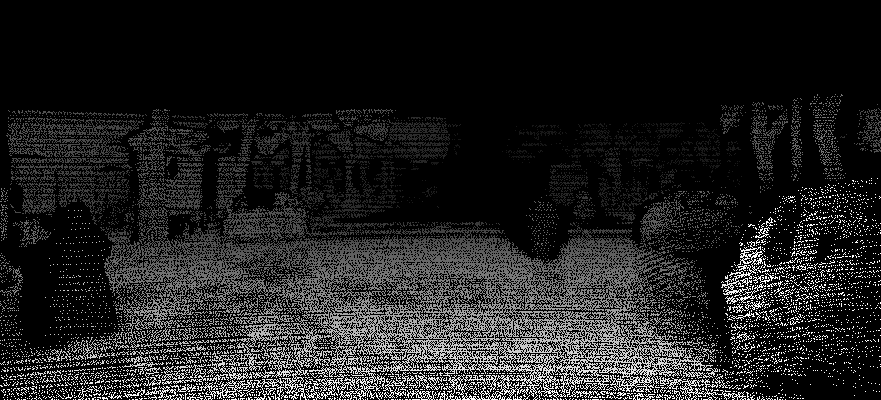} &
\includegraphics[width=\twdf]{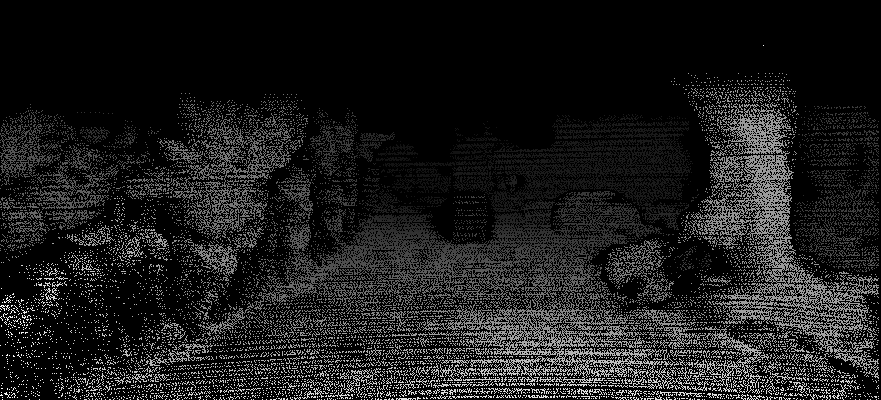} &
\includegraphics[width=\twdf]{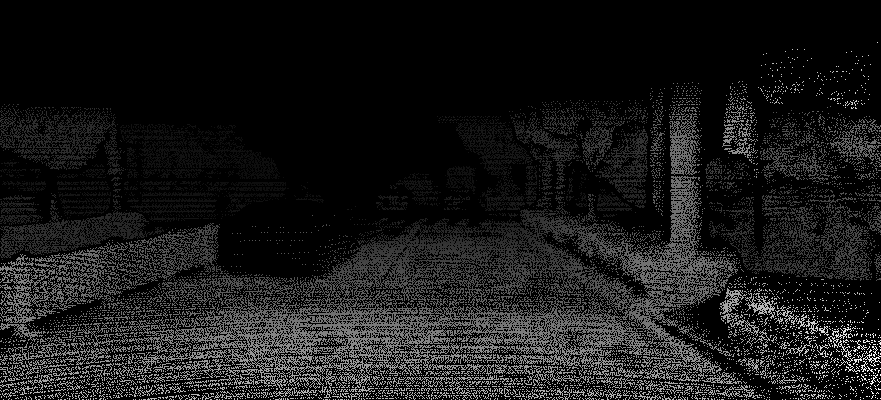} \\
& Ground Truth & \\

\includegraphics[width=\twdf]{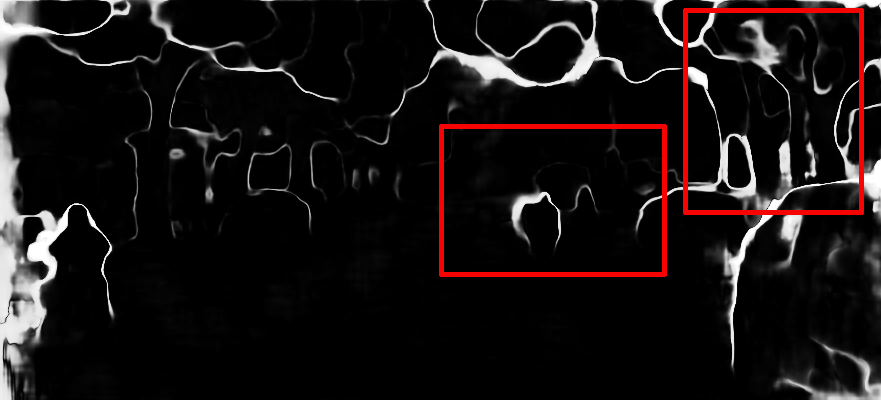} &
\includegraphics[width=\twdf]{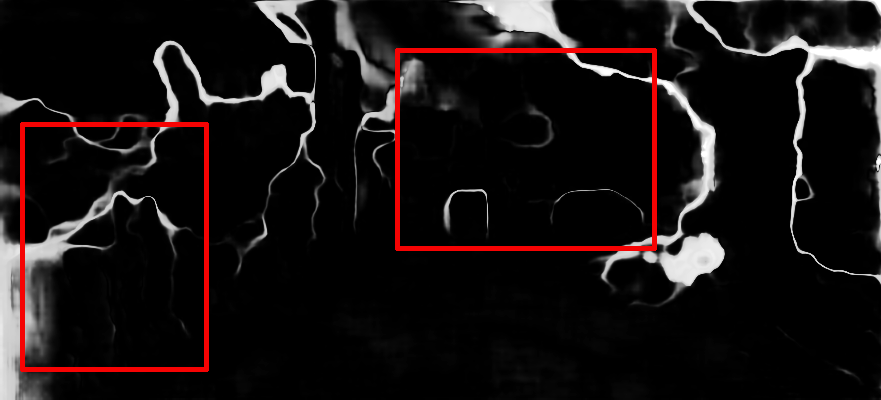} &
\includegraphics[width=\twdf]{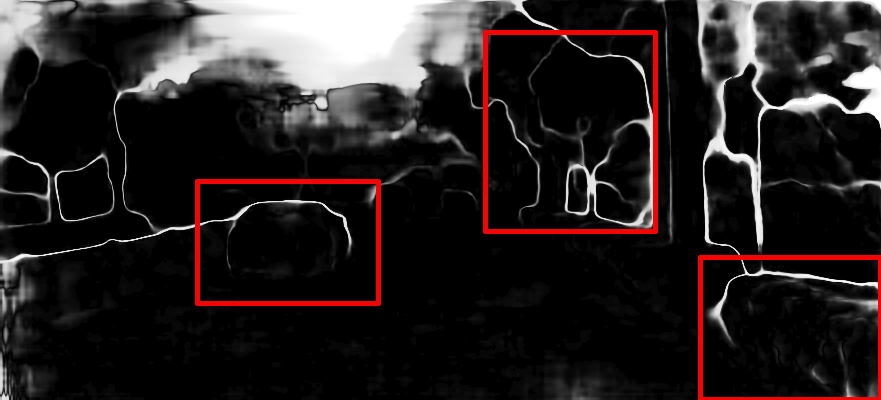} \\
& \kg & \\

\includegraphics[width=\twdf]{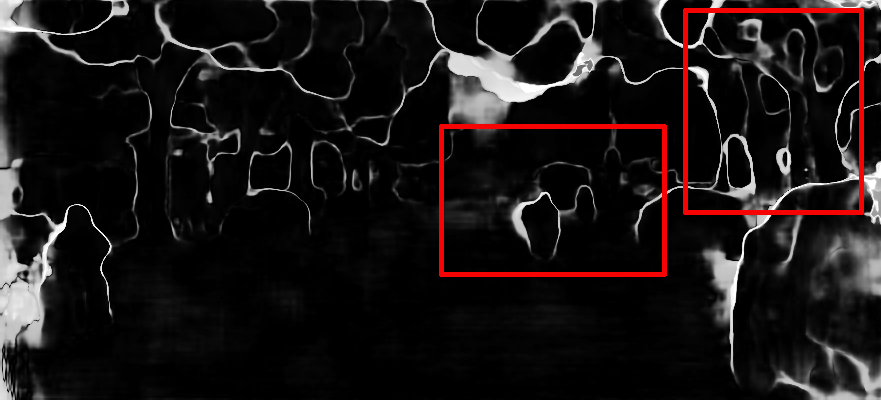} &
\includegraphics[width=\twdf]{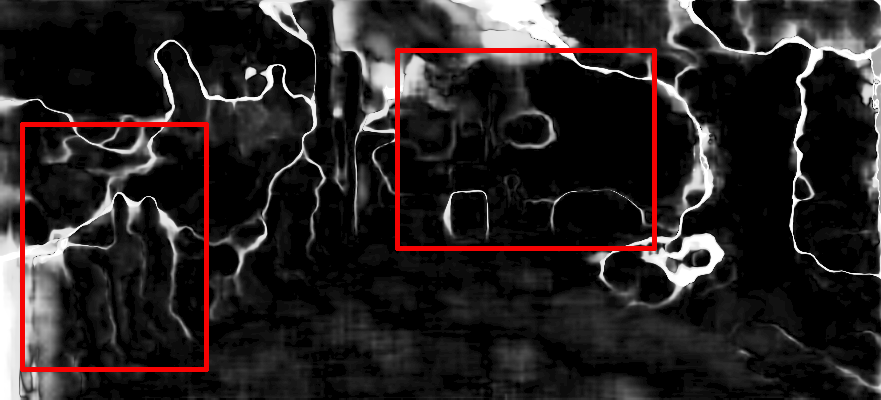} &
\includegraphics[width=\twdf]{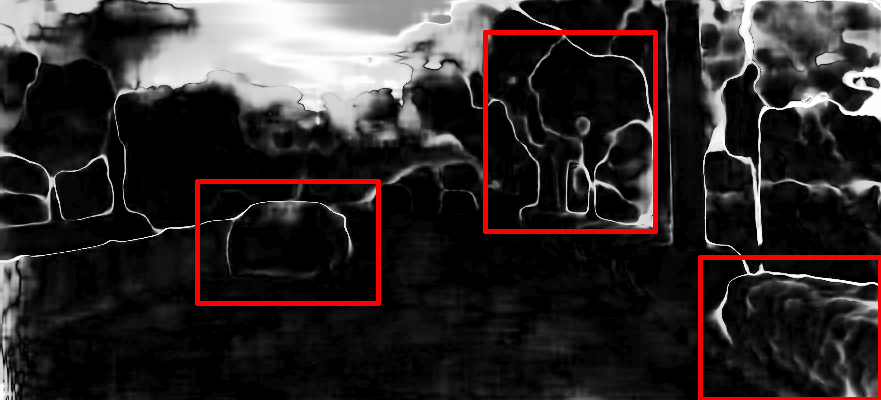} \\
& \netname &\\

\end{tabular}
\end{adjustbox}
\vspace{-4pt}
\caption{Examples of uncertainty estimation on \textit{DrivingStereo} also shown in Table~\ref{tab:in_domian}. \netname captures details more faithfully. For example, the cars, pedestrians and trees at different depths, even in overexposed parts of the images.}
\label{fig:ds_s_est}
\end{figure*}

\begin{figure*}[p]
\centering
\begin{tabular}{cc}
\includegraphics[width=\hcwdf]{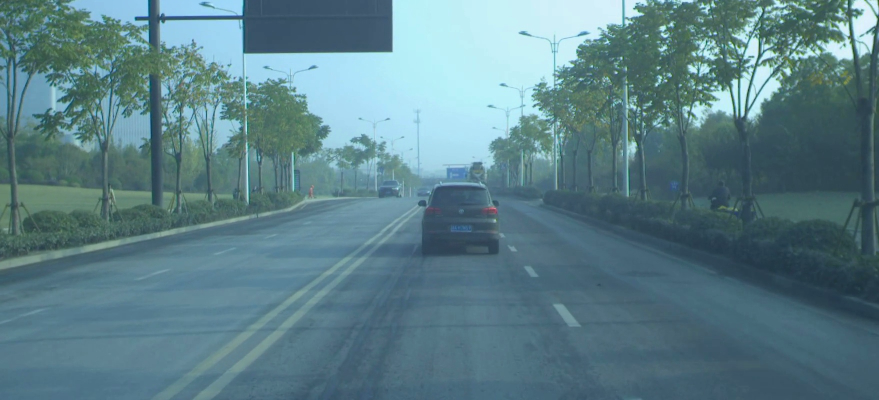} &
\includegraphics[width=\hcwdf]{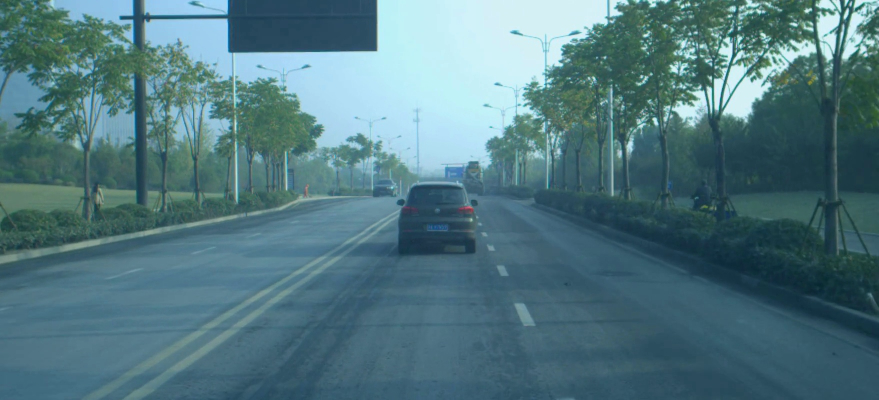}\\
Left Image & Right Image \\
\includegraphics[width=\hcwdf]{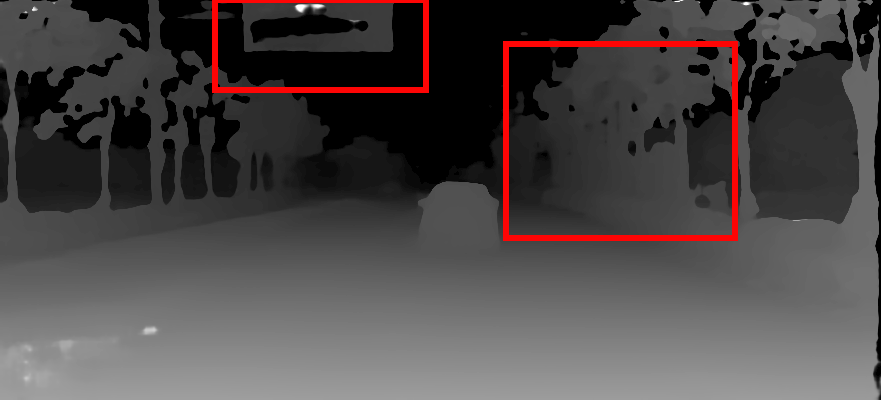} &
\includegraphics[width=\hcwdf]{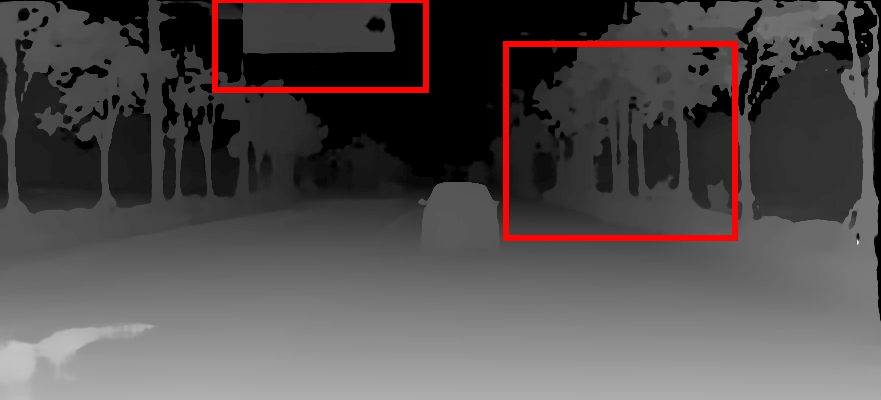}\\
\kg Disparity & \netname Disparity \\
\includegraphics[width=\hcwdf]{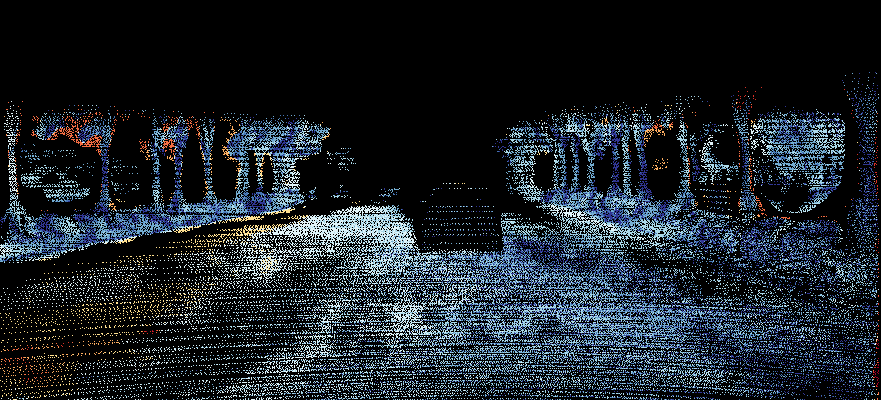} &
\includegraphics[width=\hcwdf]{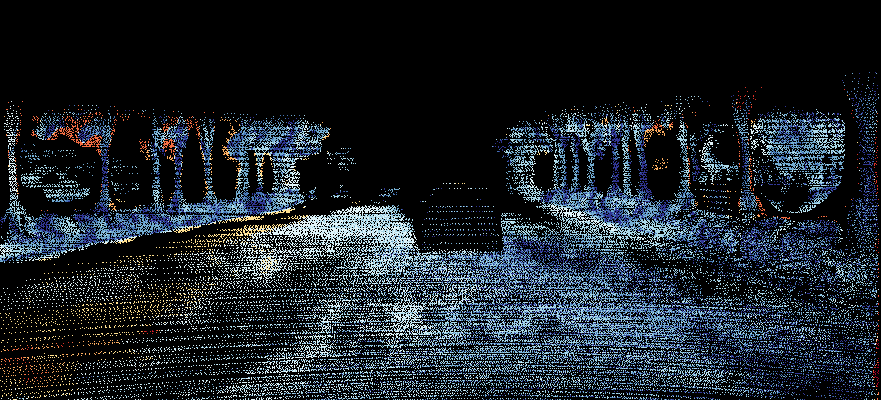}\\
\kg Error & \netname Error \\
\includegraphics[width=\hcwdf]{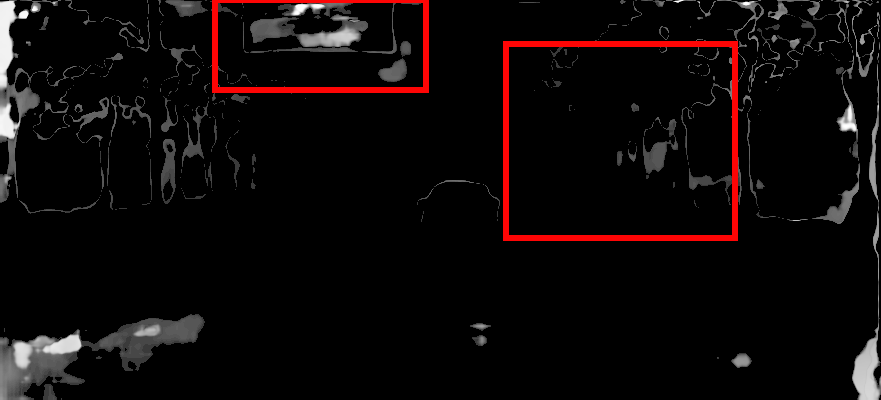} &
\includegraphics[width=\hcwdf]{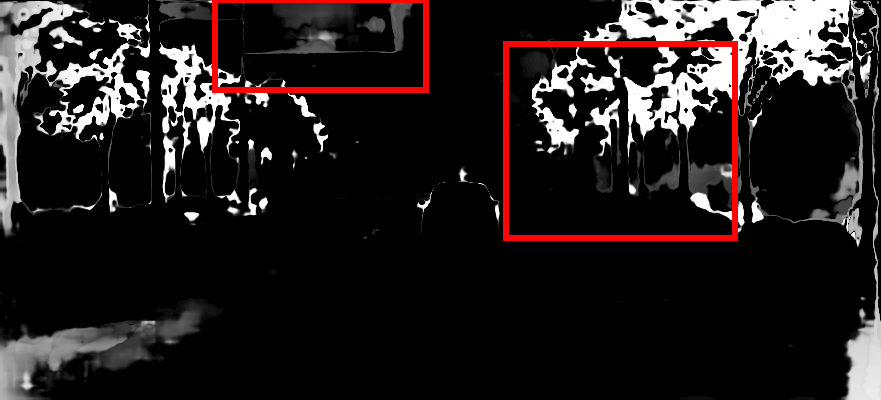}\\
\kg Uncertainty & \netname Uncertainty \\

\end{tabular}
\vspace{-4pt}
\caption{Example from DS-Foggy. 
This subset is more challenging than the corresponding synthetic data due to unmodeled sources of noise.
In this example, the thin tree trunk almost blends in with the background. Also, the billboard is very dark. \kg fails to predict the disparity of the textureless part of the billboard, as well as the space between the tree trunks on the right-hand side. The prediction of \netname is more accurate on these challenging parts.}
\label{fig:ds_1}
\end{figure*}

\begin{figure*}[p]
\centering
\begin{tabular}{cc}
\includegraphics[width=\hcwdf]{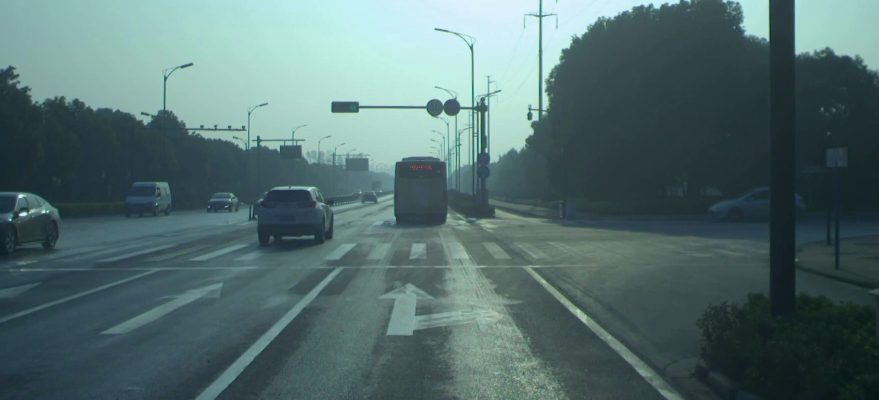} &
\includegraphics[width=\hcwdf]{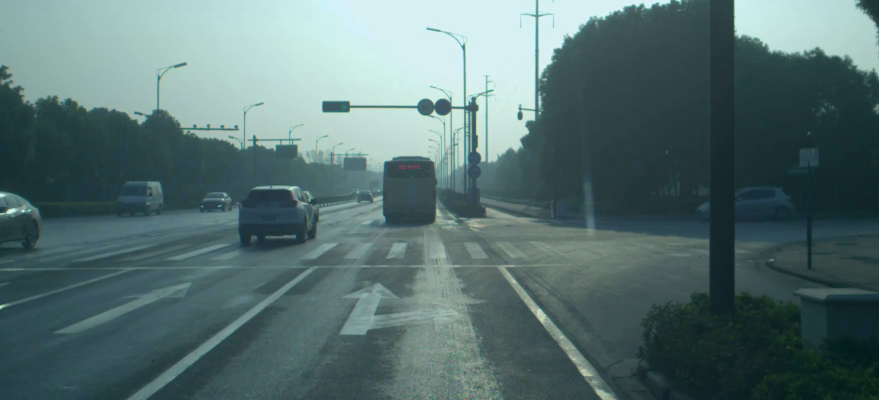}\\
Left Image & Right Image \\
\includegraphics[width=\hcwdf]{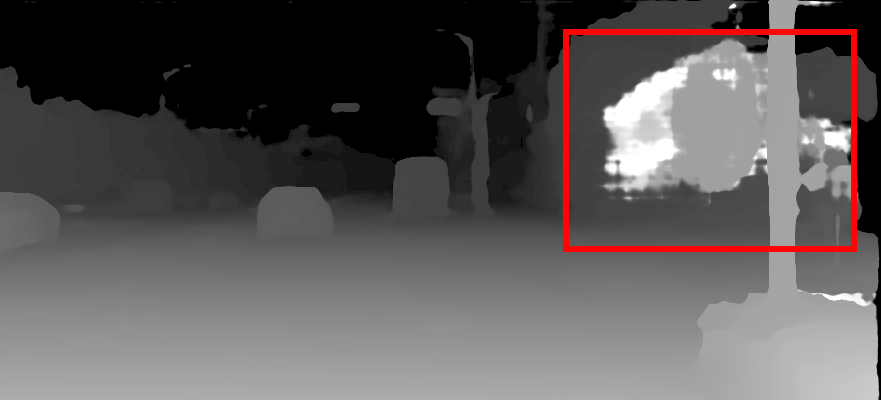} &
\includegraphics[width=\hcwdf]{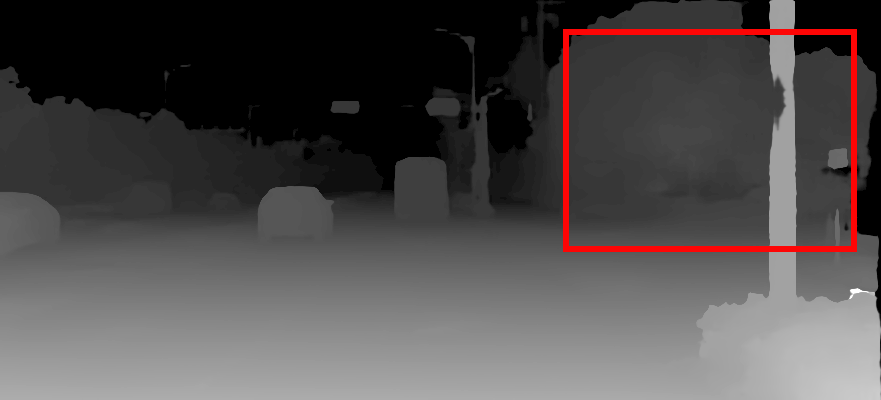}\\
\kg Disparity & \netname Disparity \\
\includegraphics[width=\hcwdf]{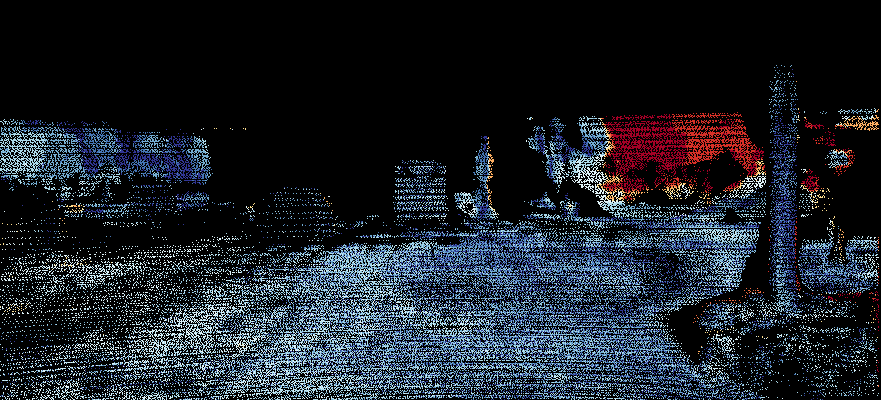} &
\includegraphics[width=\hcwdf]{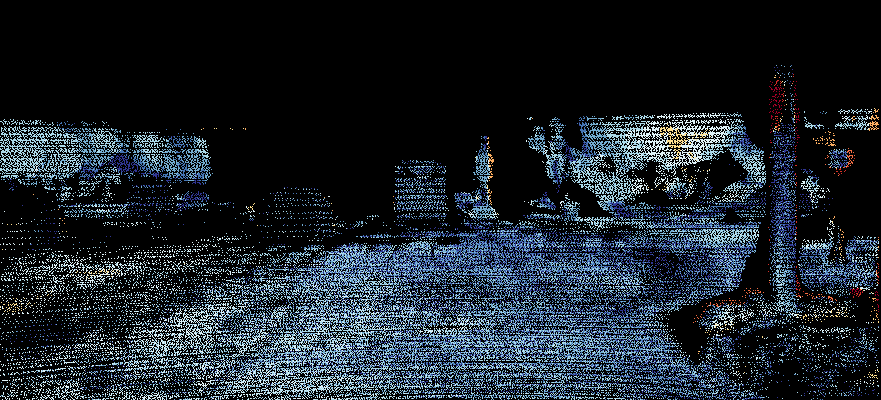}\\
\kg Error & \netname Error \\
\includegraphics[width=\hcwdf]{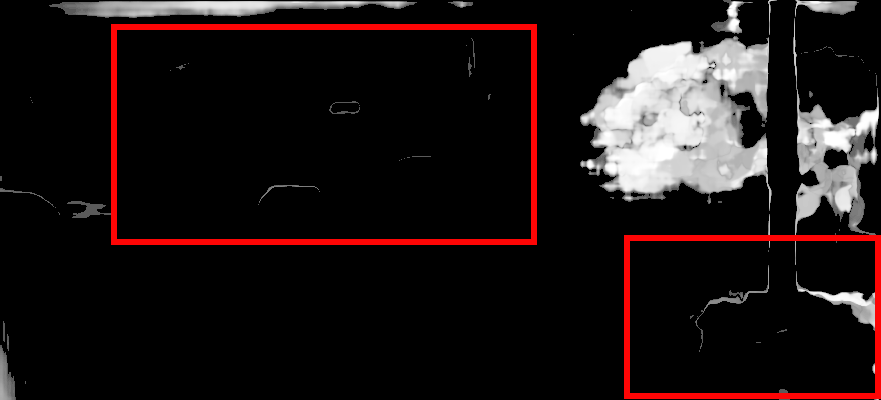} &
\includegraphics[width=\hcwdf]{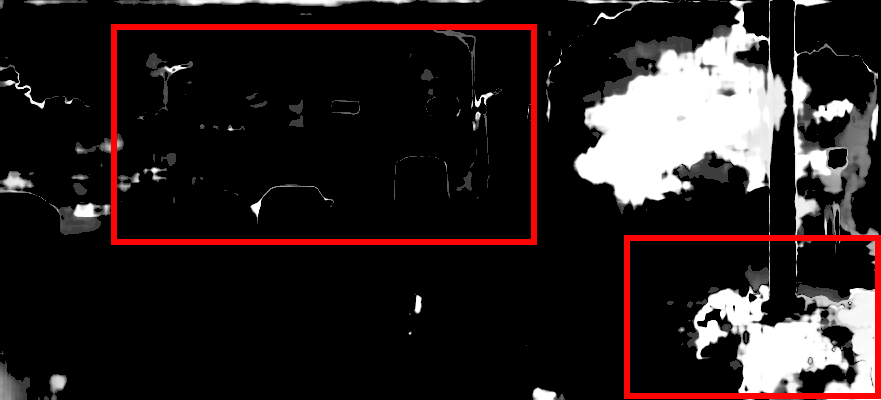}\\
\kg Uncertainty & \netname Uncertainty \\

\end{tabular}
\vspace{-4pt}
\caption{Example from DS-Foggy. Similar to the synthetic data, images from the foggy day subset are usually very dark, which makes distinguishing objects in the shadow difficult. In this example, \kg makes a mistake in predicting the disparity of the trees on the right side, since they have similar color to the post. On the other hand, the prediction of \netname is more accurate. The uncertainty maps are dominated by the right backlit regions, making it hard to see the other parts. However, zooming in the figures reveals that \netname still performs better in predicting the uncertainty of the objects far from the camera and in the bottom right dark corner.}
\label{fig:ds_s2}
\end{figure*}

\begin{figure*}[p]
\centering
\begin{tabular}{cc}
\includegraphics[width=\hcwdf]{fig/exp/DS/KG/rainy/l_img.png} &
\includegraphics[width=\hcwdf]{fig/exp/DS/KG/rainy/r_img.png}\\
Left Image & Right Image \\
\includegraphics[width=\hcwdf]{fig/exp/DS/KG/rainy/d_est.png} &
\includegraphics[width=\hcwdf]{fig/exp/DS/SED/rainy/d_est.png}\\
\kg Disparity & \netname Disparity \\
\includegraphics[width=\hcwdf]{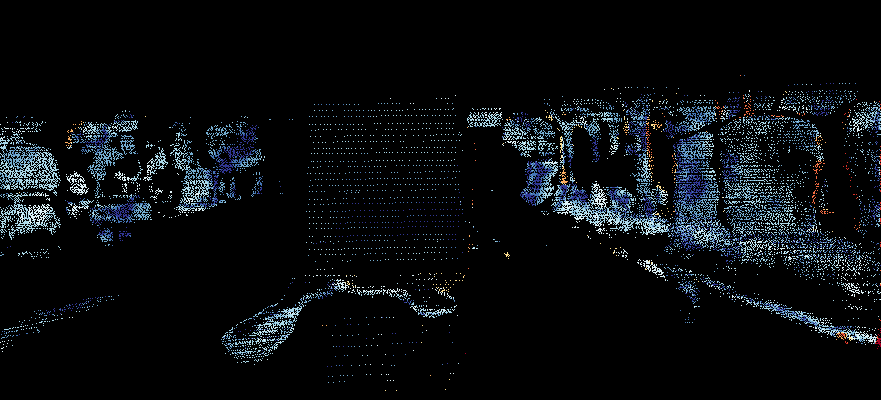} &
\includegraphics[width=\hcwdf]{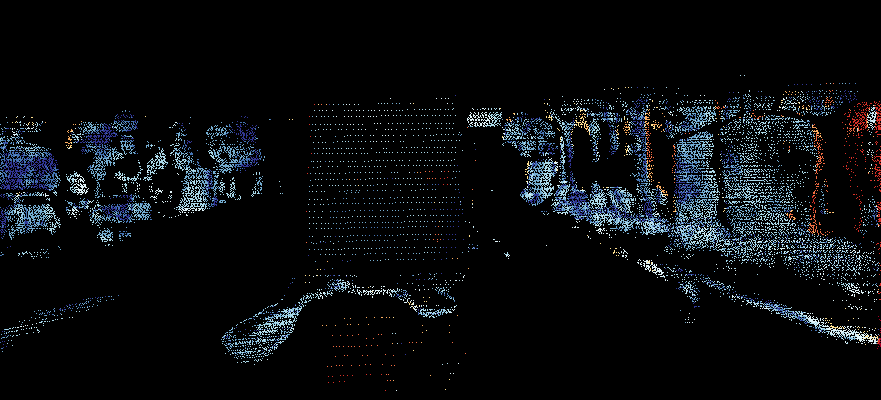}\\
\kg Error & \netname Error \\
\includegraphics[width=\hcwdf]{fig/exp/DS/KG/rainy/s_est.png} &
\includegraphics[width=\hcwdf]{fig/exp/DS/SED/rainy/s_est.png}\\
\kg Uncertainty & \netname Uncertainty \\

\end{tabular}
\vspace{-4pt}
\caption{Example from DS-Rainy. Unlike the synthetic data, the rainy-day real images do not only suffer from poor illumination, but also face challenges due to reflections in the %
water. In this example, the road is like a mirror, misleading the \kg model. Recall that the LIDAR ground truth disparity is very sparse, and is even sparser in reflective regions. Zooming in is required to see the recorded disparity errors on the road in the error maps. The disparity map of \kg fails to distinguish the car and the reflection, but \netname is able to estimate the correct disparity and the uncertainty of the car. }
\label{fig:ds_3}
\end{figure*}

\end{document}